\documentclass{article}

     \usepackage[final]{neurips_2024}

\usepackage[utf8]{inputenc} %
\usepackage[T1]{fontenc}    %
\usepackage{hyperref}       %

\usepackage{url}            %
\usepackage{booktabs}       %
\usepackage{amsfonts}       %
\usepackage{nicefrac}       %
\usepackage{microtype}      %
\usepackage{xcolor}         %

\usepackage{hyperref}
\hypersetup{
	colorlinks=true, 
    citecolor=blue, 
    linkcolor=blue,
    urlcolor=blue,
	pdfborder={0 0 0},
}

\usepackage[T1]{fontenc}
\usepackage[utf8]{inputenc}

\usepackage{here}

\usepackage{caption}
\usepackage{subcaption}

\usepackage{placeins}
\usepackage{float}

\usepackage{mathrsfs}
\usepackage{mathtools}
\usepackage{ctable}
\usepackage{footnote}
\usepackage{footnote}
\usepackage{threeparttable}
\usepackage{colortbl}
\usepackage{environ}
\usepackage{booktabs}
\usepackage{diagbox} 
\usepackage{wrapfig} 
\usepackage{bbding}
\usepackage{pifont}
\usepackage{wasysym}
\usepackage{amssymb}
\usepackage{times}
\usepackage{soul}
\usepackage{url}
\usepackage{caption}
\usepackage{graphicx}
\usepackage{amsmath}
\usepackage{amsthm}
\usepackage{tabularx}
\usepackage{multirow}
\usepackage{fancybox}
\usepackage{mathtools}
\usepackage{bold-extra}
\usepackage{lipsum}
\usepackage{siunitx}
\usepackage[export]{adjustbox}
\usepackage{comment}

\usepackage{cleveref}
\crefformat{equation}{(#2#1#3)}
\crefrangeformat{equation}{(#3#1#4) to (#5#2#6)}
\usepackage{cleveref}

\DeclarePairedDelimiterX{\infdivx}[2]{(}{)}{%
  #1\;\delimsize\|\;#2%
}

\makeatletter
\let\MYcaption\@makecaption
\makeatother
\usepackage{subcaption}
\makeatletter
\let\@makecaption\MYcaption
\makeatother

\makeatletter
\newcommand{\printfnsymbol}[1]{%
  \textsuperscript{\@fnsymbol{#1}}%
}
\makeatother
\usepackage{stmaryrd}
\usepackage{trimclip}
\makeatletter
\DeclareRobustCommand{\shortto}{%
  \mathrel{\mathpalette\short@to\relax}%
}

\usepackage{etoolbox}

\makeatletter

\newcommand*{\red}[1]{{\textcolor{red}{#1}}}

\newtheorem{theorem}{Theorem}[section]

\NewEnviron{alignSmall}{%
    \scriptsize
    \begin{align}
    \BODY
    \end{align}
    \normalsize
}
\NewEnviron{gatherSmall}{%
    \begin{gather}
    \BODY
    \end{gather}
    \normalsize
}

\usepackage{bussproofs}
\usepackage{stackengine}
\usepackage{varwidth}

\newenvironment{mathprooftree}
  {\varwidth{.9\textwidth}\centering\leavevmode}
  {\DisplayProof\endvarwidth}

\newcommand*{\colorGreenDeductionRule}[1]{{\textcolor[HTML]{3e9288}{#1}}}

\newcommand*{\colorBlueFacts}[1]{{\textcolor[HTML]{00b1f0}{#1}}}
\newcommand*{\colorRedLogicalSteps}[1]{{\textcolor[HTML]{f52727}{#1}}}
\newcommand*{\colorVioletHypothesis}[1]{{\textcolor[HTML]{4545c2}{#1}}}

\usepackage{tcolorbox}
\usepackage{bbm}
\usepackage{setspace}

\newcommand*{\numBenchmarks}{31}
\newcommand*{\PLD}[0]{Formal Logic Deduction Diverse}
\newcommand*{\PLDItalic}[0]{Formal Logic \textit{D}eduction \textit{D}iverse}
\newcommand*{\PLDAbbr}[0]{FLD$_{\times \mathbbm{2}}$}

\newcommand*{\ALT}[0]{ALT}

\newcommand*{\llamaThreeBaseline}[0]{LLaMA-3.1-8B}

\newcommand*{\llamaThreeRTALT}[0]{{$\oplus$\ALT-RT}}

\newcommand*{\llamaThreePRSingle}[0]{{$\oplus$\ALT-PRP} $_{\text{w/o RecAdam}}$}
\newcommand*{\llamaThreePRALT}[0]{{$\oplus$\ALT-PRP}}

\newcommand*{\llamaThreeFLDALT}[0]{{$\oplus$\ALT-FLD}}

\newcommand*{\llamaThreePLDALT}[0]{{$\oplus$\textbf{ALT}-{{FLD}$_{\times \mathbbm{2}}$}}}
\newcommand*{\llamaThreePLDALTWoBold}[0]{{$\oplus$ALT-{{FLD}$_{\times \mathbbm{2}}$}}}

\newcommand*{\llamaThreePLDALTBold}[0]{$\oplus$\textbf{ALT}-\textbf{{FLD}$_{\times \mathbbm{2}}$}}

\newcommand*{\llamaThreePLDALTWODOne}[0]{{ w/o  DP1}}
\newcommand*{\llamaThreePLDALTWODTwo}[0]{{ w/o  DP2}}
\newcommand*{\llamaThreePLDALTWODThreeRules}[0]{{ w/o  DP3.rules}}
\newcommand*{\llamaThreePLDALTWODThreeSteps}[0]{{ w/o  DP3.steps}}

\newcommand*{\llamaThreePLDALTWODFour}[0]{{ w/o  DP4}}

\newcommand*{\llamaThreeLargeBaseline}[0]{LLaMA-3.1-70B}

\newcommand*{\llamaThreeLargeRTALT}[0]{{$\oplus$\ALT-RT}}

\newcommand*{\llamaThreeLargePRSingle}[0]{{$\oplus$\ALT-PRP}$_{\text{w/o RecAdam}}$}
\newcommand*{\llamaThreeLargePRALT}[0]{{$\oplus$\ALT-PRP}}

\newcommand*{\llamaThreeLargeFLDALT}[0]{{$\oplus$\ALT-FLD}}

\newcommand*{\llamaThreeLargePLDALTBold}[0]{$\oplus$\textbf{ALT}-\textbf{{FLD}$_{\times \mathbbm{2}}$}}

\title{Enhancing Reasoning Capabilities of LLMs \\ via Principled Synthetic Logic Corpus}

\author{
Terufumi Morishita$^{1}$\quad \quad Gaku Morio$^{1*}$\quad \quad Atsuki Yamaguchi$^{2*\dagger}$\quad \quad Yasuhiro Sogawa$^{1}$\\
{}\\
$^1$Advanced AI Innovation Center, Hitachi \quad \quad
$^2$The University of Sheffield
}

\begin{document}

\def\thefootnote{*}\footnotetext{Equal Contribution}\def\thefootnote{\arabic{footnote}}
\def\thefootnote{$\dagger$}\footnotetext{Work done at Hitachi}\def\thefootnote{\arabic{footnote}}

\maketitle

\begin{abstract}
Large language models (LLMs) are capable of solving a wide range of tasks, yet they have struggled with reasoning.
To address this, we propose \textbf{Additional \textit{Logic} Training (\ALT)}, which aims to enhance LLMs' reasoning capabilities by program-generated logical reasoning samples.
We first establish principles for designing high-quality samples by integrating symbolic logic theory and previous empirical insights.
Then, based on these principles, we construct a synthetic corpus named \textbf{\PLDItalic} \ (\PLDAbbr), comprising numerous samples of multi-step deduction with unknown facts, diverse reasoning rules, diverse linguistic expressions, and challenging distractors.
Finally, we empirically show that \ALT \ on \PLDAbbr \ substantially enhances the reasoning capabilities of state-of-the-art LLMs, including \llamaThreeLargeBaseline.
Improvements include gains of up to 30 points on logical reasoning benchmarks, up to 10 points on math and coding benchmarks, and 5 points on the benchmark suite BBH.
\end{abstract}

\section{Introduction}   \label{sec:introduction}

Knowledge and reasoning have long been considered essential elements for achieving \textit{artificial intelligence} \citep{Mccarthy1959ProgramsWC,weizenbaum1966eliza,winograd1971procedures,colmerauer1973prolog,shortliffe1976computer,elkan1993building}.
Knowledge refers to facts about the world, e.g., ``objects with mass generate a gravitational field'' and ``the Earth has mass.''
Reasoning involves combining multiple facts according to specific rules to obtain new knowledge.
For example, the new knowledge that ``the Earth generates a gravitational field'' can be derived from the aforementioned two facts.

Recent observations suggest that LLMs can solve problems using memorized knowledge of similar samples seen during pre-training, but they cannot solve novel, unknown problems that require reasoning \citep{hodel2023response,dasgupta2023language,zhang2024careful}.
For instance, LLMs can solve famous arithmetic problems as is but not when the numbers or names are changed \citep{razeghi2022impact,mirzadeh2024gsmsymbolicunderstandinglimitationsmathematical}, and they can solve coding tests from past years before the ``knowledge cutoff'' but not from the present year \citep{melanie2023blog}.
This bias towards knowledge has been observed even in state-of-the-art LLMs such as GPT-4 \citep{liu2023evaluating,wu2023reasoning,dziri2023faith}.

\begin{figure*}[h!]
    \centering
    \includegraphics[width=1.0\linewidth]{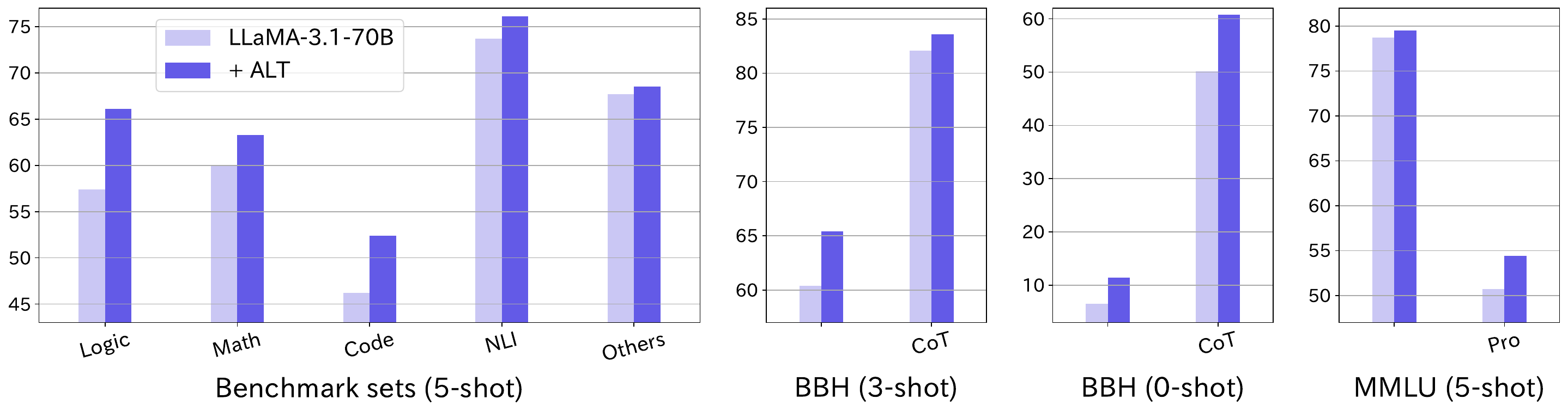}
    \vspace{-5mm}
    \caption{
        The performance gains to \llamaThreeLargeBaseline \ by Additional Logic Training (\ALT) on the proposed synthetic corpus, \PLDAbbr \ (\PLDItalic).
        Each benchmark set, such as ``Logic'' and ``Math'', comprises various benchmarks in that domain.
        \Cref{tb:performance_aggregated,tb:performance_details} shows the details.
        \label{fig:performance_comparison}
    }
    \vspace{-5mm}
\end{figure*}

LLMs' poor reasoning capabilities can stem from the lack of high-quality reasoning samples in the pre-training corpus, which primarily consists of human-written texts \citep{betz-etal-2021-critical,pmlr-v202-morishita23a}.
Indeed, reasoning samples in human-written texts often exhibit low quality, as evidenced by fallacies and biases commonly found in online debates \citep{hansson2004fallacies,guiacsu2018logical,Cheng:2017ud}.
This is unsurprising given that humans usually think reflexively rather than through rigid reasoning \citep{kahneman2011thinking,SunsteinHastie2015,Paglieri2017}.
Thus, a straightforward strategy to improve LLMs' reasoning capabilities is to prepare many high-quality reasoning samples and train LLMs on them.

\begin{figure*}[t] 
    \centering
    \includegraphics[width=1.0\linewidth]{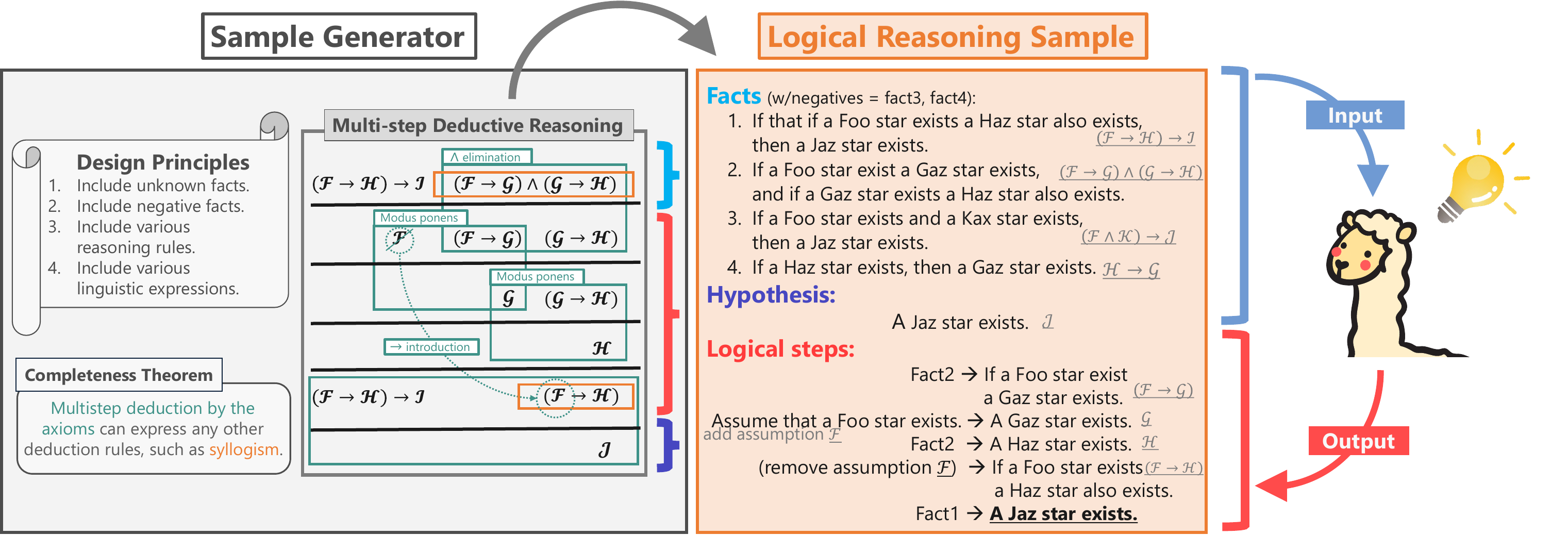}
    
    \caption{
        Our proposed \textbf{A}dditional \textbf{L}ogic \textbf{T}raining (\ALT) aims to enhance LLMs' reasoning capabilities through training on many synthetically generated logical reasoning samples.
        Our sample generator (left) first generates a sample of multi-step deductive reasoning and then converts it into a deduction sample written in English (right).
        LLMs must generate \textbf{\colorRedLogicalSteps{logical steps}} to derive a given \textbf{\colorVioletHypothesis{hypothesis}} from provided \textbf{\colorBlueFacts{facts}}.
        The sample generator adheres to theoretically and empirically grounded \textit{design principles} discussed in \Cref{sec:design_principles}.
        Refer to \Cref{appendix:fig:deduction_example} for a real sample.
        \label{fig:ALT_overview}
    }
\end{figure*}

We propose one such approach, \textbf{Additional \textit{Logic} Training (\ALT)}, which utilizes high-quality samples of \textit{logical} reasoning, the most fundamental form of reasoning.
To prepare such samples, we utilize synthetic generation \citep{clark2020transformers,betz-etal-2021-critical,tafjord-etal-2021-proofwriter,pmlr-v202-morishita23a}, where computer programs generate deductive reasoning samples in which a given hypothesis is proven or disproven by combining given facts following rigid reasoning rules.
We overview \ALT \ in \Cref{fig:ALT_overview}.

In synthetic generation, computer programs generate samples according to pre-designed patterns, so this design largely determines the quality of these samples by nature.
Thus, we start by discussing \textbf{what is the ideal design for synthetic logic samples}, incorporating symbolic logic theory and empirical findings (\Cref{sec:design_principles}).
The essence of logical reasoning lies in its ability to handle unknown facts, unlike knowledge, which deals solely with established facts, such as commonsense facts; therefore, samples must cover reasoning with unknown facts.
Samples must include both \textit{illogical} and logical reasoning to enable LLMs to distinguish between them.
The samples must cover various patterns regarding a comprehensive set of reasoning aspects, such as reasoning rules and linguistic expressions of logical statements.
We summarize these discussions into \textit{design principles}, which guide the design of synthetic logic samples.
Finally, based on these principles, we construct a synthetic corpus named \textbf{\PLDItalic} (\PLDAbbr), comprising numerous samples of multi-step deduction with unknown facts, diverse reasoning rules, diverse linguistic expressions, and challenging distractors (\Cref{sec:PLD}).

We then empirically verify that \ALT \ can enhance LLMs' reasoning capabilities (\Cref{sec:experiments,sec:results_and_discussions_method}).
Using \numBenchmarks \ benchmarks covering diverse tasks, we observed that \ALT \ on \PLDAbbr \ substantially boosts state-of-the-art LLMs' reasoning capabilities.
Even \llamaThreeLargeBaseline, the largest LLM pre-trained on over 15 trillion tokens, shows substantial improvements with \ALT \ (\Cref{fig:performance_comparison}).
Among synthetic logic corpora with different sample designs, \PLDAbbr \ yielded the largest performance gains, validating our proposed design principles.
Moreover, we discovered that employing a knowledge-forgetting prevention method during \ALT \ is critically important, as it likely prevents the LLM's knowledge of established facts from being displaced by the unknown facts included in synthetic logic corpora.

Finally, we analyze which task-solving capabilities \ALT\ can enhance and why (\Cref{sec:results_and_discussions_tasks}).
We observed a substantial improvement of up to 30 points on logical reasoning tasks (\Cref{tb:which_task_logical_reasoning}).
Surprisingly, we also observed improvements in abductive reasoning tasks, which go beyond the synthetic logic corpora's original deductive reasoning tasks.
Case analyses indicate that these improvements result from LLMs having acquired the fundamentals of the logic reflected in the design principles.
We also observed improvements of up to 10 points on math and coding tasks, indicating the generalizability of the obtained reasoning capabilities (\Cref{tb:which_task_math,tb:which_task_coding}).
We also observed improvements of up to 6 points on natural language inference (NLI) tasks (\Cref{tb:which_task_NLI}).
Case analyses suggest that LLMs successfully integrated the commonsense knowledge they had originally acquired during pre-training with the logical reasoning capabilities newly acquired from \ALT.

Improvements across various other tasks (\Cref{tb:which_task_others}) demonstrate the broad benefits of the obtained reasoning capabilities beyond standard reasoning tasks, though the modest improvements of up to 2 points indicate the need for future research on more effective application of these capabilities.

\newpage
Our contributions are summarized as follows:
\begin{itemize}
    \vspace{-0.5\baselineskip} %
    \setlength{\itemsep}{1mm}
    \setlength{\leftskip}{-8mm}
    \item We propose \textbf{A}dditional \textit{\textbf{L}ogic} \textbf{T}raining (\ALT) and empirically verify that it can enhance the reasoning capability of state-of-the-art LLMs across various sizes, from 7B to 70B.
    \item We establish systematic design principles for synthetic logic samples; then, we construct a synthetic corpus named \textbf{\PLDItalic} \ (\PLDAbbr), comprising numerous samples of multi-step deduction with unknown facts, diverse reasoning rules, diverse linguistic expressions, and challenging distractors. We empirically verify that \PLD \ indeed leads to the largest improvements among corpora with different sample designs.
    \item We demonstrate that LLMs enhanced by \ALT \ can solve not only the original logical reasoning tasks present in synthetic logic corpora but also other tasks, such as math and coding tasks, and notably NLI tasks, which require integrating knowledge and reasoning. This finding underscores the potential for advancing truly versatile AI possessing both knowledge and reasoning capabilities.
\end{itemize}
\vspace{-2mm}
We release the corpus, code, and the trained model under a permissive license \footnote{\scriptsize \url{https://github.com/hitachi-nlp/FLD}}.

\vspace{-1mm}
\section{How Should Synthetic Logic Samples Be Designed?} \label{sec:design_principles}
\vspace{-2mm}

In synthetic generation, computer programs generate samples according to pre-designed patterns, so this design largely determines the quality of the samples.
While Previous studies have examined several designs \citep{clark2020transformers,betz-etal-2021-critical,tafjord-etal-2021-proofwriter,pmlr-v202-morishita23a}, these designs were not systematically discussed, so they may not be the most effective ones.

Thus, we start by discussing how to optimally design synthetic logic samples.
To this end, we consider symbolic logic theory as suggested by \citet{pmlr-v202-morishita23a} and integrate empirical findings from previous studies.
First, we observe that the essence of logical reasoning, based solely on the logical relationships between facts, lies in its ability to handle unknown facts, unlike knowledge, which by definition deals solely with established facts (\Cref{sec:principle_unseen}).
Therefore, we argue that samples should cover reasoning with unknown facts to represent this essential aspect of logical reasoning.
We also observe that logical reasoning involves various other aspects, such as \textit{illogical} reasoning, reasoning rules, and linguistic expressions that represent logical statements (\crefrange{sec:principle_negatives}{sec:principle_linguistic_diversity}).
The samples should cover various patterns regarding these aspects to enable LLMs to solve various reasoning problems.
We summarize these discussions into the following \textit{design principles}, which guide the design of synthetic logic samples.

\clearpage

\vspace{-1mm}
\subsection{Teaching Reasoning with Unknown Facts} \label{sec:principle_unseen}
\vspace{-1mm}

We first explore the essence of logical reasoning that differentiates itself from knowledge.
Consider the following logical step:
\begin{equation}
    \small
    \begin{mathprooftree}
        \AxiomC{The Earth orbits the Sun.}
        \AxiomC{If the Earth orbits the sun, the Earth has four seasons.}
        \BinaryInfC{The Earth has four seasons.}
    \end{mathprooftree}
    \label{eq:argument:first}
\end{equation}
This step is valid because the conclusion is logically derived from the two premises.
Next, consider another logical step:
\begin{equation}
    \small
   \begin{mathprooftree}
        \AxiomC{The Earth orbits the Sun.}
        \AxiomC{If the Earth orbits the sun, the Earth \textit{does not have} four seasons.}
        \BinaryInfC{The Earth \textit{does not have} four seasons.}
    \end{mathprooftree}
    \label{eq:argument:second}
\end{equation}
The second premise and consequently, the conclusion, is factually wrong.
Nevertheless, \textit{if the premise was hypothetically correct}, the conclusion could be logically derived.
Therefore, step \Cref{eq:argument:second} is also logically valid.
Finally:
\begin{equation}
    \small
   \begin{mathprooftree}
        \AxiomC{1. A Foo star exists.}
        \AxiomC{2. If a Foo star exists, a Bar star also exists.}
        \BinaryInfC{A Bar star exists.}
    \end{mathprooftree}
    \label{eq:argument:third}
\end{equation}

``Foo star'' and ``Bar star'' are unknowns; nonetheless, we can still determine that step \Cref{eq:argument:third} is logically valid.
Steps \Crefrange{eq:argument:first}{eq:argument:third} above can be abstracted into a \textbf{deduction rule}, i.e., modus ponens, using symbols:
\begin{equation}
    \small
    \begin{mathprooftree}
        \AxiomC{$\mathcal{F}$}
        \AxiomC{$\mathcal{F} \rightarrow \mathcal{G}$}
        \RightLabel{\stackanchor{\small modus ponens}{\tiny}}
        \BinaryInfC{$\mathcal{G}$}
    \end{mathprooftree}
    \label{eq:argument:modus_ponens}
\end{equation}
As we have seen, the logical validity of a deduction rule depends solely on whether the conclusion is logically derived from the premises, not on the factual correctness of the contents of $\mathcal{F}$ and $\mathcal{G}$.
Therefore, \textit{the contents of $\mathcal{F}$ and $\mathcal{G}$ can be arbitrary.}

Now, we consider what kind of samples would be needed to teach the deduction rule \Cref{eq:argument:modus_ponens} to LLMs.
We assume a task to generate the conclusion given the premises as prompt inputs.
If the learner were human, they would be able to infer the underlying deduction rule \Cref{eq:argument:modus_ponens} by observing samples such as \Crefrange{eq:argument:first}{eq:argument:second}.
As a result, they would become able to solve the unknown problem \Cref{eq:argument:third}.

However, from a purely inductive perspective, samples \Crefrange{eq:argument:first}{eq:argument:second} cannot simply be generalized to the deduction rule \cref{eq:argument:modus_ponens}.
This is because the samples \Crefrange{eq:argument:first}{eq:argument:second} themselves do not contain the information that the contents of $\mathcal{F}$ and $\mathcal{G}$ are arbitrary.
In fact, one could generalize samples \Crefrange{eq:argument:first}{eq:argument:second} to other rules; for example, the conclusion $\mathcal{G}$ can be derived if $\mathcal{F}$ and $\mathcal{F} \rightarrow \mathcal{G}$ are given as premises \textit{and} $\mathcal{F}$ and $\mathcal{G}$ include 'Earth' as their contents.
Innumerable such deduction rules can be inductively inferred from the given samples.
In other words, induction has arbitrariness \citep{hume1748enquiry,goodman1954fact,quine1969epistemology}.

Humans prefer simpler rules \citep{russel1946,wittgenstein1922tractatus}, so they boldly induce up to the deduction rule \Cref{eq:argument:modus_ponens}.
However, it is unclear how purely inductive learners such as LLMs, which extract only what can be inferred from samples without prior preferences, induce up to \Cref{eq:argument:modus_ponens}.
For example, if only specific contents such as ``Alice is kind'' and ``Bob is smart'' are assigned to $\mathcal{F}$ and $\mathcal{G}$ in training samples, an LLM could develop into a machine that generates the conclusion $\mathcal{G}$ only when the input contains the specific contents.
In order for LLMs to accurately induce that $\mathcal{F}$ and $\mathcal{G}$ are indeed arbitrary:

\textbf{Design Principle 1} (Reasoning with Unknown Facts).
\textit{Prepare many samples assigning arbitrary contents to $\mathcal{F}$ and $\mathcal{G}$. They will make LLMs accurately induce $\mathcal{F}$ and $\mathcal{G}$ are indeed arbitrary, ultimately enabling them to reason with unknown facts.}

\vspace{-1mm}
\subsection{Teaching Illogical Reasoning} \label{sec:principle_negatives}
\vspace{-2mm}

Suppose we have LLMs trained on a large number of samples as follows:
\begin{equation}
    \small
    \begin{mathprooftree}
        \AxiomC{$\mathcal{F} \land \mathcal{G}$}
        \AxiomC{$(\mathcal{F} \land \mathcal{G}) \rightarrow \mathcal{H}$}
        \BinaryInfC{$\mathcal{H}$}
    \end{mathprooftree}
    \label{eq:argument:and_modus_ponens}
    \vspace{-0.5mm}
\end{equation}
where $\land$ denotes logical conjunction, and arbitrary contents are assigned to $\mathcal{F},\mathcal{G},\mathcal{H}$.
Suppose that we give this LLM a problem such as:
\begin{equation}
\small
    \begin{mathprooftree}
        \AxiomC{$\mathcal{F}$}
        \AxiomC{$(\mathcal{F} \land \mathcal{G}) \rightarrow \mathcal{H}$}
        \BinaryInfC{$??$}
    \end{mathprooftree}
    \label{eq:argument:and_modus_ponens_collapsed}
    \vspace{-0.5mm}
\end{equation}
Since the premises are insufficient for logically deducting the conclusion, outputting nothing is the correct answer.

Unfortunately, an LLM could output $\mathcal{H}$, which was indeed often observed in our preliminary experiments.
This is because while the LLMs can induce from sample \Cref{eq:argument:and_modus_ponens} that it can generate the conclusion $\mathcal{H}$ when the two premises of \Cref{eq:argument:and_modus_ponens} are given, the LLMs cannot induce from the sample that it is \textit{not allowed} to generate the conclusion $\mathcal{H}$ when the premises of \Cref{eq:argument:and_modus_ponens_collapsed} are given, as such information is not included in the sample \Cref{eq:argument:and_modus_ponens} itself.
Therefore,

\textbf{Design Principle 2} (Illogical Reasoning).
\textit{Include negative samples such as \Cref{eq:argument:and_modus_ponens_collapsed}. These samples will make LLMs induce that conclusions cannot be derived from insufficient premises.}

\vspace{-2mm}
\subsection{Teaching Diverse Reasoning Rules} \label{sec:principle_deduction_rules}
\vspace{-2mm}

Deduction rules other than \Cref{eq:argument:modus_ponens} exist:
\begin{align}
    \small
    \begin{mathprooftree}
    \Axiom $(\mathcal{F} \fCenter \land \mathcal{G})$
    \UnaryInf $\fCenter \mathcal{F}$
    \end{mathprooftree}
    \ 
    \begin{mathprooftree}
    \Axiom $(\mathcal{F} \fCenter \land \mathcal{G})$
    \RightLabel{\stackanchor{$\land$elimination}{}}
    \UnaryInf $\fCenter \mathcal{G}$
    \end{mathprooftree}
    &
    \ \  \ \ \
    \begin{mathprooftree}
    \Axiom $(\mathcal{F} \rightarrow \mathcal{G}) \fCenter \land (\mathcal{G} \rightarrow \mathcal{H})$
    \RightLabel{\stackanchor{\small syllogism}{}}
    \UnaryInf $\mathcal{F} \fCenter \rightarrow \mathcal{H}$
    \end{mathprooftree}
    \nonumber
    \\
    \begin{mathprooftree}
    \Axiom $\mathcal{F} \fCenter \rightarrow \mathcal{G}$
    \RightLabel{\stackanchor{\small contraposition}{}}
    \UnaryInf $\neg \mathcal{G} \fCenter \rightarrow \neg \mathcal{F}$
    \end{mathprooftree}
    &
    \ \  \ \ \
    \begin{mathprooftree}
    \Axiom $\neg(\mathcal{F} \fCenter \lor \mathcal{G})$
    \UnaryInf $\neg \mathcal{F} \fCenter \land \neg \mathcal{G}$
    \end{mathprooftree}
    \
    \begin{mathprooftree}
    \Axiom $\neg(\mathcal{F} \fCenter \land \mathcal{G})$
    \RightLabel{\stackanchor{\small De Morgan's laws}{}}
    \UnaryInf $\neg \mathcal{F} \fCenter \lor \neg \mathcal{G}$
    \end{mathprooftree}
    \label{eq:argument:others}   
    \vspace{-0.5mm}
\end{align}
where $\lor$ denotes logical disjunction and $\neg$ negation.
Since there are infinitely many possible logical formulas that can appear as premises and conclusions, there are infinitely many deduction rules.
Providing LLMs with these infinite deduction rules is obviously intractable.

Instead of directly providing these infinite deduction rules, we can take another approach.
Consider multi-step deductive reasoning (\Cref{fig:ALT_overview} left), where multiple deduction rules derive a conclusion.
Notice that the syllogism in \Cref{eq:argument:others} can be expressed by multi-step deductive reasoning using more ``atomic'' deduction rules.
Indeed, there exists a set of atomic deduction rules called \textbf{the axioms} that satisfies the following:

\begin{theorem}[Completeness of First-Order Predicate Logic \citet{godel1930uber}]
    \label{theorem:completeness}
    Any valid deduction rule can be expressed by multistep deductive reasoning constructed from the axioms.
\end{theorem}
In contrast to the axioms, the `compound' deduction rules, such as syllogism, contraposition, and De Morgan's laws, are called theorems.
According to the completeness \Cref{theorem:completeness}, if we can handle the axioms, we can effectively handle other deduction rules as well.
Indeed, \citet{pmlr-v202-morishita23a} empirically verified that a language model trained on the axioms generalizes to handle other deduction rules more effectively than those trained on non-axiom deduction rules.
Therefore,

\textbf{Design Principle 3} (Diverse Reasoning Rules).
\textit{Samples should express multi-step deduction constructed from the axioms. They will effectively teach LLMs diverse deduction rules {\small\citep{pmlr-v202-morishita23a}}}

In multi-step deductive reasoning, the number of logical steps $s$ from premises to a conclusion can vary largely depending on the problem.
Therefore:

\textbf{Design Principle 3'} (Diverse Reasoning Rules).
\textit{Samples should include diverse numbers of logical steps $s$.}

Ideally, this would be sufficient, but empirical evidence has shown that LLMs struggle with constructing multi-step deductive reasoning with large steps $s$ \citep{gontier2020measuring,pmlr-v202-morishita23a}.
Consequently, LLMs would not excel at handling theorems that require a large number of steps $s$ when expressed by the axioms.
Therefore, as an additional countermeasure:

\textbf{Design Principle 3''} (Diverse Reasoning Rules).
\textit{Samples should also include representative theorems, such as syllogism, contraposition, and De Morgan's laws.}

\vspace{-1mm}
\subsection{Teaching Diverse Linguistic Expressions that Represent Logical Statements} \label{sec:principle_linguistic_diversity}
\vspace{-1mm}

There are various linguistic structures for expressing the logical relationship $\mathcal{F} \rightarrow \mathcal{G}$, such as ``If $\mathcal{F}$ then $\mathcal{G}$'', ``$\mathcal{F}$ leads to $\mathcal{G}$'', and ``$\mathcal{F}$ results in $\mathcal{G}$''.
If we only include specific expressions in the corpora, LLMs may only learn to react to these specific expressions, which has been observed in previous experiments \citep{zhang2022paradox,yuan2023can}.
To prevent this,

\textbf{Design Principle 4} (Diverse Linguistic Expressions).
\textit{Samples should include diverse linguistic expressions that represent logical statements.}

In this chapter, we have established the principles to guide the design of synthetic logic samples.
Next, we construct a synthetic logic corpus based on these principles.

\begin{table*}[t]
    \footnotesize
    \tabcolsep 3pt
    \centering
    \caption{
        Synthetic logic corpora compared in this study, with their features categorized according to our proposed design principles (DP).
        Note that the last row of the \textit{ablation} corpora lists variations of \PLDAbbr, each of which differs from the original regarding one of the design principles.
        \label{tb:corpora}
    }
    \vspace{-1mm}

\resizebox{\textwidth}{!}{

\begin{tabular}{@{}lccccc@{}}
\toprule
\multicolumn{1}{c}{}                                                                                       & DP1                                                                                            & DP2                                                                               & \multicolumn{2}{c}{DP3}                                                                                                                                                                         & DP4                                                                                                    \\
\cmidrule(l){2-2} \cmidrule(l){3-3} \cmidrule(l){4-5} \cmidrule(l){6-6}

                                                                                                           & vocabulary size                                                                                & distractors                                                                       & deduction rules                                                                                              & logical steps                                                                    & expressions per formula                                                                                \\ \midrule
\begin{tabular}[c]{@{}l@{}}RuleTaker {\tiny \citep{clark2020transformers}}\\ \ (RT)\end{tabular}            & \begin{tabular}[c]{@{}c@{}}$\leq$ 100\\ {\scriptsize (hand-selected)}\end{tabular}             & \begin{tabular}[c]{@{}c@{}}random\\ formula\end{tabular}                          & \begin{tabular}[c]{@{}c@{}}2\\ {\scriptsize (implication)}\end{tabular}                                      & 1--5                                                                             & $\mathcal{O}(1)$                                                                                       \\ \midrule
\begin{tabular}[c]{@{}l@{}}PARARULE-Plus {\tiny \citep{bao2022multi}}\\ \ (PRP)\end{tabular}                & \begin{tabular}[c]{@{}c@{}}$\leq$ 100\\ {\scriptsize (hand-selected)}\end{tabular}             & \begin{tabular}[c]{@{}c@{}}random\\ formula\end{tabular}                          & \begin{tabular}[c]{@{}c@{}}2\\ {\scriptsize (implication)}\end{tabular}                                      & 1--5                                                                             & $\mathcal{O}(1)$                                                                                       \\ \midrule
FLD {\tiny \citep{pmlr-v202-morishita23a}}                                                                  & \begin{tabular}[c]{@{}c@{}}$\simeq$ 15k\\ {\scriptsize (WordNet, subset)}\end{tabular}         & \begin{tabular}[c]{@{}c@{}}random\\ formula\end{tabular}                          & \begin{tabular}[c]{@{}c@{}}13\\ {\scriptsize (axioms)}\end{tabular}                                          & \textbf{1--8}                                                                    & 10$\sim$100                                                                                            \\ \midrule
\textbf{\PLDAbbr}                                                                                          & \textbf{\begin{tabular}[c]{@{}c@{}}$\simeq$ 100k\\ {\scriptsize (WordNet, full)}\end{tabular}} & \textbf{\begin{tabular}[c]{@{}c@{}}adversarial\\ formula\end{tabular}}            & \textbf{\begin{tabular}[c]{@{}c@{}}$\simeq$ 50\\ {\scriptsize (axioms and theorems)}\end{tabular}}           & \textbf{1--8}                                                                    & \textbf{\begin{tabular}[c]{@{}c@{}}10$\sim$100\\ {\scriptsize (more extensive than FLD)}\end{tabular}} \\ \midrule
\begin{tabular}[c]{@{}l@{}}\ \ \ \ \PLDAbbr\\ \ \ \ \ \textit{ablation} corpora $\rightarrow$\end{tabular} & \begin{tabular}[c]{@{}c@{}}100\\ $\rightarrow$ \textit{w/o DP1}\end{tabular}                   & \begin{tabular}[c]{@{}c@{}}not used\\ $\rightarrow$ \textit{w/o DP2}\end{tabular} & \begin{tabular}[c]{@{}c@{}}2 {\scriptsize (implication)}\\ $\rightarrow$ \textit{w/o DP3.rules}\end{tabular} & \begin{tabular}[c]{@{}c@{}}1\\ $\rightarrow$ \textit{w/o DP3.steps}\end{tabular} & \begin{tabular}[c]{@{}c@{}}1\\ $\rightarrow$ \textit{w/o DP4}\end{tabular}                             \\ \bottomrule
\end{tabular}

}

\vspace{-2mm}

\end{table*}

\section{Creating a Synthetic Corpus based on Design Principles} \label{sec:PLD}
\vspace{-3mm}

To prepare diverse samples reflecting the design principles 1 to 4 (DP1-4), 
we built a novel sample generator by extending the previous one by \citet{pmlr-v202-morishita23a} and then generated the synthetic logic corpus named \PLDAbbr \ (\PLDItalic).
\Cref{fig:ALT_overview} shows a schematic of our generator and a deduction sample.
\Cref{tb:corpora} compares \PLDAbbr \ with existing corpora.
\Cref{appendix:fig:deduction_example} provides an actual deduction sample included in \PLDAbbr.

More specifically, our generator generates deduction samples through the following steps.
First, the generator randomly generates a sample of multi-step deductive reasoning written in logical formulas, as shown on the left side of \Cref{fig:ALT_overview}, where a conclusion is derived from premises using multiple \textbf{\colorGreenDeductionRule{deduction rules}} (See \Cref{appendix:sec:FLD_proof_tree_generation} for more details of this generation procedure).
At this time, the generator also generates `distractor' logical formulas, which express negative premises of DP2.
Next, the generator converts each logical formula into English expressions.
To achieve this, the generator first randomly selects a template from pre-defined options, such as ``If $\mathcal{F}$, then $\mathcal{G}$,'' ``$\mathcal{F}$ leads to $\mathcal{G}$,'' or ``$\mathcal{F}$ results in $\mathcal{G}$,'' for the logical formula ``$\mathcal{F} \rightarrow \mathcal{G}$.''
It then assigns English content randomly constructed from a vocabulary, such as ``(that) a Foo star exists'' and ``(that) a Bar star exists,'' to each symbol, such as $\mathcal{F}$ and $\mathcal{G}$.
Finally, it converts the multi-step deduction into a deduction sample (right side of \Cref{fig:ALT_overview}) by using the premises as \textbf{\colorBlueFacts{`facts'}}, the conclusion as \textbf{\colorVioletHypothesis{`hypothesis'}}, and the intermediate logical steps as \textbf{\colorRedLogicalSteps{`logical steps'}}.
The deduction sample requires LLMs to generate \textbf{\colorRedLogicalSteps{logical steps}} that derive a given \textbf{\colorVioletHypothesis{hypothesis}} based on the given \textbf{\colorBlueFacts{facts}}.

\Cref{tb:corpora} outlines the comparison of \PLDAbbr \ with other existing corpora \citep{clark2020transformers,bao2022multi,pmlr-v202-morishita23a} in terms of DP1-4, which is detailed as follows:
\begin{itemize}
\vspace{-0.4\baselineskip} %
\setlength{\itemsep}{0.0mm}
\setlength{\leftskip}{-10mm}
    \item DP1: We assign $\mathcal{F}$ and $\mathcal{G}$ content randomly constructed from a vocabulary.
    While the existing corpora used small-sized vocabulary of up to 15k, we use a large vocabulary of around 100k words built from WordNet \citep{miller1995wordnet}.
    This will teach LLMs that $\mathcal{F}$ and $\mathcal{G}$ are truly arbitrary, ultimately enabling them to reason with unknown facts.
    \item DP2: The existing corpora used randomly generated logical formulas as distractors. In contrast, we implement adversarial distractors. For example, for a premise $\mathcal{F} \land \mathcal{G}$, we use $\mathcal{F}$ with missing information (see \Cref{eq:argument:and_modus_ponens,eq:argument:and_modus_ponens_collapsed}), and for a premise $\mathcal{F} \rightarrow \mathcal{H}$, we use $\mathcal{F} \land \mathcal{G} \rightarrow \mathcal{H}$ with missing information as distractors. These distractors teach LLMs precisely when a conclusion can and cannot be derived. As with previous corpora, we include a variable number of distractors in each sample, randomly chosen from a range of 0 to 20.
    \item DP3-3'': While the existing corpora used a small number of deduction rules of up to 13 (refer to Figure B.4 of \citet{pmlr-v202-morishita23a}), we include diverse deduction rules, encompassing the axioms and representative theorems, such as modus ponens, syllogisms, and contraposition, totaling about 50 rules.  We include samples with up to $s=8$ logical steps, following \citep{pmlr-v202-morishita23a}.
    \item DP4: We manually craft several more English templates \textit{per} logical formulas than those used in FLD.
    Since the templates have a nested structure, they yield combinatorially more diverse English expressions.
    While counting the exact number of the resulting expressions is intractable, we observed at least dozens of expressions per logical formula, including minor variations.
    See \Cref{appendix:sec:PLD_linguistic_diversity} for details.
\end{itemize}

\vspace{-2mm}
\section{Experimental Setup} \label{sec:experiments}
\vspace{-3mm}

We briefly explain the experimental settings.
Refer to \Cref{appendix:sec:experiments} for the details.
\vspace{-1mm}

\textbf{Synthetic Logic Corpora:}
We examine the proposed \PLDAbbr \ and previous corpora (\Cref{tb:corpora}).

\textbf{LLMs:}
We used the state-of-the-art LLM, LLaMA-3.1 (8B and 70B) \citep{llama3modelcard}.

\textbf{Training Settings:}
We trained the LLMs by a method similar to supervised fine-tuning; as illustrated in \Cref{fig:ALT_overview}, we used the facts and hypothesis as inputs and logical steps and additional answer label (see \Cref{appendix:sec:answer_label}) as outputs.
We excluded loss computation for the inputs to prevent LLMs from learning to generate unknown facts.
We trained the LLMs for 1 epoch on 100k samples ($\sim$ 0.1B tokens) from the training split of each corpus, with a batch size of 256, resulting in 390 steps, with a linear warmup for 200 steps.
We used the learning rate of 2e-05 for the 8B model and 3e-06 for the 70B model.
We used Huggingface \citep{wolf-et-al-2019-huggingface} for implementation.

\textbf{Prevention of Knowledge Forgetting by Recall Adam Optimizer:}
Synthetic logic corpora include many samples with unknown facts, so training on them should cause LLMs to forget their knowledge of existing facts.
To prevent this, we employed the Recall Adam optimizer \citep{recadam}, which regularizes parameter updates to avoid deviating too far from the pre-training parameters.
Recall Adam stands out for LLM training for several reasons (see \Cref{appendix:sec:experiments_recadam} for details).
We used our re-implemented version \footnote{\scriptsize \url{https://github.com/hitachi-nlp/rec-adam}}.
The hyperparameters were: $\beta_1=0.9, \beta_2=0.999, \epsilon=10^{-6}, \text{fisher coefficient}=4000$ for the 8B model and $2000$ for the 70B model.

\textbf{Benchmarks:}
We evaluated the trained LLMs on \numBenchmarks \ benchmarks shown in \Cref{appendix:tb:benchmarks} using 5-shot in-context learning, except for BBH and AbuductionRules, which used 3-shot in-context learning.
These benchmarks cover a wide range of tasks and are prominent in LLM evaluation.
Note that we excluded the synthetic logic corpora used for training, as training on them often leads to overfitting to their superficial and statistical cues \citep{zhang2022paradox,yuan2023can}, failing to measure truly generalizable reasoning capabilities.
We used lm-evaluation-harness \citep{eval-harness} and bigcode-evaluation-harness \citep{bigcode-evaluation-harness} for the implementation.

\section{Can Additional Logic Training Enhance LLMs' Capabilities?} \label{sec:results_and_discussions_method}
\vspace{-2.5mm}

\Cref{tb:performance_aggregated} show the performance of LLMs before and after \ALT.
Most LLMs trained with \ALT \ outperformed their counterparts without \ALT.
Notably, \ALT \ yielded substantial gains of up to 10 points even for \llamaThreeLargeBaseline, the largest LLM pre-trained on over 15 trillion tokens.
These results verify that \ALT \ can enhance the capabilities of state-of-the-art LLMs.

Among the LLMs trained with \ALT, the one trained on \PLDAbbr \ (i.e., \llamaThreePLDALT) achieved the highest generalization performance across the benchmarks.
\Cref{tb:performance_aggregated_llama_three_ablation} shows the performance of the LLMs trained on \textit{ablated} \PLDAbbr \ corpora, each of which lacks one of the design principles.
As seen, ablating any design principle almost always led to performance degradation.
These results demonstrate that the proposed design principles are critical to obtaining the maximum possible gain from \ALT, and each principle is indispensable.

\Cref{appendix:tb:performance_aggregated} shows that the LLMs trained with \ALT \ without preventing knowledge forgetting by Recall Adam optimizer underperformed compared to their counterparts trained with knowledge forgetting prevention and even the LLM without \ALT.
This behavior presumably occurred because the unknown facts included in synthetic logic corpora displaced the LLM's knowledge of existing facts.
Therefore, knowledge-forgetting prevention is critically important for the success of \ALT.

\begin{table*}[t!]
    \centering
    \tabcolsep 4pt
    \footnotesize
    \caption{
    5-shot performance of LLMs before and after \ALT.
    $\oplus$\textbf{\ALT}-$x$ denotes the LLM trained with \ALT \ on the synthetic logic corpus $x$ from \Cref{tb:corpora}.
    The color shows the rank in each column (darker is better).
    Each benchmark set, such as ``Logic'' and ``Math'', comprises various benchmarks in that domain (see \Cref{appendix:tb:benchmarks}).
    ``Avg.'' represents the micro-average of all the benchmarks.
    \label{tb:performance_aggregated}
    }
    \vspace{-0.5mm}

    \begin{subfigure}{1.0\linewidth}
        \centering
        \subcaption{\llamaThreeBaseline.    \label{tb:performance_aggregated_llama_three}}
        \vspace{-1mm}
        \resizebox{\textwidth}{!}{

\begin{tabular}{lcccccccccccc}
\toprule
{} & Avg. & Logic & Math & Code & NLI & Others & \multicolumn{2}{c}{BBH (3-shot)} & \multicolumn{2}{c}{BBH (0-shot)} & \multicolumn{2}{c}{MMLU} \\
\cmidrule(l){2-2} \cmidrule(l){3-3} \cmidrule(l){4-4} \cmidrule(l){5-5} \cmidrule(l){6-6}  \cmidrule(l){7-7} \cmidrule(l){8-9} \cmidrule(l){10-11} \cmidrule(l){12-13}
{} & {} & {} & {} & {} & {} & {} & {} & CoT & {} & CoT & {} & Pro \\
\midrule

\llamaThreeBaseline & \cellcolor{blue!10} 47.9 & \cellcolor{blue!10} 42.8\scalebox{0.8}{$_{\pm\text{0.4}}$} & \cellcolor{blue!20} 39.6\scalebox{0.8}{$_{\pm\text{0.5}}$} & \cellcolor{blue!10} 35.4 & \cellcolor{blue!10} 65.4\scalebox{0.8}{$_{\pm\text{0.3}}$} & \cellcolor{blue!10} 60.7\scalebox{0.8}{$_{\pm\text{0.3}}$} & \cellcolor{blue!20} 44.9\scalebox{0.8}{$_{\pm\text{0.4}}$} & \cellcolor{blue!20} 61.9\scalebox{0.8}{$_{\pm\text{0.4}}$} & \cellcolor{blue!20} 8.2\scalebox{0.8}{$_{\pm\text{0.2}}$} & \cellcolor{blue!20} 36.5\scalebox{0.8}{$_{\pm\text{0.4}}$} & \cellcolor{blue!30} 65.3\scalebox{0.8}{$_{\pm\text{0.4}}$} & \cellcolor{blue!30} 35.8\scalebox{0.8}{$_{\pm\text{0.4}}$} \\
\llamaThreePRALT & \cellcolor{blue!20} 48.1 & \cellcolor{blue!20} 43.7\scalebox{0.8}{$_{\pm\text{0.2}}$} & \cellcolor{blue!10} 39.2\scalebox{0.8}{$_{\pm\text{0.3}}$} & \cellcolor{blue!20} 35.7 & \cellcolor{blue!20} 65.6\scalebox{0.8}{$_{\pm\text{0.2}}$} & \cellcolor{blue!20} 60.8\scalebox{0.8}{$_{\pm\text{0.2}}$} & \cellcolor{blue!20} 44.9\scalebox{0.8}{$_{\pm\text{0.2}}$} & \cellcolor{blue!10} 61.8\scalebox{0.8}{$_{\pm\text{0.2}}$} & \cellcolor{blue!20} 8.2\scalebox{0.8}{$_{\pm\text{0.1}}$} & \cellcolor{blue!10} 36.4\scalebox{0.8}{$_{\pm\text{0.2}}$} & \cellcolor{blue!30} 65.3\scalebox{0.8}{$_{\pm\text{0.2}}$} & \cellcolor{blue!10} 35.3\scalebox{0.8}{$_{\pm\text{0.2}}$} \\
\llamaThreeRTALT & \cellcolor{blue!30} 50.1 & \cellcolor{blue!30} 46.8\scalebox{0.8}{$_{\pm\text{0.1}}$} & \cellcolor{blue!30} 42.4\scalebox{0.8}{$_{\pm\text{0.2}}$} & \cellcolor{blue!30} 36.5 & \cellcolor{blue!30} 68.6\scalebox{0.8}{$_{\pm\text{0.1}}$} & \cellcolor{blue!30} 61.3\scalebox{0.8}{$_{\pm\text{0.1}}$} & \cellcolor{blue!50} 46.9\scalebox{0.8}{$_{\pm\text{0.2}}$} & \cellcolor{blue!30} 63.5\scalebox{0.8}{$_{\pm\text{0.2}}$} & \cellcolor{blue!50} 13.7\scalebox{0.8}{$_{\pm\text{0.1}}$} & \cellcolor{blue!30} 38.4\scalebox{0.8}{$_{\pm\text{0.2}}$} & \cellcolor{blue!30} 65.3\scalebox{0.8}{$_{\pm\text{0.1}}$} & \cellcolor{blue!20} 35.7\scalebox{0.8}{$_{\pm\text{0.2}}$} \\
\llamaThreeFLDALT & \cellcolor{blue!40} 51.9 & \cellcolor{blue!40} 51.6\scalebox{0.8}{$_{\pm\text{0.1}}$} & \cellcolor{blue!50} 43.4\scalebox{0.8}{$_{\pm\text{0.2}}$} & \cellcolor{blue!50} 38.1 & \cellcolor{blue!40} 70.1\scalebox{0.8}{$_{\pm\text{0.1}}$} & \cellcolor{blue!50} 61.5\scalebox{0.8}{$_{\pm\text{0.1}}$} & \cellcolor{blue!40} 46.7\scalebox{0.8}{$_{\pm\text{0.2}}$} & \cellcolor{blue!40} 64.9\scalebox{0.8}{$_{\pm\text{0.2}}$} & \cellcolor{blue!40} 11.9\scalebox{0.8}{$_{\pm\text{0.1}}$} & \cellcolor{blue!50} 39.6\scalebox{0.8}{$_{\pm\text{0.2}}$} & \cellcolor{blue!40} 65.4\scalebox{0.8}{$_{\pm\text{0.1}}$} & \cellcolor{blue!40} 36.2\scalebox{0.8}{$_{\pm\text{0.2}}$} \\
\llamaThreePLDALTBold & \cellcolor{blue!50} 52.0 & \cellcolor{blue!50} 52.2\scalebox{0.8}{$_{\pm\text{0.1}}$} & \cellcolor{blue!40} 43.2\scalebox{0.8}{$_{\pm\text{0.2}}$} & \cellcolor{blue!40} 38.0 & \cellcolor{blue!50} 70.7\scalebox{0.8}{$_{\pm\text{0.1}}$} & \cellcolor{blue!50} 61.5\scalebox{0.8}{$_{\pm\text{0.1}}$} & \cellcolor{blue!30} 46.5\scalebox{0.8}{$_{\pm\text{0.2}}$} & \cellcolor{blue!50} 65.3\scalebox{0.8}{$_{\pm\text{0.2}}$} & \cellcolor{blue!30} 11.3\scalebox{0.8}{$_{\pm\text{0.1}}$} & \cellcolor{blue!40} 38.7\scalebox{0.8}{$_{\pm\text{0.2}}$} & \cellcolor{blue!50} 65.5\scalebox{0.8}{$_{\pm\text{0.1}}$} & \cellcolor{blue!50} 36.4\scalebox{0.8}{$_{\pm\text{0.2}}$} \\
\bottomrule

\end{tabular}

        }
    \end{subfigure}
    \begin{subfigure}{1.0\linewidth}
        \centering
        \vspace{3mm}
        \subcaption{\llamaThreeLargeBaseline.    \label{tb:performance_aggregated_llama_three_large}}
        \vspace{-1mm}
        \resizebox{\textwidth}{!}{

\begin{tabular}{lcccccccccccc}
\toprule
{} & Avg. & Logic & Math & Code & NLI & Others & \multicolumn{2}{c}{BBH (3-shot)} & \multicolumn{2}{c}{BBH (0-shot)} & \multicolumn{2}{c}{MMLU} \\
\cmidrule(l){2-2} \cmidrule(l){3-3} \cmidrule(l){4-4} \cmidrule(l){5-5} \cmidrule(l){6-6}  \cmidrule(l){7-7} \cmidrule(l){8-9} \cmidrule(l){10-11} \cmidrule(l){12-13}
{} & {} & {} & {} & {} & {} & {} & {} & CoT & {} & CoT & {} & Pro \\
\midrule

LLaMA-3.1-70B & \cellcolor{blue!10} 60.0 & \cellcolor{blue!10} 57.4\scalebox{0.8}{$_{\pm\text{0.4}}$} & \cellcolor{blue!20} 60.0\scalebox{0.8}{$_{\pm\text{0.5}}$} & \cellcolor{blue!10} 46.2 & \cellcolor{blue!20} 73.7\scalebox{0.8}{$_{\pm\text{0.3}}$} & \cellcolor{blue!20} 67.7\scalebox{0.8}{$_{\pm\text{0.3}}$} & \cellcolor{blue!20} 60.4\scalebox{0.8}{$_{\pm\text{0.3}}$} & \cellcolor{blue!10} 82.1\scalebox{0.8}{$_{\pm\text{0.2}}$} & \cellcolor{blue!20} 6.5\scalebox{0.8}{$_{\pm\text{0.1}}$} & \cellcolor{blue!20} 50.1\scalebox{0.8}{$_{\pm\text{0.3}}$} & \cellcolor{blue!20} 78.7\scalebox{0.8}{$_{\pm\text{0.3}}$} & \cellcolor{blue!10} 50.7\scalebox{0.8}{$_{\pm\text{0.4}}$} \\
\llamaThreeLargePRALT & \cellcolor{blue!20} 60.4 & \cellcolor{blue!20} 57.7\scalebox{0.8}{$_{\pm\text{0.4}}$} & \cellcolor{blue!10} 59.8\scalebox{0.8}{$_{\pm\text{0.5}}$} & \cellcolor{blue!20} 49.2 & \cellcolor{blue!10} 73.5\scalebox{0.8}{$_{\pm\text{0.3}}$} & \cellcolor{blue!10} 67.6\scalebox{0.8}{$_{\pm\text{0.3}}$} & \cellcolor{blue!20} 60.4\scalebox{0.8}{$_{\pm\text{0.4}}$} & \cellcolor{blue!20} 82.2\scalebox{0.8}{$_{\pm\text{0.3}}$} & \cellcolor{blue!10} 6.0\scalebox{0.8}{$_{\pm\text{0.2}}$} & \cellcolor{blue!20} 50.1\scalebox{0.8}{$_{\pm\text{0.4}}$} & \cellcolor{blue!20} 78.7\scalebox{0.8}{$_{\pm\text{0.3}}$} & \cellcolor{blue!20} 50.9\scalebox{0.8}{$_{\pm\text{0.4}}$} \\
\llamaThreeLargeRTALT & \cellcolor{blue!30} 62.7 & \cellcolor{blue!30} 61.4\scalebox{0.8}{$_{\pm\text{0.2}}$} & \cellcolor{blue!30} 62.1\scalebox{0.8}{$_{\pm\text{0.3}}$} & \cellcolor{blue!30} 50.8 & \cellcolor{blue!40} 75.4\scalebox{0.8}{$_{\pm\text{0.2}}$} & \cellcolor{blue!30} 68.4\scalebox{0.8}{$_{\pm\text{0.2}}$} & \cellcolor{blue!30} 64.1\scalebox{0.8}{$_{\pm\text{0.3}}$} & \cellcolor{blue!30} 82.5\scalebox{0.8}{$_{\pm\text{0.2}}$} & \cellcolor{blue!40} 11.5\scalebox{0.8}{$_{\pm\text{0.2}}$} & \cellcolor{blue!30} 59.2\scalebox{0.8}{$_{\pm\text{0.3}}$} & \cellcolor{blue!30} 79.0\scalebox{0.8}{$_{\pm\text{0.2}}$} & \cellcolor{blue!30} 52.4\scalebox{0.8}{$_{\pm\text{0.3}}$} \\
\llamaThreeLargeFLDALT & \cellcolor{blue!40} 64.2 & \cellcolor{blue!40} 65.7\scalebox{0.8}{$_{\pm\text{0.1}}$} & \cellcolor{blue!50} 63.6\scalebox{0.8}{$_{\pm\text{0.2}}$} & \cellcolor{blue!40} 52.0 & \cellcolor{blue!30} 75.3\scalebox{0.8}{$_{\pm\text{0.1}}$} & \cellcolor{blue!50} 68.5\scalebox{0.8}{$_{\pm\text{0.1}}$} & \cellcolor{blue!40} 65.0\scalebox{0.8}{$_{\pm\text{0.2}}$} & \cellcolor{blue!50} 83.6\scalebox{0.8}{$_{\pm\text{0.1}}$} & \cellcolor{blue!50} 12.1\scalebox{0.8}{$_{\pm\text{0.1}}$} & \cellcolor{blue!40} 59.9\scalebox{0.8}{$_{\pm\text{0.2}}$} & \cellcolor{blue!40} 79.3\scalebox{0.8}{$_{\pm\text{0.1}}$} & \cellcolor{blue!50} 54.4\scalebox{0.8}{$_{\pm\text{0.2}}$} \\
\llamaThreeLargePLDALTBold & \cellcolor{blue!50} 64.4 & \cellcolor{blue!50} 66.1\scalebox{0.8}{$_{\pm\text{0.1}}$} & \cellcolor{blue!40} 63.3\scalebox{0.8}{$_{\pm\text{0.2}}$} & \cellcolor{blue!50} 52.4 & \cellcolor{blue!50} 76.1\scalebox{0.8}{$_{\pm\text{0.1}}$} & \cellcolor{blue!50} 68.5\scalebox{0.8}{$_{\pm\text{0.1}}$} & \cellcolor{blue!50} 65.4\scalebox{0.8}{$_{\pm\text{0.2}}$} & \cellcolor{blue!50} 83.6\scalebox{0.8}{$_{\pm\text{0.2}}$} & \cellcolor{blue!30} 11.4\scalebox{0.8}{$_{\pm\text{0.1}}$} & \cellcolor{blue!50} 60.8\scalebox{0.8}{$_{\pm\text{0.2}}$} & \cellcolor{blue!50} 79.5\scalebox{0.8}{$_{\pm\text{0.1}}$} & \cellcolor{blue!50} 54.4\scalebox{0.8}{$_{\pm\text{0.2}}$} \\
\bottomrule

\end{tabular}

        }
    \end{subfigure}

\vspace{-7mm}
\end{table*}

\begin{table*}[t!]
    \centering
    \tabcolsep 4pt
    \footnotesize
    \vspace{0mm}
    \caption{
    \llamaThreeBaseline \ trained on the ablation corpora.  \label{tb:performance_aggregated_llama_three_ablation}
    }
    \vspace{-0.5mm}

\resizebox{\textwidth}{!}{

\begin{tabular}{lcccccccccccc}
\toprule
{} & Avg. & Logic & Math & Code & NLI & Others & \multicolumn{2}{c}{BBH (3-shot)} & \multicolumn{2}{c}{BBH (0-shot)} & \multicolumn{2}{c}{MMLU} \\
\cmidrule(l){2-2} \cmidrule(l){3-3} \cmidrule(l){4-4} \cmidrule(l){5-5} \cmidrule(l){6-6}  \cmidrule(l){7-7} \cmidrule(l){8-9} \cmidrule(l){10-11} \cmidrule(l){12-13}
{} & {} & {} & {} & {} & {} & {} & {} & CoT & {} & CoT & {} & Pro \\
\midrule

\llamaThreePLDALTBold & \textbf{52.0} & \textbf{52.2}\scalebox{0.8}{$_{\pm\text{0.1}}$} & \textbf{43.2}\scalebox{0.8}{$_{\pm\text{0.2}}$} & 38.0 & \underline{70.7}\scalebox{0.8}{$_{\pm\text{0.1}}$} & \textbf{61.5}\scalebox{0.8}{$_{\pm\text{0.1}}$} & 46.5\scalebox{0.8}{$_{\pm\text{0.2}}$} & \textbf{65.3}\scalebox{0.8}{$_{\pm\text{0.2}}$} & 11.3\scalebox{0.8}{$_{\pm\text{0.1}}$} & 38.7\scalebox{0.8}{$_{\pm\text{0.2}}$} & \underline{65.5}\scalebox{0.8}{$_{\pm\text{0.1}}$} & \textbf{36.4}\scalebox{0.8}{$_{\pm\text{0.2}}$} \\
\ \ \llamaThreePLDALTWODOne & \underline{51.4} & \textbf{52.2}\scalebox{0.8}{$_{\pm\text{0.1}}$} & \underline{43.1}\scalebox{0.8}{$_{\pm\text{0.2}}$} & \textbf{39.2} & 70.0\scalebox{0.8}{$_{\pm\text{0.1}}$} & 59.4\scalebox{0.8}{$_{\pm\text{0.1}}$} & \underline{46.7}\scalebox{0.8}{$_{\pm\text{0.2}}$} & 64.7\scalebox{0.8}{$_{\pm\text{0.2}}$} & 11.5\scalebox{0.8}{$_{\pm\text{0.1}}$} & \underline{38.9}\scalebox{0.8}{$_{\pm\text{0.2}}$} & 65.4\scalebox{0.8}{$_{\pm\text{0.1}}$} & 36.1\scalebox{0.8}{$_{\pm\text{0.2}}$} \\
\ \ \llamaThreePLDALTWODTwo & 50.6 & 49.9\scalebox{0.8}{$_{\pm\text{0.1}}$} & \underline{43.1}\scalebox{0.8}{$_{\pm\text{0.2}}$} & 38.1 & \textbf{71.1}\scalebox{0.8}{$_{\pm\text{0.1}}$} & 59.3\scalebox{0.8}{$_{\pm\text{0.1}}$} & 46.1\scalebox{0.8}{$_{\pm\text{0.2}}$} & 64.6\scalebox{0.8}{$_{\pm\text{0.2}}$} & 10.4\scalebox{0.8}{$_{\pm\text{0.1}}$} & 37.4\scalebox{0.8}{$_{\pm\text{0.2}}$} & 65.4\scalebox{0.8}{$_{\pm\text{0.1}}$} & 35.7\scalebox{0.8}{$_{\pm\text{0.2}}$} \\
\ \ \llamaThreePLDALTWODThreeRules & 50.7 & 50.4\scalebox{0.8}{$_{\pm\text{0.1}}$} & 42.8\scalebox{0.8}{$_{\pm\text{0.2}}$} & 38.3 & 69.5\scalebox{0.8}{$_{\pm\text{0.1}}$} & 59.4\scalebox{0.8}{$_{\pm\text{0.1}}$} & 46.4\scalebox{0.8}{$_{\pm\text{0.2}}$} & 64.0\scalebox{0.8}{$_{\pm\text{0.2}}$} & 11.8\scalebox{0.8}{$_{\pm\text{0.1}}$} & 38.3\scalebox{0.8}{$_{\pm\text{0.2}}$} & \textbf{65.6}\scalebox{0.8}{$_{\pm\text{0.1}}$} & 36.2\scalebox{0.8}{$_{\pm\text{0.2}}$} \\
\ \ \llamaThreePLDALTWODThreeSteps & 51.1 & \underline{51.5}\scalebox{0.8}{$_{\pm\text{0.1}}$} & \underline{43.1}\scalebox{0.8}{$_{\pm\text{0.2}}$} & \underline{38.7} & 69.6\scalebox{0.8}{$_{\pm\text{0.1}}$} & \underline{59.5}\scalebox{0.8}{$_{\pm\text{0.1}}$} & \textbf{46.8}\scalebox{0.8}{$_{\pm\text{0.2}}$} & \underline{65.0}\scalebox{0.8}{$_{\pm\text{0.2}}$} & \underline{12.3}\scalebox{0.8}{$_{\pm\text{0.1}}$} & 38.8\scalebox{0.8}{$_{\pm\text{0.2}}$} & \textbf{65.6}\scalebox{0.8}{$_{\pm\text{0.1}}$} & \underline{36.3}\scalebox{0.8}{$_{\pm\text{0.2}}$} \\
\ \ \llamaThreePLDALTWODFour & 51.3 & \textbf{52.2}\scalebox{0.8}{$_{\pm\text{0.1}}$} & 42.8\scalebox{0.8}{$_{\pm\text{0.2}}$} & 38.4 & 70.3\scalebox{0.8}{$_{\pm\text{0.1}}$} & \underline{59.5}\scalebox{0.8}{$_{\pm\text{0.1}}$} & 46.1\scalebox{0.8}{$_{\pm\text{0.2}}$} & 64.8\scalebox{0.8}{$_{\pm\text{0.2}}$} & \textbf{12.8}\scalebox{0.8}{$_{\pm\text{0.1}}$} & \textbf{39.3}\scalebox{0.8}{$_{\pm\text{0.2}}$} & \underline{65.5}\scalebox{0.8}{$_{\pm\text{0.1}}$} & \underline{36.3}\scalebox{0.8}{$_{\pm\text{0.2}}$} \\
\bottomrule

\end{tabular}

}

\vspace{-5mm}
\end{table*}

\vspace{-1mm}
\section{What Capabilities Can Additional Logic Training Enhance and Why?}   \label{sec:results_and_discussions_tasks}
\vspace{-2mm}

We analyze the results on each benchmark or each case and discuss whether and why the LLM's capabilities to solve the tasks can or cannot be enhanced by \ALT.

\vspace{-1mm}
\subsection{Logical Reasoning Tasks} \label{sec:which_task_logical_reasoning}
\vspace{-1mm}

\begin{table}[t!]
\centering
\footnotesize
\tabcolsep 1.5pt

\caption{
    Benchmark-wise 5-shot performance of \llamaThreeLargeBaseline \ before and \textbf{after} \ALT\ on \PLDAbbr.
    Refer to \Cref{appendix:tb:performance_details} \ for \llamaThreeBaseline \ results.
    \Cref{appendix:tb:benchmarks} details each benchmark. 
    \label{tb:performance_details}
}

\begin{subfigure}{\linewidth}
    \centering
    \subcaption{Logic.   \label{tb:which_task_logical_reasoning}}
    \vspace{-1mm}
    \tabcolsep 2pt
    
\resizebox{\textwidth}{!}{

\begin{tabular}{lccccccccc}
\toprule
{} & bAbiD & FOLIO & LogicNLI & RobustLR & AR-LSAT & LogiQA & ReClor & AbductionR & ART \\
\midrule
\llamaThreeLargeBaseline & \textbf{83.8}\scalebox{0.8}{$_{\pm\text{1.2}}$} & 58.9\scalebox{0.8}{$_{\pm\text{1.6}}$} & 34.9\scalebox{0.8}{$_{\pm\text{1.1}}$} & 49.6\scalebox{0.8}{$_{\pm\text{0.9}}$} & 21.5\scalebox{0.8}{$_{\pm\text{1.0}}$} & 64.3\scalebox{0.8}{$_{\pm\text{1.2}}$} & 33.7\scalebox{0.8}{$_{\pm\text{0.7}}$} & 84.0\scalebox{0.8}{$_{\pm\text{0.7}}$} & 85.4\scalebox{0.8}{$_{\pm\text{0.9}}$} \\
\llamaThreeLargePLDALTBold & 83.5\scalebox{0.8}{$_{\pm\text{0.5}}$} & \textbf{66.7}\scalebox{0.8}{$_{\pm\text{0.6}}$} & \textbf{50.9}\scalebox{0.8}{$_{\pm\text{0.5}}$} & \textbf{81.6}\scalebox{0.8}{$_{\pm\text{0.3}}$} & \textbf{25.0}\scalebox{0.8}{$_{\pm\text{0.4}}$} & \textbf{69.4}\scalebox{0.8}{$_{\pm\text{0.5}}$} & \textbf{36.3}\scalebox{0.8}{$_{\pm\text{0.3}}$} & \textbf{95.7}\scalebox{0.8}{$_{\pm\text{0.2}}$} & \textbf{85.5}\scalebox{0.8}{$_{\pm\text{0.4}}$} \\
\bottomrule
\end{tabular}

}

\end{subfigure}

\begin{subfigure}{\linewidth}
    \centering
    \subcaption{Math.   \label{tb:which_task_math}}
    \vspace{-1mm}
    \tabcolsep 3pt

\begin{tabular}{lccccc}
\toprule

{} & \multicolumn{3}{c}{GSM8k} & MATH & MathQA \\
\cmidrule(l){2-4} \cmidrule(l){5-5} \cmidrule(l){6-6}
{} & {} & CoT & CoT (0-shot) & - & - \\
\midrule

\llamaThreeLargeBaseline & 80.9\scalebox{0.8}{$_{\pm\text{1.1}}$} & 75.2\scalebox{0.8}{$_{\pm\text{1.2}}$} & 65.4\scalebox{0.8}{$_{\pm\text{1.3}}$} & 23.7\scalebox{0.8}{$_{\pm\text{0.6}}$} & 55.0\scalebox{0.8}{$_{\pm\text{0.9}}$} \\
\llamaThreeLargePLDALTBold & \textbf{83.3}\scalebox{0.8}{$_{\pm\text{0.4}}$} & \textbf{80.4}\scalebox{0.8}{$_{\pm\text{0.4}}$} & \textbf{73.0}\scalebox{0.8}{$_{\pm\text{0.5}}$} & \textbf{24.4}\scalebox{0.8}{$_{\pm\text{0.2}}$} & \textbf{55.4}\scalebox{0.8}{$_{\pm\text{0.4}}$} \\
\bottomrule
\end{tabular}

\end{subfigure}

\begin{subfigure}{\linewidth}
    \centering
    \subcaption{Code.   \label{tb:which_task_coding}}
    \vspace{-1mm}
    \tabcolsep 4pt

\begin{tabular}{lccccc}

\toprule
{} & HumanEval & MBPP & MBPP+ & MultiPL-E (cpp) & MultiPL-E (go) \\
\midrule
\llamaThreeLargeBaseline & 32.3 & 43.4 & 48.7 & 29.8 & 76.6 \\
\llamaThreeLargePLDALTBold & \textbf{42.6} & \textbf{49.5} & \textbf{52.5} & \textbf{38.7} & \textbf{78.6} \\
\bottomrule

\end{tabular}

\end{subfigure}

\begin{subfigure}{\linewidth}
    \centering
    \subcaption{Natural language inference (NLI).   \label{tb:which_task_NLI}}
    \vspace{-1mm}
    \tabcolsep 4pt

\begin{tabular}{lcccc}
\toprule
{} & HELP & MNLI & RTE & SNLI \\
\midrule
\llamaThreeLargeBaseline & 45.8\scalebox{0.8}{$_{\pm\text{0.5}}$} & 82.2\scalebox{0.8}{$_{\pm\text{0.4}}$} & 84.0\scalebox{0.8}{$_{\pm\text{0.7}}$} & \textbf{82.6}\scalebox{0.8}{$_{\pm\text{0.4}}$} \\
\llamaThreeLargePLDALTBold & \textbf{51.3}\scalebox{0.8}{$_{\pm\text{0.2}}$} & \textbf{83.7}\scalebox{0.8}{$_{\pm\text{0.2}}$} & \textbf{87.2}\scalebox{0.8}{$_{\pm\text{0.3}}$} & 82.3\scalebox{0.8}{$_{\pm\text{0.2}}$} \\
\bottomrule
\end{tabular}

\end{subfigure}

\begin{subfigure}{\linewidth}
    \centering
    \subcaption{Others.   \label{tb:which_task_others}}
    \vspace{-1mm}
    \tabcolsep 2pt

\resizebox{\textwidth}{!}{

\begin{tabular}{lccccccccc}
\toprule
{} & CommonsenseQA & HellaSwag & SQuAD & WinoGrande & ARCe & ARCc & GPQA & OpenBookQA & SciQ \\
\midrule
\llamaThreeLargeBaseline & 81.2\scalebox{0.8}{$_{\pm\text{1.1}}$} & 69.2\scalebox{0.8}{$_{\pm\text{0.5}}$} & 38.5\scalebox{0.8}{$_{\pm\text{0.0}}$} & 85.6\scalebox{0.8}{$_{\pm\text{1.0}}$} & 89.1\scalebox{0.8}{$_{\pm\text{0.6}}$} & 65.3\scalebox{0.8}{$_{\pm\text{1.4}}$} & \textbf{40.7}\scalebox{0.8}{$_{\pm\text{1.4}}$} & 41.4\scalebox{0.8}{$_{\pm\text{0.7}}$} & \textbf{98.5}\scalebox{0.8}{$_{\pm\text{0.4}}$} \\
\llamaThreeLargePLDALTBold & \textbf{82.5}\scalebox{0.8}{$_{\pm\text{0.4}}$} & \textbf{69.6}\scalebox{0.8}{$_{\pm\text{0.2}}$} & \textbf{40.1}\scalebox{0.8}{$_{\pm\text{0.0}}$} & \textbf{86.1}\scalebox{0.8}{$_{\pm\text{0.4}}$} & \textbf{89.4}\scalebox{0.8}{$_{\pm\text{0.3}}$} & \textbf{66.7}\scalebox{0.8}{$_{\pm\text{0.6}}$} & 40.6\scalebox{0.8}{$_{\pm\text{0.6}}$} & \textbf{42.8}\scalebox{0.8}{$_{\pm\text{0.3}}$} & \textbf{98.5}\scalebox{0.8}{$_{\pm\text{0.2}}$} \\
\bottomrule
\end{tabular}

}

\end{subfigure}

\end{table}

\Cref{tb:which_task_logical_reasoning} shows that \ALT \ substantially boosted \llamaThreeLargeBaseline's performance by up to 30 points on various benchmarks dealing with logical reasoning tasks.
Surprisingly, we also observed improvements on abductive reasoning tasks, which go beyond the original deductive reasoning tasks in synthetic logic corpora.
Abductive reasoning involves guessing the missing premises that caused the observed conclusion rather than deriving a conclusion from the premises.
For example, from the observed conclusion, ``the window glass at home was broken and the room was ransacked,'' we guess the premise ``a burglar broke in.''
The improvements would be due to the fact that, while the surface form of abductive reasoning problems differs from that of deductive reasoning, they share the fundamentals of logic reflected in the design principles.

Next, we conduct case analyses to see whether the LLM enhanced by \ALT \ acquired the abilities intended by the proposed design principles (DP1-4).
\Cref{tb:case_study} shows problems where \llamaThreeLargeBaseline's errors have been corrected by \ALT.
The first problem is very simple, so it is surprising that \llamaThreeLargeBaseline \ failed to solve it, indicating the inherent difficulty of learning logical reasoning solely from pre-training. 
In contrast, \llamaThreePLDALTWoBold, which was additionally trained on \PLDAbbr, solved the problem correctly.
The premises of the problem are randomly constructed to express unknown facts.
Therefore, the result suggests that \llamaThreePLDALTWoBold \ acquired genuine logical reasoning ability, which can handle unknown facts (DP1).

In the second problem, \llamaThreePLDALTWoBold \ correctly answered ``neutral'', indicating that it successfully learned that conclusions cannot be derived from insufficient facts (DP2).

The third problem comes from the FOLIO benchmark.
To solve this problem, LLMs must use syllogism at the first step as follows: ``All eels are fish, and no fish are plants. Therefore, all ells are not plants.''
\llamaThreePLDALTWoBold \ answered this problem correctly, suggesting that it successfully learned diverse deduction rules (DP3).

FOLIO problems are created based on Wikipedia topics, describing them in more natural and realistic linguistic expressions than in other benchmarks.
As seen in the fourth problem, \llamaThreePLDALTWoBold \ understands such expressions, suggesting the effect of diverse expressions from DP4 and/or that LLMs can integrate their original linguistic ability with the newly acquired logical reasoning ability.

\vspace{-1mm}
\subsection{Math and Coding Tasks} \label{sec:which_task_math_coding}
\vspace{-1mm}

\Cref{tb:which_task_math,tb:which_task_coding} shows that \ALT \ substantially boosted the \llamaThreeLargeBaseline's performance by up to 7 and 10 points on math and coding tasks, respectively.
The math improvements are reasonable, as understanding predicate logic is a prerequisite for solving mathematical problems.
For coding, some recent studies have verified the opposite direction, namely, that training on coding data improves logical reasoning abilities \citep{jiang2024logicproimprovingcomplexlogical,ma2024at,uchiyama2024programminglanguagefeaturespretraining}.

\subsection{NLI Tasks} \label{sec:which_task_NLI}
\vspace{-3mm}

\Cref{tb:which_task_NLI} shows that \ALT \ substantially boosted the \llamaThreeLargeBaseline's performance by up to 6 points on various natural language inference (NLI) benchmarks.
NLI is similar to deductive reasoning in assessing whether a premise supports or contradicts a hypothesis.
However, the main difference is that this judgment requires a rich set of commonsense knowledge beyond the given premise.

Consider the fifth problem in \Cref{tb:case_study}: by supplementing the given fact ``An Indian woman is dancing with her partner'' with the commonsense knowledge ``If someone is dancing, then he/she is moving.'', we can derive the hypothesis ``A woman is moving.''
The sixth problem is more challenging as we have to trace multiple logical steps while supplementing with sufficient commonsense knowledge as follows: ``a church choir sings at a church,'' ``baseball is often played at a baseball field,'' ``a person cannot be in two or more places at the same time,'' ``therefore, a church choir cannot sing for baseball.''

Since synthetic logic corpora only contain unknown facts, LLMs cannot acquire new knowledge from them.
Therefore, the commonsense knowledge used to solve the above problems must have been acquired by the LLMs from pre-training.
This suggests that LLMs can integrate their original knowledge with the logical reasoning capabilities newly acquired from \ALT \ to solve problems.

\vspace{-2mm}
\subsection{Other Tasks} \label{sec:which_task_others}
\vspace{-3mm}

\begin{table*}[t]
\centering
\tabcolsep 2.0pt

\caption{
    Problems where \llamaThreeLargeBaseline \ initially answered incorrectly and then correctly after training with \ALT \ on \PLDAbbr.
    \red{Red} highlights the premises related to the hypothesis.
    \label{tb:case_study}
}
\vspace{-2mm}

\resizebox{\textwidth}{!}{

\begin{tabular}{@{}cllcc@{}}
\toprule
benchmark                 & premises                                                                                                                                                                                                                                                                                                                                                                                                                                  & hypothesis                                                                                  & \begin{tabular}[c]{@{}c@{}}answer\\ (\llamaThreeLargeBaseline/gold)\end{tabular} & \begin{tabular}[c]{@{}c@{}}required\\ ability\end{tabular}                                          \\ \midrule
\multirow{2}{*}{LogicNLI} & \begin{tabular}[c]{@{}l@{}}Mice are afraid of wolves. \red{Cats are afraid of sheep.}\\ \red{Jessica is a cat.} Wolves are afraid of cats.\\ Winona is a wolf. Sheep are afraid of cats.\end{tabular}                                                                                                                                                                                                                                     & \begin{tabular}[c]{@{}l@{}}Jessica is\\ afraid of sheep.\end{tabular}                       & \begin{tabular}[c]{@{}c@{}}neutral/\\ entailment\end{tabular}                    & DP1                                                                                                 \\ \cmidrule(l){2-5} 
                          & \begin{tabular}[c]{@{}l@{}}Rhett is not modest.  Vivian is confused.\\ Rhett is lazy.  If someone is modest or not confused, then he is not eager.\end{tabular}                                                                                                                                                                                                                                                                           & \begin{tabular}[c]{@{}l@{}}Rhett is\\ confused.\end{tabular}                                & \begin{tabular}[c]{@{}c@{}}entailment/\\ neutral\end{tabular}                    & DP2                                                                                                 \\ \midrule
\multirow{2}{*}{FOLIO}    & \begin{tabular}[c]{@{}l@{}}\red{All eels are fish.} \red{No fish are plants.} \\ \red{Everything displayed in the collection is either a plant or an animal.}\\ \red{All animals displayed in the collection are multicellular.}\\ \red{A sea eel is displayed in the collection.}\\ The sea eel is an eel or an animal or not a plant.\end{tabular}                                                                                      & \begin{tabular}[c]{@{}l@{}}The sea eel\\ is multicellular\\ or is bacteria.\end{tabular}    & \begin{tabular}[c]{@{}c@{}}neutral/\\ entailment\end{tabular}                    & DP3                                                                                                 \\ \cmidrule(l){2-5} 
                          & \begin{tabular}[c]{@{}l@{}}\red{Common utilities include water, electricity, gas, heating, sewer, trash, and recycling.}\\ Many apartment rents cover the cost of water and electricity.\\ Susan lives in an apartment where the rent covers all utilities.\\ \red{The rent of the apartment where Ava lives does not cover any utility expenses.}\\ \red{Noah lives in an apartment where the rent does not cover heating.}\end{tabular} & \begin{tabular}[c]{@{}l@{}}Noah and Ava both\\ need to pay\\ the heating bill.\end{tabular} & \begin{tabular}[c]{@{}c@{}}neutral/\\ entailment\end{tabular}                    & DP4                                                                                                 \\ \midrule
\multirow{2}{*}{SNLI}     & An Indian woman \red{is dancing} with her partner.                                                                                                                                                                                                                                                                                                                                                                                        & A woman is moving.                                                                          & \begin{tabular}[c]{@{}c@{}}neutral/\\ entailment\end{tabular}                    & \multirow{2}{*}{\begin{tabular}[c]{@{}c@{}}reasoning\\ with\\ commonsense\\ knowledge\end{tabular}} \\ \cmidrule(lr){2-4}
                          & \begin{tabular}[c]{@{}l@{}}\red{This church choir sings} to the masses\\ as they sing joyous songs from the book \red{at a church.}\end{tabular}                                                                                                                                                                                                                                                                                          & \begin{tabular}[c]{@{}l@{}}A choir is singing\\ at a baseball game.\end{tabular}            & \begin{tabular}[c]{@{}c@{}}entailment/\\ contradiction\end{tabular}              &                                                                                                     \\ \bottomrule
\end{tabular}

}

\vspace{-4mm}
\end{table*}

\begin{table*}[t]
\centering
\tabcolsep 3.0pt

\vspace{1mm}
\caption{
    Problems that \llamaThreeLargeBaseline \ trained with \ALT \ on \PLDAbbr \ still cannot solve.
    \label{tb:case_study_negatives}
}
\vspace{-1mm}

\resizebox{\textwidth}{!}{

\begin{tabular}{@{}cll@{}}
\toprule
\multicolumn{1}{l}{benchmark}                             & question                                                                                                                                                                                                                                                                                                                                                                    & answer                                                                                                                               \\ \midrule
\begin{tabular}[c]{@{}c@{}}ARC\\ (challenge)\end{tabular} & \begin{tabular}[c]{@{}l@{}}The end result in the process of photosynthesis is the production of\\ sugar and oxygen. Which step signals the beginning of photosynthesis?\end{tabular}                                                                                                                                                                                        & \begin{tabular}[c]{@{}l@{}}Chlorophyll in the\\ leaf captures light energy.\end{tabular}                                             \\ \midrule
GPQA                                                      & \begin{tabular}[c]{@{}l@{}}A spin-half particle is in a linear superposition $0.8|\uparrow\rangle+0.6|\downarrow\rangle$ of its spin-up\\ and spin-down states. If $|\uparrow\rangle$ and $|\downarrow \rangle$ are the eigenstates of $\sigma_{z}$, then what\\ is the expectation value up to one decimal place, of the operator $10\sigma_{z}+5\sigma_{x}$?\end{tabular} & $-0.7$                                                                                                                               \\ \midrule
\begin{tabular}[c]{@{}c@{}}ARC\\ (challenge)\end{tabular} & \begin{tabular}[c]{@{}l@{}}Beavers build their homes in ponds and streams. Which characteristic\\ is least critical to building homes in an aquatic environment?\end{tabular}                                                                                                                                                                                               & \begin{tabular}[c]{@{}l@{}}(A) waterproof fur (B) webbed hind feet\\ \textbf{(C) arge, sharp teeth} (D) flat, wide tail\end{tabular} \\ \bottomrule
\end{tabular}

}

\vspace{-4mm}
\end{table*}

Improvements across various other tasks (\Cref{tb:which_task_others}) demonstrate the broad benefits of the obtained reasoning capabilities beyond standard reasoning tasks; though the improvements were modest at up to 2 percentage points, which may be due to the following reasons.
First, these benchmarks include problems that purely test knowledge, such as the first one in \Cref{tb:case_study_negatives}.
Since \ALT \ does not aim to provide new knowledge, the ability to solve such problems does not improve by nature.
Next, some problems may require knowledge that is too advanced for LLMs, so potential improvements by the enhanced reasoning capabilities may be bottlenecked.
For example, the second problem does involve reasoning but requires sufficient quantum mechanics knowledge as a prerequisite.
However, these knowledge-related issues should be solved by improving the quantity and quality of pre-training.

Finally, LLMs may not be able to fully utilize the potential of enhanced reasoning capabilities for problems that require complex procedures.
To solve the third problem, LLMs first must attempt reasoning related to each choice as follows: ``To build homes in an aquatic environment, one needs to maintain body heat and insulation despite being frequently submerged in cold water. Therefore, the waterproof fur of (A) is essential'',
and ``To build \dots, one must gather and process natural materials like wood. Large, sharp teeth of (C) are critical as they allow beavers to cut down trees and shape branches.''
Next, while reasoning traces on (A) to (D) all seem reasonable, LLMs must choose the single best answer, considering the subtle nuance of the question context, as follows: ``Since the question emphasizes the aquatic environment, the least related reasoning trace should be (C).''
This complex procedure contrasts with logical reasoning and NLI problems, where LLMs can directly obtain an answer from a single reasoning trace.
Previous studies also observed that such procedure on multiple-choice QA problems are challenging for LLMs \citep{joshua2023mcq,chuijie2024mcq,wang2024answers}.
Since \ALT \ alone does not teach LLMs such task-specific procedures, additional training on these procedures should be necessary to solve these problems.

\vspace{-4mm}
\section{Conclusion}    \label{sec:conclusion}
\vspace{-4mm}
Towards versatile artificial intelligence with reasoning capabilities, we proposed \textbf{A}dditional \textit{\textbf{L}ogic} \textbf{T}raining on synthetic logic samples.
We established systematic design principles well-grounded on symbolic logic theory and previous empirical findings.
We constructed a corpus named \PLD\ (\PLDAbbr) based on the design principles.
We empirically showed that \ALT \ on \PLDAbbr \ substantially enhances the capabilities of state-of-the-art LLMs.

\clearpage

\section*{Acknowledgement}
Computational resources of AI Bridging Cloud Infrastructure (ABCI) provided by the National Institute of Advanced Industrial Science and Technology (AIST) were used.
We thank Dr. Masaaki Shimizu at Hitachi for the convenience of additional computational resources.
We thank Dr. Naoaki Okazaki, a professor at the Tokyo Institute of Technology, for the keen comments.

\bibliography{neurips_2024}

\begin{thebibliography}{135}
\expandafter\ifx\csname natexlab\endcsname\relax\def\natexlab#1{#1}\fi

\bibitem[{AI@Meta(2024)}]{llama3modelcard}
AI@Meta. 2024.
\newblock \href {https://github.com/meta-llama/llama3/blob/main/MODEL_CARD.md} {Llama 3 model card}.

\bibitem[{Amini et~al.(2019)Amini, Gabriel, Lin, Koncel-Kedziorski, Choi, and Hajishirzi}]{amini-etal-2019-mathqa}
Aida Amini, Saadia Gabriel, Shanchuan Lin, Rik Koncel-Kedziorski, Yejin Choi, and Hannaneh Hajishirzi. 2019.
\newblock \href {https://doi.org/10.18653/v1/N19-1245} {{M}ath{QA}: Towards interpretable math word problem solving with operation-based formalisms}.
\newblock In \emph{Proceedings of the 2019 Conference of the North {A}merican Chapter of the Association for Computational Linguistics: Human Language Technologies, Volume 1 (Long and Short Papers)}, pages 2357--2367, Minneapolis, Minnesota. Association for Computational Linguistics.

\bibitem[{Ando et~al.(2023)Ando, Morishita, Abe, Mineshima, and Okada}]{ando-etal-2023-evaluating}
Risako Ando, Takanobu Morishita, Hirohiko Abe, Koji Mineshima, and Mitsuhiro Okada. 2023.
\newblock \href {https://aclanthology.org/2023.naloma-1.1} {Evaluating large language models with {N}eu{BAROCO}: Syllogistic reasoning ability and human-like biases}.
\newblock In \emph{Proceedings of the 4th Natural Logic Meets Machine Learning Workshop}, pages 1--11, Nancy, France. Association for Computational Linguistics.

\bibitem[{Aoki et~al.(2024)Aoki, Kudo, Kuribayashi, Sone, Taniguchi, Sakaguchi, and Inui}]{aoki2024heuristicrationaldynamicuse}
Yoichi Aoki, Keito Kudo, Tatsuki Kuribayashi, Shusaku Sone, Masaya Taniguchi, Keisuke Sakaguchi, and Kentaro Inui. 2024.
\newblock \href {http://arxiv.org/abs/2406.16078} {First heuristic then rational: Dynamic use of heuristics in language model reasoning}.

\bibitem[{Askell(2020)}]{askell2020gpt}
Amanda Askell. 2020.
\newblock Gpt-3: Towards renaissance models.
\newblock \emph{Daily Nous Blog: Philosophers On GPT-3}.

\bibitem[{Austin et~al.(2021)Austin, Odena, Nye, Bosma, Michalewski, Dohan, Jiang, Cai, Terry, Le et~al.}]{austin2021program}
Jacob Austin, Augustus Odena, Maxwell Nye, Maarten Bosma, Henryk Michalewski, David Dohan, Ellen Jiang, Carrie Cai, Michael Terry, Quoc Le, et~al. 2021.
\newblock Program synthesis with large language models.
\newblock \emph{arXiv preprint arXiv:2108.07732}.

\bibitem[{Bao et~al.(2022)Bao, Peng, Hartill, Tan, Deng, Witbrock, and Liu}]{bao2022multi}
Qiming Bao, Alex~Yuxuan Peng, Tim Hartill, Neset Tan, Zhenyun Deng, Michael Witbrock, and Jiamou Liu. 2022.
\newblock Multi-step deductive reasoning over natural language: An empirical study on out-of-distribution generalisation.
\newblock In \emph{Proceedings of the 16th International Workshop on Neural-Symbolic Learning and Reasoning as part of the 2nd International Joint Conference on Learning \& Reasoning (IJCLR 2022)}, pages 202--217, Cumberland Lodge, Windsor Great Park, United Kingdom.

\bibitem[{Ben~Allal et~al.(2024)Ben~Allal, Lozhkov, Penedo, Wolf, and von Werra}]{benallal2024cosmopedia}
Loubna Ben~Allal, Anton Lozhkov, Guilherme Penedo, Thomas Wolf, and Leandro von Werra. 2024.
\newblock \href {https://huggingface.co/datasets/HuggingFaceTB/cosmopedia} {Cosmopedia}.

\bibitem[{Ben~Allal et~al.(2022)Ben~Allal, Muennighoff, Kumar~Umapathi, Lipkin, and von Werra}]{bigcode-evaluation-harness}
Loubna Ben~Allal, Niklas Muennighoff, Logesh Kumar~Umapathi, Ben Lipkin, and Leandro von Werra. 2022.
\newblock A framework for the evaluation of code generation models.
\newblock \url{https://github.com/bigcode-project/bigcode-evaluation-harness}.

\bibitem[{Bentivogli et~al.(2009)Bentivogli, Dagan, Dang, Giampiccolo, and Magnini}]{bentivogli2009fifth}
Luisa Bentivogli, Ido Dagan, Hoa~Trang Dang, Danilo Giampiccolo, and Bernardo Magnini. 2009.
\newblock The fifth pascal recognizing textual entailment challenge.
\newblock In \emph{Text Analysis Conference}.

\bibitem[{Bertolazzi et~al.(2024)Bertolazzi, Gatt, and Bernardi}]{bertolazzi2024systematicanalysislargelanguage}
Leonardo Bertolazzi, Albert Gatt, and Raffaella Bernardi. 2024.
\newblock \href {http://arxiv.org/abs/2406.11341} {A systematic analysis of large language models as soft reasoners: The case of syllogistic inferences}.

\bibitem[{Bertrand()}]{russel1946}
Russell Bertrand.
\newblock A history of western philosophy.

\bibitem[{Betz et~al.(2021)Betz, Voigt, and Richardson}]{betz-etal-2021-critical}
Gregor Betz, Christian Voigt, and Kyle Richardson. 2021.
\newblock \href {https://aclanthology.org/2021.iwcs-1.7} {Critical thinking for language models}.
\newblock In \emph{Proceedings of the 14th International Conference on Computational Semantics (IWCS)}, pages 63--75, Groningen, The Netherlands (online). Association for Computational Linguistics.

\bibitem[{Bhagavatula et~al.(2019)Bhagavatula, Bras, Malaviya, Sakaguchi, Holtzman, Rashkin, Downey, Yih, and Choi}]{bhagavatula2019abductive}
Chandra Bhagavatula, Ronan~Le Bras, Chaitanya Malaviya, Keisuke Sakaguchi, Ari Holtzman, Hannah Rashkin, Doug Downey, Scott Wen-tau Yih, and Yejin Choi. 2019.
\newblock Abductive commonsense reasoning.
\newblock \emph{arXiv preprint arXiv:1908.05739}.

\bibitem[{Bhuiya et~al.(2024)Bhuiya, Schlegel, and Winkler}]{bhuiya2024seeminglyplausibledistractorsmultihop}
Neeladri Bhuiya, Viktor Schlegel, and Stefan Winkler. 2024.
\newblock \href {http://arxiv.org/abs/2409.05197} {Seemingly plausible distractors in multi-hop reasoning: Are large language models attentive readers?}

\bibitem[{Bostrom et~al.(2021)Bostrom, Zhao, Chaudhuri, and Durrett}]{bostrom-etal-2021-flexible}
Kaj Bostrom, Xinyu Zhao, Swarat Chaudhuri, and Greg Durrett. 2021.
\newblock \href {https://doi.org/10.18653/v1/2021.emnlp-main.506} {Flexible generation of natural language deductions}.
\newblock In \emph{Proceedings of the 2021 Conference on Empirical Methods in Natural Language Processing}, pages 6266--6278, Online and Punta Cana, Dominican Republic. Association for Computational Linguistics.

\bibitem[{Bowman et~al.()Bowman, Angeli, Potts, and Manning}]{bowmanlarge}
Samuel~R Bowman, Gabor Angeli, Christopher Potts, and Christopher~D Manning.
\newblock A large annotated corpus for learning natural language inference.

\bibitem[{Cassano et~al.(2023)Cassano, Gouwar, Nguyen, Nguyen, Phipps-Costin, Pinckney, Yee, Zi, Anderson, Feldman, Guha, Greenberg, and Jangda}]{multiple}
Federico Cassano, John Gouwar, Daniel Nguyen, Sydney Nguyen, Luna Phipps-Costin, Donald Pinckney, Ming-Ho Yee, Yangtian Zi, Carolyn~Jane Anderson, Molly~Q Feldman, Arjun Guha, Michael Greenberg, and Abhinav Jangda. 2023.
\newblock \href {https://doi.org/10.1109/TSE.2023.3267446} {Multipl-e: A scalable and polyglot approach to benchmarking neural code generation}.
\newblock \emph{IEEE Transactions on Software Engineering}, 49(7):3675--3691.

\bibitem[{Chen et~al.(2021)Chen, Tworek, Jun, Yuan, de~Oliveira~Pinto, Kaplan, Edwards, Burda, Joseph, Brockman, Ray, Puri, Krueger, Petrov, Khlaaf, Sastry, Mishkin, Chan, Gray, Ryder, Pavlov, Power, Kaiser, Bavarian, Winter, Tillet, Such, Cummings, Plappert, Chantzis, Barnes, Herbert-Voss, Guss, Nichol, Paino, Tezak, Tang, Babuschkin, Balaji, Jain, Saunders, Hesse, Carr, Leike, Achiam, Misra, Morikawa, Radford, Knight, Brundage, Murati, Mayer, Welinder, McGrew, Amodei, McCandlish, Sutskever, and Zaremba}]{chen2021codex}
Mark Chen, Jerry Tworek, Heewoo Jun, Qiming Yuan, Henrique~Ponde de~Oliveira~Pinto, Jared Kaplan, Harri Edwards, Yuri Burda, Nicholas Joseph, Greg Brockman, Alex Ray, Raul Puri, Gretchen Krueger, Michael Petrov, Heidy Khlaaf, Girish Sastry, Pamela Mishkin, Brooke Chan, Scott Gray, Nick Ryder, Mikhail Pavlov, Alethea Power, Lukasz Kaiser, Mohammad Bavarian, Clemens Winter, Philippe Tillet, Felipe~Petroski Such, Dave Cummings, Matthias Plappert, Fotios Chantzis, Elizabeth Barnes, Ariel Herbert-Voss, William~Hebgen Guss, Alex Nichol, Alex Paino, Nikolas Tezak, Jie Tang, Igor Babuschkin, Suchir Balaji, Shantanu Jain, William Saunders, Christopher Hesse, Andrew~N. Carr, Jan Leike, Josh Achiam, Vedant Misra, Evan Morikawa, Alec Radford, Matthew Knight, Miles Brundage, Mira Murati, Katie Mayer, Peter Welinder, Bob McGrew, Dario Amodei, Sam McCandlish, Ilya Sutskever, and Wojciech Zaremba. 2021.
\newblock \href {http://arxiv.org/abs/2107.03374} {Evaluating large language models trained on code}.

\bibitem[{Chen et~al.(2020)Chen, Hou, Cui, Che, Liu, and Yu}]{recadam}
Sanyuan Chen, Yutai Hou, Yiming Cui, Wanxiang Che, Ting Liu, and Xiangzhan Yu. 2020.
\newblock \href {https://www.aclweb.org/anthology/2020.emnlp-main.634} {Recall and learn: Fine-tuning deep pretrained language models with less forgetting}.
\newblock In \emph{Proceedings of the 2020 Conference on Empirical Methods in Natural Language Processing (EMNLP)}, pages 7870--7881, Online. Association for Computational Linguistics.

\bibitem[{Chen et~al.(2024)Chen, Chi, Wang, and Zhou}]{pmlr-v235-chen24i}
Xinyun Chen, Ryan~Andrew Chi, Xuezhi Wang, and Denny Zhou. 2024.
\newblock \href {https://proceedings.mlr.press/v235/chen24i.html} {Premise order matters in reasoning with large language models}.
\newblock In \emph{Proceedings of the 41st International Conference on Machine Learning}, volume 235 of \emph{Proceedings of Machine Learning Research}, pages 6596--6620. PMLR.

\bibitem[{Cheng et~al.(2017)Cheng, Bernstein, Danescu-Niculescu-Mizil, and Leskovec}]{Cheng:2017ud}
J.~Cheng, M.~Bernstein, C.~Danescu-Niculescu-Mizil, and J.~Leskovec. 2017.
\newblock \href {https://doi.org/10.1145/2998181.2998213} {Anyone can become a troll: Causes of trolling behavior in online discussions}.
\newblock \emph{CSCW: Proceedings of the Conference on Computer-Supported Cooperative Work. Conference on Computer-Supported Cooperative Work, 2017}.

\bibitem[{Clark et~al.(2018)Clark, Cowhey, Etzioni, Khot, Sabharwal, Schoenick, and Tafjord}]{clark2018think}
Peter Clark, Isaac Cowhey, Oren Etzioni, Tushar Khot, Ashish Sabharwal, Carissa Schoenick, and Oyvind Tafjord. 2018.
\newblock Think you have solved question answering? try arc, the ai2 reasoning challenge.
\newblock \emph{arXiv preprint arXiv:1803.05457}.

\bibitem[{Clark et~al.(2021)Clark, Tafjord, and Richardson}]{clark2020transformers}
Peter Clark, Oyvind Tafjord, and Kyle Richardson. 2021.
\newblock Transformers as soft reasoners over language.
\newblock In \emph{Proceedings of the Twenty-Ninth International Conference on International Joint Conferences on Artificial Intelligence}, pages 3882--3890.

\bibitem[{Cobbe et~al.(2021)Cobbe, Kosaraju, Bavarian, Chen, Jun, Kaiser, Plappert, Tworek, Hilton, Nakano, Hesse, and Schulman}]{cobbe2021gsm8k}
Karl Cobbe, Vineet Kosaraju, Mohammad Bavarian, Mark Chen, Heewoo Jun, Lukasz Kaiser, Matthias Plappert, Jerry Tworek, Jacob Hilton, Reiichiro Nakano, Christopher Hesse, and John Schulman. 2021.
\newblock Training verifiers to solve math word problems.
\newblock \emph{arXiv preprint arXiv:2110.14168}.

\bibitem[{Colmerauer and Roussel(1973)}]{colmerauer1973prolog}
A.~Colmerauer and P~Roussel. 1973.
\newblock The birth of prolog.
\newblock \emph{The ALP Newsletter}.

\bibitem[{Dagan et~al.(2005)Dagan, Glickman, and Magnini}]{dagan2005pascal}
Ido Dagan, Oren Glickman, and Bernardo Magnini. 2005.
\newblock The {PASCAL} recognising textual entailment challenge.
\newblock pages 177--190.

\bibitem[{Dalvi et~al.(2021)Dalvi, Jansen, Tafjord, Xie, Smith, Pipatanangkura, and Clark}]{dalvi-etal-2021-explaining}
Bhavana Dalvi, Peter Jansen, Oyvind Tafjord, Zhengnan Xie, Hannah Smith, Leighanna Pipatanangkura, and Peter Clark. 2021.
\newblock \href {https://doi.org/10.18653/v1/2021.emnlp-main.585} {Explaining answers with entailment trees}.
\newblock In \emph{Proceedings of the 2021 Conference on Empirical Methods in Natural Language Processing}, pages 7358--7370, Online and Punta Cana, Dominican Republic. Association for Computational Linguistics.

\bibitem[{Dasgupta et~al.(2023)Dasgupta, Lampinen, Chan, Sheahan, Creswell, Kumaran, McClelland, and Hill}]{dasgupta2023language}
Ishita Dasgupta, Andrew~K. Lampinen, Stephanie C.~Y. Chan, Hannah~R. Sheahan, Antonia Creswell, Dharshan Kumaran, James~L. McClelland, and Felix Hill. 2023.
\newblock \href {http://arxiv.org/abs/2207.07051} {Language models show human-like content effects on reasoning tasks}.

\bibitem[{Dougrez-Lewis et~al.(2024)Dougrez-Lewis, Akhter, He, and Liakata}]{dougrezlewis2024assessingreasoningabilitieschatgpt}
John Dougrez-Lewis, Mahmud~Elahi Akhter, Yulan He, and Maria Liakata. 2024.
\newblock \href {http://arxiv.org/abs/2402.10735} {Assessing the reasoning abilities of chatgpt in the context of claim verification}.

\bibitem[{Dziri et~al.(2023)Dziri, Lu, Sclar, Li, Jiang, Lin, West, Bhagavatula, Bras, Hwang, Sanyal, Welleck, Ren, Ettinger, Harchaoui, and Choi}]{dziri2023faith}
Nouha Dziri, Ximing Lu, Melanie Sclar, Xiang~Lorraine Li, Liwei Jiang, Bill~Yuchen Lin, Peter West, Chandra Bhagavatula, Ronan~Le Bras, Jena~D. Hwang, Soumya Sanyal, Sean Welleck, Xiang Ren, Allyson Ettinger, Zaid Harchaoui, and Yejin Choi. 2023.
\newblock \href {http://arxiv.org/abs/2305.18654} {Faith and fate: Limits of transformers on compositionality}.

\bibitem[{Eisape et~al.(2024)Eisape, Tessler, Dasgupta, Sha, Steenkiste, and Linzen}]{eisape-etal-2024-systematic}
Tiwalayo Eisape, Michael Tessler, Ishita Dasgupta, Fei Sha, Sjoerd Steenkiste, and Tal Linzen. 2024.
\newblock \href {https://doi.org/10.18653/v1/2024.naacl-long.466} {A systematic comparison of syllogistic reasoning in humans and language models}.
\newblock In \emph{Proceedings of the 2024 Conference of the North American Chapter of the Association for Computational Linguistics: Human Language Technologies (Volume 1: Long Papers)}, pages 8425--8444, Mexico City, Mexico. Association for Computational Linguistics.

\bibitem[{Elkan and Greiner(1993)}]{elkan1993building}
Charles Elkan and Russell Greiner. 1993.
\newblock Building large knowledge-based systems: Representation and inference in the cyc project: Db lenat and rv guha.

\bibitem[{Gao et~al.(2023)Gao, Tow, Abbasi, Biderman, Black, DiPofi, Foster, Golding, Hsu, Le~Noac'h, Li, McDonell, Muennighoff, Ociepa, Phang, Reynolds, Schoelkopf, Skowron, Sutawika, Tang, Thite, Wang, Wang, and Zou}]{eval-harness}
Leo Gao, Jonathan Tow, Baber Abbasi, Stella Biderman, Sid Black, Anthony DiPofi, Charles Foster, Laurence Golding, Jeffrey Hsu, Alain Le~Noac'h, Haonan Li, Kyle McDonell, Niklas Muennighoff, Chris Ociepa, Jason Phang, Laria Reynolds, Hailey Schoelkopf, Aviya Skowron, Lintang Sutawika, Eric Tang, Anish Thite, Ben Wang, Kevin Wang, and Andy Zou. 2023.
\newblock \href {https://doi.org/10.5281/zenodo.10256836} {A framework for few-shot language model evaluation}.

\bibitem[{Giampiccolo et~al.(2007)Giampiccolo, Magnini, Dagan, and Dolan}]{giampiccolo2007third}
Danilo Giampiccolo, Bernardo Magnini, Ido Dagan, and William~B Dolan. 2007.
\newblock The third {PASCAL} recognizing textual entailment challenge.
\newblock In \emph{ACL-PASCAL Workshop on Textual Entailment and Paraphrasing}, pages 1--9.

\bibitem[{G\"{o}del(1930)}]{godel1930uber}
Kurt G\"{o}del. 1930.
\newblock \emph{Uber die vollst{\"a}ndigkeit des logikkalk{\"u}ls}.
\newblock Ph.D. thesis, Ph. D. dissertation, University of Vienna.

\bibitem[{Gontier et~al.(2020)Gontier, Sinha, Reddy, and Pal}]{gontier2020measuring}
Nicolas Gontier, Koustuv Sinha, Siva Reddy, and Chris Pal. 2020.
\newblock Measuring systematic generalization in neural proof generation with transformers.
\newblock \emph{Advances in Neural Information Processing Systems}, 33:22231--22242.

\bibitem[{Goodman(1954)}]{goodman1954fact}
Nelson Goodman. 1954.
\newblock Fact, fiction, and forecast. london: University of london.

\bibitem[{Guia{\c{s}}u and Tindale(2018)}]{guiacsu2018logical}
Radu~Cornel Guia{\c{s}}u and Christopher~W Tindale. 2018.
\newblock Logical fallacies and invasion biology.
\newblock \emph{Biology \& philosophy}, 33(5-6):34.

\bibitem[{Habernal et~al.(2018)Habernal, Wachsmuth, Gurevych, and Stein}]{habernal-etal-2018-argument}
Ivan Habernal, Henning Wachsmuth, Iryna Gurevych, and Benno Stein. 2018.
\newblock \href {https://doi.org/10.18653/v1/N18-1175} {The argument reasoning comprehension task: Identification and reconstruction of implicit warrants}.
\newblock In \emph{Proceedings of the 2018 Conference of the North {A}merican Chapter of the Association for Computational Linguistics: Human Language Technologies, Volume 1 (Long Papers)}, pages 1930--1940, New Orleans, Louisiana. Association for Computational Linguistics.

\bibitem[{Han et~al.(2024)Han, Song, Yu, and You}]{han2024incontextlearningelicittrustworthy}
Pengrui Han, Peiyang Song, Haofei Yu, and Jiaxuan You. 2024.
\newblock \href {http://arxiv.org/abs/2409.15454} {In-context learning may not elicit trustworthy reasoning: A-not-b errors in pretrained language models}.

\bibitem[{Han et~al.(2022)Han, Schoelkopf, Zhao, Qi, Riddell, Benson, Sun, Zubova, Qiao, Burtell et~al.}]{han2022folio}
Simeng Han, Hailey Schoelkopf, Yilun Zhao, Zhenting Qi, Martin Riddell, Luke Benson, Lucy Sun, Ekaterina Zubova, Yujie Qiao, Matthew Burtell, et~al. 2022.
\newblock Folio: Natural language reasoning with first-order logic.
\newblock \emph{arXiv e-prints}, pages arXiv--2209.

\bibitem[{Hansson(2004)}]{hansson2004fallacies}
Sven~Ove Hansson. 2004.
\newblock Fallacies of risk.
\newblock \emph{Journal of Risk Research}, 7(3):353--360.

\bibitem[{Hendrycks et~al.(2021{\natexlab{a}})Hendrycks, Burns, Basart, Zou, Mazeika, Song, and Steinhardt}]{hendryckstest2021}
Dan Hendrycks, Collin Burns, Steven Basart, Andy Zou, Mantas Mazeika, Dawn Song, and Jacob Steinhardt. 2021{\natexlab{a}}.
\newblock Measuring massive multitask language understanding.
\newblock \emph{Proceedings of the International Conference on Learning Representations (ICLR)}.

\bibitem[{Hendrycks et~al.(2021{\natexlab{b}})Hendrycks, Burns, Kadavath, Arora, Basart, Tang, Song, and Steinhardt}]{hendrycksmath2021}
Dan Hendrycks, Collin Burns, Saurav Kadavath, Akul Arora, Steven Basart, Eric Tang, Dawn Song, and Jacob Steinhardt. 2021{\natexlab{b}}.
\newblock Measuring mathematical problem solving with the math dataset.
\newblock \emph{NeurIPS}.

\bibitem[{Ho et~al.(2023)Ho, Schmid, and Yun}]{ho2023large}
Namgyu Ho, Laura Schmid, and Se-Young Yun. 2023.
\newblock \href {http://arxiv.org/abs/2212.10071} {Large language models are reasoning teachers}.

\bibitem[{Hodel and West(2023)}]{hodel2023response}
Damian Hodel and Jevin West. 2023.
\newblock \href {http://arxiv.org/abs/2308.16118} {Response: Emergent analogical reasoning in large language models}.

\bibitem[{Hong et~al.(2024)Hong, Zhang, Pang, Yu, and Zhang}]{hong-etal-2024-closer}
Ruixin Hong, Hongming Zhang, Xinyu Pang, Dong Yu, and Changshui Zhang. 2024.
\newblock \href {https://doi.org/10.18653/v1/2024.naacl-long.52} {A closer look at the self-verification abilities of large language models in logical reasoning}.
\newblock In \emph{Proceedings of the 2024 Conference of the North American Chapter of the Association for Computational Linguistics: Human Language Technologies (Volume 1: Long Papers)}, pages 900--925, Mexico City, Mexico. Association for Computational Linguistics.

\bibitem[{Hu et~al.(2024)Hu, Gao, Gao, Chen, and Huang}]{hu2024largelanguagemodelslimited}
Peng Hu, Changjiang Gao, Ruiqi Gao, Jiajun Chen, and Shujian Huang. 2024.
\newblock \href {http://arxiv.org/abs/2406.07393} {Large language models are limited in out-of-context knowledge reasoning}.

\bibitem[{Huang et~al.(2024)Huang, Chen, Mishra, Zheng, Yu, Song, and Zhou}]{huang2024large}
Jie Huang, Xinyun Chen, Swaroop Mishra, Huaixiu~Steven Zheng, Adams~Wei Yu, Xinying Song, and Denny Zhou. 2024.
\newblock \href {https://openreview.net/forum?id=IkmD3fKBPQ} {Large language models cannot self-correct reasoning yet}.
\newblock In \emph{The Twelfth International Conference on Learning Representations}.

\bibitem[{Hume(1748)}]{hume1748enquiry}
David Hume. 1748.
\newblock An enquiry concerning human understanding (section iv).
\newblock \emph{Recuperado de http://www. clorenzano. com. ar}.

\bibitem[{Jiang et~al.(2024{\natexlab{a}})Jiang, Xie, Hao, Wang, Mallick, Su, Taylor, and Roth}]{jiang2024peektokenbiaslarge}
Bowen Jiang, Yangxinyu Xie, Zhuoqun Hao, Xiaomeng Wang, Tanwi Mallick, Weijie~J. Su, Camillo~J. Taylor, and Dan Roth. 2024{\natexlab{a}}.
\newblock \href {http://arxiv.org/abs/2406.11050} {A peek into token bias: Large language models are not yet genuine reasoners}.

\bibitem[{Jiang et~al.(2024{\natexlab{b}})Jiang, Yan, Liu, Jin, Peng, Zhang, Cai, Cao, Gao, and Tang}]{jiang2024logicproimprovingcomplexlogical}
Jin Jiang, Yuchen Yan, Yang Liu, Yonggang Jin, Shuai Peng, Mengdi Zhang, Xunliang Cai, Yixin Cao, Liangcai Gao, and Zhi Tang. 2024{\natexlab{b}}.
\newblock \href {http://arxiv.org/abs/2409.12929} {Logicpro: Improving complex logical reasoning via program-guided learning}.

\bibitem[{Kahneman(2011)}]{kahneman2011thinking}
Daniel Kahneman. 2011.
\newblock \emph{Thinking, fast and slow}.
\newblock Macmillan.

\bibitem[{Lanham et~al.(2023)Lanham, Chen, Radhakrishnan, Steiner, Denison, Hernandez, Li, Durmus, Hubinger, Kernion, Lukošiūtė, Nguyen, Cheng, Joseph, Schiefer, Rausch, Larson, McCandlish, Kundu, Kadavath, Yang, Henighan, Maxwell, Telleen-Lawton, Hume, Hatfield-Dodds, Kaplan, Brauner, Bowman, and Perez}]{lanham2023measuring}
Tamera Lanham, Anna Chen, Ansh Radhakrishnan, Benoit Steiner, Carson Denison, Danny Hernandez, Dustin Li, Esin Durmus, Evan Hubinger, Jackson Kernion, Kamilė Lukošiūtė, Karina Nguyen, Newton Cheng, Nicholas Joseph, Nicholas Schiefer, Oliver Rausch, Robin Larson, Sam McCandlish, Sandipan Kundu, Saurav Kadavath, Shannon Yang, Thomas Henighan, Timothy Maxwell, Timothy Telleen-Lawton, Tristan Hume, Zac Hatfield-Dodds, Jared Kaplan, Jan Brauner, Samuel~R. Bowman, and Ethan Perez. 2023.
\newblock \href {http://arxiv.org/abs/2307.13702} {Measuring faithfulness in chain-of-thought reasoning}.

\bibitem[{Li et~al.(2023)Li, Hessel, Yu, Ren, Chang, and Choi}]{li-etal-2023-symbolic}
Liunian~Harold Li, Jack Hessel, Youngjae Yu, Xiang Ren, Kai-Wei Chang, and Yejin Choi. 2023.
\newblock \href {https://doi.org/10.18653/v1/2023.acl-long.150} {Symbolic chain-of-thought distillation: Small models can also {``}think{''} step-by-step}.
\newblock In \emph{Proceedings of the 61st Annual Meeting of the Association for Computational Linguistics (Volume 1: Long Papers)}, pages 2665--2679, Toronto, Canada. Association for Computational Linguistics.

\bibitem[{Li et~al.(2022)Li, Chen, Shen, Chen, Zhang, Li, Wang, Qian, Peng, Mao, Chen, and Yan}]{li2022explanations}
Shiyang Li, Jianshu Chen, Yelong Shen, Zhiyu Chen, Xinlu Zhang, Zekun Li, Hong Wang, Jing Qian, Baolin Peng, Yi~Mao, Wenhu Chen, and Xifeng Yan. 2022.
\newblock \href {http://arxiv.org/abs/2210.06726} {Explanations from large language models make small reasoners better}.

\bibitem[{Liu et~al.(2023{\natexlab{a}})Liu, Liu, Cui, Teng, Duan, Zhou, and Zhang}]{logiqa2}
Hanmeng Liu, Jian Liu, Leyang Cui, Zhiyang Teng, Nan Duan, Ming Zhou, and Yue Zhang. 2023{\natexlab{a}}.
\newblock \href {https://doi.org/10.1109/TASLP.2023.3293046} {Logiqa 2.0—an improved dataset for logical reasoning in natural language understanding}.
\newblock \emph{IEEE/ACM Transactions on Audio, Speech, and Language Processing}, 31:2947--2962.

\bibitem[{Liu et~al.(2023{\natexlab{b}})Liu, Ning, Teng, Liu, Zhou, and Zhang}]{liu2023evaluating}
Hanmeng Liu, Ruoxi Ning, Zhiyang Teng, Jian Liu, Qiji Zhou, and Yue Zhang. 2023{\natexlab{b}}.
\newblock \href {http://arxiv.org/abs/2304.03439} {Evaluating the logical reasoning ability of chatgpt and gpt-4}.

\bibitem[{Liu et~al.(2023{\natexlab{c}})Liu, Teng, Cui, Zhang, Zhou, and Zhang}]{liu-etal-2023-logicot}
Hanmeng Liu, Zhiyang Teng, Leyang Cui, Chaoli Zhang, Qiji Zhou, and Yue Zhang. 2023{\natexlab{c}}.
\newblock \href {https://doi.org/10.18653/v1/2023.findings-emnlp.191} {{L}ogi{C}o{T}: Logical chain-of-thought instruction tuning}.
\newblock In \emph{Findings of the Association for Computational Linguistics: EMNLP 2023}, pages 2908--2921, Singapore. Association for Computational Linguistics.

\bibitem[{Liu et~al.(2020)Liu, Cui, Liu, Huang, Wang, and Zhang}]{ijcai2020p501}
Jian Liu, Leyang Cui, Hanmeng Liu, Dandan Huang, Yile Wang, and Yue Zhang. 2020.
\newblock \href {https://doi.org/10.24963/ijcai.2020/501} {Logiqa: A challenge dataset for machine reading comprehension with logical reasoning}.
\newblock In \emph{Proceedings of the Twenty-Ninth International Joint Conference on Artificial Intelligence, {IJCAI-20}}, pages 3622--3628. International Joint Conferences on Artificial Intelligence Organization.
\newblock Main track.

\bibitem[{Liu et~al.(2023{\natexlab{d}})Liu, Xia, Wang, and Zhang}]{evalplus}
Jiawei Liu, Chunqiu~Steven Xia, Yuyao Wang, and Lingming Zhang. 2023{\natexlab{d}}.
\newblock \href {https://openreview.net/forum?id=1qvx610Cu7} {Is your code generated by chat{GPT} really correct? rigorous evaluation of large language models for code generation}.
\newblock In \emph{Thirty-seventh Conference on Neural Information Processing Systems}.

\bibitem[{Liu et~al.(2019)Liu, Ott, Goyal, Du, Joshi, Chen, Levy, Lewis, Zettlemoyer, and Stoyanov}]{liu-et-al-2019-roberta}
Yinhan Liu, Myle Ott, Naman Goyal, Jingfei Du, Mandar Joshi, Danqi Chen, Omer Levy, Mike Lewis, Luke Zettlemoyer, and Veselin Stoyanov. 2019.
\newblock {RoBERTa}: A robustly optimized {BERT} pretraining approach.
\newblock \emph{arXiv preprint arXiv:1907.11692}.

\bibitem[{Liu et~al.(2024)Liu, Lee, Du, Sanyal, and Zhao}]{liu2024selfcontradictoryreasoningevaluationdetection}
Ziyi Liu, Isabelle Lee, Yongkang Du, Soumya Sanyal, and Jieyu Zhao. 2024.
\newblock \href {http://arxiv.org/abs/2311.09603} {Self-contradictory reasoning evaluation and detection}.

\bibitem[{Lu et~al.(2024)Lu, Zhou, Ren, Wang, Shi, Pan, Zhan, and Li}]{lu2024mathgenie}
Zimu Lu, Aojun Zhou, Houxing Ren, Ke~Wang, Weikang Shi, Junting Pan, Mingjie Zhan, and Hongsheng Li. 2024.
\newblock \href {http://arxiv.org/abs/2402.16352} {Mathgenie: Generating synthetic data with question back-translation for enhancing mathematical reasoning of llms}.

\bibitem[{MA et~al.(2024)MA, Liu, Yu, Zhang, Jiang, Wang, and Li}]{ma2024at}
YINGWEI MA, Yue Liu, Yue Yu, Yuanliang Zhang, Yu~Jiang, Changjian Wang, and Shanshan Li. 2024.
\newblock \href {https://openreview.net/forum?id=KIPJKST4gw} {At which training stage does code data help {LLM}s reasoning?}
\newblock In \emph{The Twelfth International Conference on Learning Representations}.

\bibitem[{Magister et~al.(2023)Magister, Mallinson, Adamek, Malmi, and Severyn}]{magister-etal-2023-teaching}
Lucie~Charlotte Magister, Jonathan Mallinson, Jakub Adamek, Eric Malmi, and Aliaksei Severyn. 2023.
\newblock \href {https://doi.org/10.18653/v1/2023.acl-short.151} {Teaching small language models to reason}.
\newblock In \emph{Proceedings of the 61st Annual Meeting of the Association for Computational Linguistics (Volume 2: Short Papers)}, pages 1773--1781, Toronto, Canada. Association for Computational Linguistics.

\bibitem[{McCarthy(1959)}]{Mccarthy1959ProgramsWC}
John~W. McCarthy. 1959.
\newblock Programs with common sense.
\newblock In \emph{Proc. Tedding Conf. on the Mechanization of Thought Processes}, pages 75--91.

\bibitem[{Mihaylov et~al.(2018)Mihaylov, Clark, Khot, and Sabharwal}]{OpenBookQA2018}
Todor Mihaylov, Peter Clark, Tushar Khot, and Ashish Sabharwal. 2018.
\newblock Can a suit of armor conduct electricity? a new dataset for open book question answering.
\newblock In \emph{EMNLP}.

\bibitem[{Miller(1995)}]{miller1995wordnet}
George~A Miller. 1995.
\newblock Wordnet: a lexical database for english.
\newblock \emph{Communications of the ACM}, 38(11):39--41.

\bibitem[{Mirzadeh et~al.(2024)Mirzadeh, Alizadeh, Shahrokhi, Tuzel, Bengio, and Farajtabar}]{mirzadeh2024gsmsymbolicunderstandinglimitationsmathematical}
Iman Mirzadeh, Keivan Alizadeh, Hooman Shahrokhi, Oncel Tuzel, Samy Bengio, and Mehrdad Farajtabar. 2024.
\newblock \href {http://arxiv.org/abs/2410.05229} {Gsm-symbolic: Understanding the limitations of mathematical reasoning in large language models}.

\bibitem[{Mitchell(2023)}]{melanie2023blog}
Melanie Mitchell. 2023.
\newblock Can large language models reason?
\newblock \emph{blog}, pages https://aiguide.substack.com/p/can--large--language--models--reason.

\bibitem[{Mitra et~al.(2023)Mitra, Corro, Mahajan, Codas, Simoes, Agarwal, Chen, Razdaibiedina, Jones, Aggarwal, Palangi, Zheng, Rosset, Khanpour, and Awadallah}]{mitra2023orca}
Arindam Mitra, Luciano~Del Corro, Shweti Mahajan, Andres Codas, Clarisse Simoes, Sahaj Agarwal, Xuxi Chen, Anastasia Razdaibiedina, Erik Jones, Kriti Aggarwal, Hamid Palangi, Guoqing Zheng, Corby Rosset, Hamed Khanpour, and Ahmed Awadallah. 2023.
\newblock \href {http://arxiv.org/abs/2311.11045} {Orca 2: Teaching small language models how to reason}.

\bibitem[{Mondorf and Plank(2024)}]{mondorf2024liarliarlogicalmire}
Philipp Mondorf and Barbara Plank. 2024.
\newblock \href {http://arxiv.org/abs/2406.12546} {Liar, liar, logical mire: A benchmark for suppositional reasoning in large language models}.

\bibitem[{Morishita et~al.(2023)Morishita, Morio, Yamaguchi, and Sogawa}]{pmlr-v202-morishita23a}
Terufumi Morishita, Gaku Morio, Atsuki Yamaguchi, and Yasuhiro Sogawa. 2023.
\newblock \href {https://proceedings.mlr.press/v202/morishita23a.html} {Learning deductive reasoning from synthetic corpus based on formal logic}.
\newblock In \emph{Proceedings of the 40th International Conference on Machine Learning}, volume 202 of \emph{Proceedings of Machine Learning Research}, pages 25254--25274. PMLR.

\bibitem[{Morishita et~al.(2024)Morishita, Yamaguchi, Morio, Tomonari, Imaichi, and Sogawa}]{morishita-etal-2024-jfld}
Terufumi Morishita, Atsuki Yamaguchi, Gaku Morio, Hikaru Tomonari, Osamu Imaichi, and Yasuhiro Sogawa. 2024.
\newblock \href {https://aclanthology.org/2024.lrec-main.832} {{JFLD}: A {J}apanese benchmark for deductive reasoning based on formal logic}.
\newblock In \emph{Proceedings of the 2024 Joint International Conference on Computational Linguistics, Language Resources and Evaluation (LREC-COLING 2024)}, pages 9526--9535, Torino, Italia. ELRA and ICCL.

\bibitem[{Nafar et~al.(2024)Nafar, Venable, and Kordjamshidi}]{nafar-etal-2024-teaching}
Aliakbar Nafar, K.~Brent Venable, and Parisa Kordjamshidi. 2024.
\newblock \href {https://aclanthology.org/2024.findings-eacl.112} {Teaching probabilistic logical reasoning to transformers}.
\newblock In \emph{Findings of the Association for Computational Linguistics: EACL 2024}, pages 1615--1632, St. Julian{'}s, Malta. Association for Computational Linguistics.

\bibitem[{Ozeki et~al.(2024)Ozeki, Ando, Morishita, Abe, Mineshima, and Okada}]{ozeki-etal-2024-exploring}
Kentaro Ozeki, Risako Ando, Takanobu Morishita, Hirohiko Abe, Koji Mineshima, and Mitsuhiro Okada. 2024.
\newblock \href {https://doi.org/10.18653/v1/2024.findings-acl.950} {Exploring reasoning biases in large language models through syllogism: Insights from the {N}eu{BAROCO} dataset}.
\newblock In \emph{Findings of the Association for Computational Linguistics ACL 2024}, pages 16063--16077, Bangkok, Thailand and virtual meeting. Association for Computational Linguistics.

\bibitem[{Paglieri(2017)}]{Paglieri2017}
Fabio Paglieri. 2017.
\newblock \href {https://doi.org/10.1007/s13347-016-0222-6} {A plea for ecological argument technologies}.
\newblock \emph{Philosophy {\&} Technology}, 30(2):209--238.

\bibitem[{Parmar et~al.(2024)Parmar, Patel, Varshney, Nakamura, Luo, Mashetty, Mitra, and Baral}]{parmar-etal-2024-logicbench}
Mihir Parmar, Nisarg Patel, Neeraj Varshney, Mutsumi Nakamura, Man Luo, Santosh Mashetty, Arindam Mitra, and Chitta Baral. 2024.
\newblock \href {https://doi.org/10.18653/v1/2024.acl-long.739} {{L}ogic{B}ench: Towards systematic evaluation of logical reasoning ability of large language models}.
\newblock In \emph{Proceedings of the 62nd Annual Meeting of the Association for Computational Linguistics (Volume 1: Long Papers)}, pages 13679--13707, Bangkok, Thailand. Association for Computational Linguistics.

\bibitem[{Patel et~al.(2024)Patel, Kulkarni, Parmar, Budhiraja, Nakamura, Varshney, and Baral}]{patel2024multilogievalevaluatingmultisteplogical}
Nisarg Patel, Mohith Kulkarni, Mihir Parmar, Aashna Budhiraja, Mutsumi Nakamura, Neeraj Varshney, and Chitta Baral. 2024.
\newblock \href {http://arxiv.org/abs/2406.17169} {Multi-logieval: Towards evaluating multi-step logical reasoning ability of large language models}.

\bibitem[{Pi et~al.(2022)Pi, Liu, Chen, Ziyadi, Lin, Fu, Gao, Lou, and Chen}]{pi-etal-2022-reasoning}
Xinyu Pi, Qian Liu, Bei Chen, Morteza Ziyadi, Zeqi Lin, Qiang Fu, Yan Gao, Jian-Guang Lou, and Weizhu Chen. 2022.
\newblock \href {https://doi.org/10.18653/v1/2022.emnlp-main.48} {Reasoning like program executors}.
\newblock In \emph{Proceedings of the 2022 Conference on Empirical Methods in Natural Language Processing}, pages 761--779, Abu Dhabi, United Arab Emirates. Association for Computational Linguistics.

\bibitem[{Quine(1969)}]{quine1969epistemology}
Willard Van~Orman Quine. 1969.
\newblock Epistemology naturalized. ontological relativity and other essays.
\newblock \emph{New York: Columbia UP}.

\bibitem[{Radford et~al.(2019)Radford, Wu, Child, Luan, Amodei, and Sutskever}]{radford2019language}
Alec Radford, Jeff Wu, Rewon Child, David Luan, Dario Amodei, and Ilya Sutskever. 2019.
\newblock Language models are unsupervised multitask learners.

\bibitem[{Rae et~al.(2021)Rae, Borgeaud, Cai, Millican, Hoffmann, Song, Aslanides, Henderson, Ring, Young et~al.}]{rae2021scaling}
Jack~W Rae, Sebastian Borgeaud, Trevor Cai, Katie Millican, Jordan Hoffmann, Francis Song, John Aslanides, Sarah Henderson, Roman Ring, Susannah Young, et~al. 2021.
\newblock Scaling language models: Methods, analysis \& insights from training gopher.
\newblock \emph{arXiv preprint arXiv:2112.11446}.

\bibitem[{Rajpurkar et~al.(2018)Rajpurkar, Jia, and Liang}]{rajpurkar-etal-2018-know}
Pranav Rajpurkar, Robin Jia, and Percy Liang. 2018.
\newblock \href {https://doi.org/10.18653/v1/P18-2124} {Know what you don{'}t know: Unanswerable questions for {SQ}u{AD}}.
\newblock In \emph{Proceedings of the 56th Annual Meeting of the Association for Computational Linguistics (Volume 2: Short Papers)}, pages 784--789, Melbourne, Australia. Association for Computational Linguistics.

\bibitem[{Razeghi et~al.(2022)Razeghi, Logan~IV, Gardner, and Singh}]{razeghi2022impact}
Yasaman Razeghi, Robert~L Logan~IV, Matt Gardner, and Sameer Singh. 2022.
\newblock Impact of pretraining term frequencies on few-shot numerical reasoning.
\newblock In \emph{Findings of the Association for Computational Linguistics: EMNLP 2022}, pages 840--854.

\bibitem[{Rein et~al.(2023)Rein, Hou, Stickland, Petty, Pang, Dirani, Michael, and Bowman}]{rein2023gpqa}
David Rein, Betty~Li Hou, Asa~Cooper Stickland, Jackson Petty, Richard~Yuanzhe Pang, Julien Dirani, Julian Michael, and Samuel~R Bowman. 2023.
\newblock Gpqa: A graduate-level google-proof q\&a benchmark.
\newblock \emph{arXiv preprint arXiv:2311.12022}.

\bibitem[{Robinson and Wingate(2023)}]{joshua2023mcq}
Joshua Robinson and David Wingate. 2023.
\newblock \href {https://openreview.net/forum?id=yKbprarjc5B} {Leveraging large language models for multiple choice question answering}.
\newblock In \emph{The Eleventh International Conference on Learning Representations}.

\bibitem[{Saeed et~al.(2021)Saeed, Ahmadi, Nakov, and Papotti}]{saeed-etal-2021-rulebert}
Mohammed Saeed, Naser Ahmadi, Preslav Nakov, and Paolo Papotti. 2021.
\newblock \href {https://doi.org/10.18653/v1/2021.emnlp-main.110} {{R}ule{BERT}: Teaching soft rules to pre-trained language models}.
\newblock In \emph{Proceedings of the 2021 Conference on Empirical Methods in Natural Language Processing}, pages 1460--1476, Online and Punta Cana, Dominican Republic. Association for Computational Linguistics.

\bibitem[{Saha et~al.(2020)Saha, Ghosh, Srivastava, and Bansal}]{saha-etal-2020-prover}
Swarnadeep Saha, Sayan Ghosh, Shashank Srivastava, and Mohit Bansal. 2020.
\newblock \href {https://doi.org/10.18653/v1/2020.emnlp-main.9} {{PR}over: Proof generation for interpretable reasoning over rules}.
\newblock In \emph{Proceedings of the 2020 Conference on Empirical Methods in Natural Language Processing (EMNLP)}, pages 122--136, Online. Association for Computational Linguistics.

\bibitem[{Sakaguchi et~al.(2021)Sakaguchi, Bras, Bhagavatula, and Choi}]{sakaguchi2021winogrande}
Keisuke Sakaguchi, Ronan~Le Bras, Chandra Bhagavatula, and Yejin Choi. 2021.
\newblock Winogrande: An adversarial winograd schema challenge at scale.
\newblock \emph{Communications of the ACM}, 64(9):99--106.

\bibitem[{Sanyal et~al.(2022{\natexlab{a}})Sanyal, Liao, and Ren}]{sanyal2022robustlr}
Soumya Sanyal, Zeyi Liao, and Xiang Ren. 2022{\natexlab{a}}.
\newblock Robustlr: Evaluating robustness to logical perturbation in deductive reasoning.
\newblock \emph{arXiv preprint arXiv:2205.12598}.

\bibitem[{Sanyal et~al.(2022{\natexlab{b}})Sanyal, Singh, and Ren}]{sanyal2022fairr}
Soumya Sanyal, Harman Singh, and Xiang Ren. 2022{\natexlab{b}}.
\newblock Fairr: Faithful and robust deductive reasoning over natural language.
\newblock In \emph{Proceedings of the 60th Annual Meeting of the Association for Computational Linguistics (Volume 1: Long Papers)}, pages 1075--1093.

\bibitem[{Shortliffe(1976)}]{shortliffe1976computer}
eh~Shortliffe. 1976.
\newblock Computer based medical consultations: Mycin.
\newblock \emph{Elsevier}.

\bibitem[{Shridhar et~al.(2023)Shridhar, Stolfo, and Sachan}]{shridhar-etal-2023-distilling}
Kumar Shridhar, Alessandro Stolfo, and Mrinmaya Sachan. 2023.
\newblock \href {https://doi.org/10.18653/v1/2023.findings-acl.441} {Distilling reasoning capabilities into smaller language models}.
\newblock In \emph{Findings of the Association for Computational Linguistics: ACL 2023}, pages 7059--7073, Toronto, Canada. Association for Computational Linguistics.

\bibitem[{Sileo(2024)}]{sileo2024scalingsyntheticlogicalreasoning}
Damien Sileo. 2024.
\newblock \href {http://arxiv.org/abs/2406.11035} {Scaling synthetic logical reasoning datasets with context-sensitive declarative grammars}.

\bibitem[{Sprague et~al.(2024)Sprague, Ye, Bostrom, Chaudhuri, and Durrett}]{sprague2024musr}
Zayne~Rea Sprague, Xi~Ye, Kaj Bostrom, Swarat Chaudhuri, and Greg Durrett. 2024.
\newblock \href {https://openreview.net/forum?id=jenyYQzue1} {Mu{SR}: Testing the limits of chain-of-thought with multistep soft reasoning}.
\newblock In \emph{The Twelfth International Conference on Learning Representations}.

\bibitem[{Sunstein and Hastie(2015)}]{SunsteinHastie2015}
Cass~R Sunstein and Reid Hastie. 2015.
\newblock \emph{Wiser: getting beyond groupthink to make groups smarter}.
\newblock Harvard Business Review Press, Boston.

\bibitem[{Suzgun et~al.(2022)Suzgun, Scales, Sch{\"a}rli, Gehrmann, Tay, Chung, Chowdhery, Le, Chi, Zhou, , and Wei}]{suzgun2022challenging}
Mirac Suzgun, Nathan Scales, Nathanael Sch{\"a}rli, Sebastian Gehrmann, Yi~Tay, Hyung~Won Chung, Aakanksha Chowdhery, Quoc~V Le, Ed~H Chi, Denny Zhou, , and Jason Wei. 2022.
\newblock Challenging big-bench tasks and whether chain-of-thought can solve them.
\newblock \emph{arXiv preprint arXiv:2210.09261}.

\bibitem[{Tafjord et~al.(2021)Tafjord, Dalvi, and Clark}]{tafjord-etal-2021-proofwriter}
Oyvind Tafjord, Bhavana Dalvi, and Peter Clark. 2021.
\newblock \href {https://doi.org/10.18653/v1/2021.findings-acl.317} {{P}roof{W}riter: Generating implications, proofs, and abductive statements over natural language}.
\newblock In \emph{Findings of the Association for Computational Linguistics: ACL-IJCNLP 2021}, pages 3621--3634, Online. Association for Computational Linguistics.

\bibitem[{Talmor et~al.(2018)Talmor, Herzig, Lourie, and Berant}]{talmor2018commonsenseqa}
Alon Talmor, Jonathan Herzig, Nicholas Lourie, and Jonathan Berant. 2018.
\newblock Commonsenseqa: A question answering challenge targeting commonsense knowledge.
\newblock \emph{arXiv preprint arXiv:1811.00937}.

\bibitem[{Tian et~al.(2021)Tian, Li, Chen, Xiao, He, and Jin}]{tian2021diagnosing}
Jidong Tian, Yitian Li, Wenqing Chen, Liqiang Xiao, Hao He, and Yaohui Jin. 2021.
\newblock Diagnosing the first-order logical reasoning ability through logicnli.
\newblock In \emph{Proceedings of the 2021 Conference on Empirical Methods in Natural Language Processing}, pages 3738--3747.

\bibitem[{Trinh et~al.(2024)Trinh, Wu, Le, He, and Luong}]{trinh2024solving}
Trieu~H Trinh, Yuhuai Wu, Quoc~V Le, He~He, and Thang Luong. 2024.
\newblock Solving olympiad geometry without human demonstrations.
\newblock \emph{Nature}, 625(7995):476--482.

\bibitem[{Turpin et~al.(2023)Turpin, Michael, Perez, and Bowman}]{turpin2023language}
Miles Turpin, Julian Michael, Ethan Perez, and Samuel~R. Bowman. 2023.
\newblock \href {http://arxiv.org/abs/2305.04388} {Language models don't always say what they think: Unfaithful explanations in chain-of-thought prompting}.

\bibitem[{Uchiyama et~al.(2024)Uchiyama, Kojima, Gambardella, Cao, Iwasawa, and Matsuo}]{uchiyama2024programminglanguagefeaturespretraining}
Fumiya Uchiyama, Takeshi Kojima, Andrew Gambardella, Qi~Cao, Yusuke Iwasawa, and Yutaka Matsuo. 2024.
\newblock \href {http://arxiv.org/abs/2410.06735} {Which programming language and what features at pre-training stage affect downstream logical inference performance?}

\bibitem[{Vaswani et~al.(2017)Vaswani, Shazeer, Parmar, Uszkoreit, Jones, Gomez, Kaiser, and Polosukhin}]{vaswani2017attention}
Ashish Vaswani, Noam Shazeer, Niki Parmar, Jakob Uszkoreit, Llion Jones, Aidan~N Gomez, \L~ukasz Kaiser, and Illia Polosukhin. 2017.
\newblock Attention is all you need.
\newblock In \emph{Advances in Neural Information Processing Systems}, volume~30.

\bibitem[{Wan et~al.(2024)Wan, Wang, Yang, Yuan, tse Huang, He, Jiao, and Lyu}]{wan2024logicaskerevaluatingimprovinglogical}
Yuxuan Wan, Wenxuan Wang, Yiliu Yang, Youliang Yuan, Jen tse Huang, Pinjia He, Wenxiang Jiao, and Michael~R. Lyu. 2024.
\newblock \href {http://arxiv.org/abs/2401.00757} {Logicasker: Evaluating and improving the logical reasoning ability of large language models}.

\bibitem[{Wang et~al.(2024{\natexlab{a}})Wang, Zhao, Qiang, Qin, and Liu}]{wang2024answers}
Haochun Wang, Sendong Zhao, Zewen Qiang, Bing Qin, and Ting Liu. 2024{\natexlab{a}}.
\newblock \href {http://arxiv.org/abs/2402.01349} {Beyond the answers: Reviewing the rationality of multiple choice question answering for the evaluation of large language models}.

\bibitem[{Wang et~al.(2023)Wang, Wang, Li, Gao, Yin, and Ren}]{wang-etal-2023-scott}
Peifeng Wang, Zhengyang Wang, Zheng Li, Yifan Gao, Bing Yin, and Xiang Ren. 2023.
\newblock \href {https://doi.org/10.18653/v1/2023.acl-long.304} {{SCOTT}: Self-consistent chain-of-thought distillation}.
\newblock In \emph{Proceedings of the 61st Annual Meeting of the Association for Computational Linguistics (Volume 1: Long Papers)}, pages 5546--5558, Toronto, Canada. Association for Computational Linguistics.

\bibitem[{Wang et~al.(2024{\natexlab{b}})Wang, Wei, Choi, and Ren}]{wang-etal-2024-llms}
Siyuan Wang, Zhongyu Wei, Yejin Choi, and Xiang Ren. 2024{\natexlab{b}}.
\newblock \href {https://doi.org/10.18653/v1/2024.acl-long.406} {Can {LLM}s reason with rules? logic scaffolding for stress-testing and improving {LLM}s}.
\newblock In \emph{Proceedings of the 62nd Annual Meeting of the Association for Computational Linguistics (Volume 1: Long Papers)}, pages 7523--7543, Bangkok, Thailand. Association for Computational Linguistics.

\bibitem[{Wang et~al.(2024{\natexlab{c}})Wang, Ma, Zhang, Ni, Chandra, Guo, Ren, Arulraj, He, Jiang, Li, Ku, Wang, Zhuang, Fan, Yue, and Chen}]{wang2024mmluprorobustchallengingmultitask}
Yubo Wang, Xueguang Ma, Ge~Zhang, Yuansheng Ni, Abhranil Chandra, Shiguang Guo, Weiming Ren, Aaran Arulraj, Xuan He, Ziyan Jiang, Tianle Li, Max Ku, Kai Wang, Alex Zhuang, Rongqi Fan, Xiang Yue, and Wenhu Chen. 2024{\natexlab{c}}.
\newblock \href {http://arxiv.org/abs/2406.01574} {Mmlu-pro: A more robust and challenging multi-task language understanding benchmark (published at neurips 2024 track datasets and benchmarks)}.

\bibitem[{Wei et~al.(2022)Wei, Wang, Schuurmans, Bosma, brian ichter, Xia, Chi, Le, and Zhou}]{wei2022chain}
Jason Wei, Xuezhi Wang, Dale Schuurmans, Maarten Bosma, brian ichter, Fei Xia, Ed~H. Chi, Quoc~V Le, and Denny Zhou. 2022.
\newblock \href {https://openreview.net/forum?id=_VjQlMeSB_J} {Chain of thought prompting elicits reasoning in large language models}.
\newblock In \emph{Advances in Neural Information Processing Systems}.

\bibitem[{Weizenbaum(1966)}]{weizenbaum1966eliza}
Joseph Weizenbaum. 1966.
\newblock Eliza—a computer program for the study of natural language communication between man and machine.
\newblock \emph{Communications of the ACM}, 9(1):36--45.

\bibitem[{Welbl et~al.(2017)Welbl, Liu, and Gardner}]{welbl-etal-2017-crowdsourcing}
Johannes Welbl, Nelson~F. Liu, and Matt Gardner. 2017.
\newblock \href {https://doi.org/10.18653/v1/W17-4413} {Crowdsourcing multiple choice science questions}.
\newblock In \emph{Proceedings of the 3rd Workshop on Noisy User-generated Text}, pages 94--106, Copenhagen, Denmark. Association for Computational Linguistics.

\bibitem[{Weston et~al.(2015)Weston, Bordes, Chopra, Rush, Van~Merri{\"e}nboer, Joulin, and Mikolov}]{weston2015towards}
Jason Weston, Antoine Bordes, Sumit Chopra, Alexander~M Rush, Bart Van~Merri{\"e}nboer, Armand Joulin, and Tomas Mikolov. 2015.
\newblock Towards ai-complete question answering: A set of prerequisite toy tasks.
\newblock \emph{arXiv preprint arXiv:1502.05698}.

\bibitem[{Williams et~al.(2018)Williams, Nangia, and Bowman}]{williams2018broad}
Adina Williams, Nikita Nangia, and Samuel~R Bowman. 2018.
\newblock A broad-coverage challenge corpus for sentence understanding through inference.
\newblock In \emph{Proceedings of NAACL-HLT}, pages 1112--1122.

\bibitem[{Winograd(1971)}]{winograd1971procedures}
T~Winograd. 1971.
\newblock Procedures as a representation for data in a computer program for understanding natural language, mit ai technical report 235.

\bibitem[{Wittgenstein(1922)}]{wittgenstein1922tractatus}
Ludwig Wittgenstein. 1922.
\newblock \emph{Tractatus Logico Philosophicus: Logical-Philosophical Treatise}.
\newblock Really Simple Media.

\bibitem[{Wolf et~al.(2020)Wolf, Debut, Sanh, Chaumond, Delangue, Moi, Cistac, Rault, Louf, Funtowicz, Davison, Shleifer, von Platen, Ma, Jernite, Plu, Xu, Le~Scao, Gugger, Drame, Lhoest, and Rush}]{wolf-et-al-2019-huggingface}
Thomas Wolf, Lysandre Debut, Victor Sanh, Julien Chaumond, Clement Delangue, Anthony Moi, Pierric Cistac, Tim Rault, Remi Louf, Morgan Funtowicz, Joe Davison, Sam Shleifer, Patrick von Platen, Clara Ma, Yacine Jernite, Julien Plu, Canwen Xu, Teven Le~Scao, Sylvain Gugger, Mariama Drame, Quentin Lhoest, and Alexander Rush. 2020.
\newblock Transformers: State-of-the-art natural language processing.
\newblock In \emph{Empirical Methods in Natural Language Processing: System Demonstrations}, pages 38--45.

\bibitem[{Wu et~al.(2023)Wu, Qiu, Ross, Akyürek, Chen, Wang, Kim, Andreas, and Kim}]{wu2023reasoning}
Zhaofeng Wu, Linlu Qiu, Alexis Ross, Ekin Akyürek, Boyuan Chen, Bailin Wang, Najoung Kim, Jacob Andreas, and Yoon Kim. 2023.
\newblock \href {http://arxiv.org/abs/2307.02477} {Reasoning or reciting? exploring the capabilities and limitations of language models through counterfactual tasks}.

\bibitem[{Yanaka et~al.(2019)Yanaka, Mineshima, Bekki, Inui, Sekine, Abzianidze, and Bos}]{yanaka2019help}
Hitomi Yanaka, Koji Mineshima, Daisuke Bekki, Kentaro Inui, Satoshi Sekine, Lasha Abzianidze, and Johan Bos. 2019.
\newblock Help: A dataset for identifying shortcomings of neural models in monotonicity reasoning.
\newblock \emph{arXiv preprint arXiv:1904.12166}.

\bibitem[{Young et~al.(2022)Young, Bao, Bensemann, and Witbrock}]{young2022abductionrules}
Nathan Young, Qiming Bao, Joshua Bensemann, and Michael~J Witbrock. 2022.
\newblock Abductionrules: Training transformers to explain unexpected inputs.
\newblock In \emph{Findings of the Association for Computational Linguistics: ACL 2022}, pages 218--227.

\bibitem[{Yu et~al.(2020)Yu, Jiang, Dong, and Feng}]{yu2020reclor}
Weihao Yu, Zihang Jiang, Yanfei Dong, and Jiashi Feng. 2020.
\newblock Reclor: A reading comprehension dataset requiring logical reasoning.
\newblock In \emph{International Conference on Learning Representations (ICLR)}.

\bibitem[{Yuan et~al.(2023)Yuan, Hu, Vuli{\'c}, Korhonen, and Meng}]{yuan2023can}
Zhangdie Yuan, Songbo Hu, Ivan Vuli{\'c}, Anna Korhonen, and Zaiqiao Meng. 2023.
\newblock Can pretrained language models (yet) reason deductively?
\newblock In \emph{Proceedings of the 17th Conference of the European Chapter of the Association for Computational Linguistics}, pages 1439--1454.

\bibitem[{Zellers et~al.(2019)Zellers, Holtzman, Bisk, Farhadi, and Choi}]{zellers2019hellaswag}
Rowan Zellers, Ari Holtzman, Yonatan Bisk, Ali Farhadi, and Yejin Choi. 2019.
\newblock Hellaswag: Can a machine really finish your sentence?
\newblock In \emph{Proceedings of the 57th Annual Meeting of the Association for Computational Linguistics}, pages 4791--4800.

\bibitem[{Zhang et~al.(2022)Zhang, Li, Meng, Chang, and den Broeck}]{zhang2022paradox}
Honghua Zhang, Liunian~Harold Li, Tao Meng, Kai-Wei Chang, and Guy~Van den Broeck. 2022.
\newblock \href {http://arxiv.org/abs/2205.11502} {On the paradox of learning to reason from data}.

\bibitem[{Zhang et~al.(2024)Zhang, Da, Lee, Robinson, Wu, Song, Zhao, Raja, Slack, Lyu, Hendryx, Kaplan, Lunati, and Yue}]{zhang2024careful}
Hugh Zhang, Jeff Da, Dean Lee, Vaughn Robinson, Catherine Wu, Will Song, Tiffany Zhao, Pranav Raja, Dylan Slack, Qin Lyu, Sean Hendryx, Russell Kaplan, Michele Lunati, and Summer Yue. 2024.
\newblock \href {http://arxiv.org/abs/2405.00332} {A careful examination of large language model performance on grade school arithmetic}.

\bibitem[{Zhao et~al.(2024{\natexlab{a}})Zhao, Tong, Mou, Zhang, Zhang, and Huang}]{zhao2024exploringcompositionaldeficiencylarge}
Jun Zhao, Jingqi Tong, Yurong Mou, Ming Zhang, Qi~Zhang, and Xuanjing Huang. 2024{\natexlab{a}}.
\newblock \href {http://arxiv.org/abs/2405.06680} {Exploring the compositional deficiency of large language models in mathematical reasoning}.

\bibitem[{Zhao et~al.(2024{\natexlab{b}})Zhao, Chiu, Hwang, Brahman, Hessel, Choudhury, Choi, Li, and Suhr}]{zhao-etal-2024-uncommonsense}
Wenting Zhao, Justin Chiu, Jena Hwang, Faeze Brahman, Jack Hessel, Sanjiban Choudhury, Yejin Choi, Xiang Li, and Alane Suhr. 2024{\natexlab{b}}.
\newblock \href {https://doi.org/10.18653/v1/2024.naacl-long.469} {{UN}commonsense reasoning: Abductive reasoning about uncommon situations}.
\newblock In \emph{Proceedings of the 2024 Conference of the North American Chapter of the Association for Computational Linguistics: Human Language Technologies (Volume 1: Long Papers)}, pages 8487--8505, Mexico City, Mexico. Association for Computational Linguistics.

\bibitem[{Zheng et~al.(2024)Zheng, Zhou, Meng, Zhou, and Huang}]{chuijie2024mcq}
Chujie Zheng, Hao Zhou, Fandong Meng, Jie Zhou, and Minlie Huang. 2024.
\newblock \href {https://openreview.net/forum?id=shr9PXz7T0} {Large language models are not robust multiple choice selectors}.
\newblock In \emph{The Twelfth International Conference on Learning Representations}.

\bibitem[{Zhong et~al.(2021)Zhong, Wang, Tang, Xu, Guo, Wang, Yin, Zhou, and Duan}]{zhong2021ar}
Wanjun Zhong, Siyuan Wang, Duyu Tang, Zenan Xu, Daya Guo, Jiahai Wang, Jian Yin, Ming Zhou, and Nan Duan. 2021.
\newblock Ar-lsat: Investigating analytical reasoning of text.
\newblock \emph{arXiv preprint arXiv:2104.06598}.

\bibitem[{Zhou et~al.(2024{\natexlab{a}})Zhou, Staats, Li, Szegedy, Weinberger, and Wu}]{zhou2024dont}
Jin~Peng Zhou, Charles~E Staats, Wenda Li, Christian Szegedy, Kilian~Q Weinberger, and Yuhuai Wu. 2024{\natexlab{a}}.
\newblock \href {https://openreview.net/forum?id=V5tdi14ple} {Don't trust: Verify -- grounding {LLM} quantitative reasoning with autoformalization}.
\newblock In \emph{The Twelfth International Conference on Learning Representations}.

\bibitem[{Zhou et~al.(2024{\natexlab{b}})Zhou, Zhu, Antognini, Kim, and Zhang}]{zhou-etal-2024-paraphrase}
Yue Zhou, Yada Zhu, Diego Antognini, Yoon Kim, and Yang Zhang. 2024{\natexlab{b}}.
\newblock \href {https://doi.org/10.18653/v1/2024.naacl-long.153} {Paraphrase and solve: Exploring and exploiting the impact of surface form on mathematical reasoning in large language models}.
\newblock In \emph{Proceedings of the 2024 Conference of the North American Chapter of the Association for Computational Linguistics: Human Language Technologies (Volume 1: Long Papers)}, pages 2793--2804, Mexico City, Mexico. Association for Computational Linguistics.

\bibitem[{Zhu et~al.(2024)Zhu, Chen, Wang, Gong, Yang, and Xie}]{zhu2024dyval}
Kaijie Zhu, Jiaao Chen, Jindong Wang, Neil~Zhenqiang Gong, Diyi Yang, and Xing Xie. 2024.
\newblock \href {https://openreview.net/forum?id=gjfOL9z5Xr} {Dyval: Dynamic evaluation of large language models for reasoning tasks}.
\newblock In \emph{The Twelfth International Conference on Learning Representations}.

\end{thebibliography}

\clearpage

\appendix

\numberwithin{equation}{section}
\renewcommand{\thefigure}{\Alph{section}.\arabic{figure}}
\renewcommand{\thetable}{\Alph{section}.\arabic{table}}

\section{Related Work}   \label{appendix:sec:related_work}

\subsection{Investigation of Reasoning Capabilities of LLMs}
Many studies examine LLMs' reasoning capabilities \citep{askell2020gpt,rae2021scaling,razeghi2022impact,liu2023evaluating,turpin2023language,lanham2023measuring,wu2023reasoning,hodel2023response,dziri2023faith,dasgupta2023language}.
\citet{patel2024multilogievalevaluatingmultisteplogical} observed LLMs' performance significantly declines as reasoning steps increase in multi-step logical reasoning tasks.
\citet{dougrezlewis2024assessingreasoningabilitieschatgpt} revealed ChatGPT struggles with abductive reasoning when verifying claims by decomposing their evidence into atomic reasoning steps.
\citet{wang-etal-2024-llms} found that GPT-series models showed significant gaps compared to humans in dealing with inference rules.
\citet{parmar-etal-2024-logicbench} introduced LogicBench and showed that existing LLMs struggle with instances involving complex reasoning and negations. 
\citet{wan2024logicaskerevaluatingimprovinglogical} introduced LogicAsker, which assesses whether LLMs can employ a set of atomic reasoning skills grounded in propositional and predicate logic and found significant gaps in LLMs' learning of logical rules.
\citet{bhuiya2024seeminglyplausibledistractorsmultihop} proposed a challenging multi-hop reasoning benchmark with seemingly plausible but incorrect multi-hop reasoning chains and found that state-of-the-art LLMs' capabilities to perform multi-hop reasoning is affected by such chains.
\citet{mondorf2024liarliarlogicalmire} introduced TruthQuest, which assesses LLMs' capabilities to conduct suppositional reasoning, i.e., reasoning where each statement can be false, and found that LLMs exhibit significant difficulties solving these tasks.
\citet{sprague2024musr} introduced a complex multi-step reasoning benchmark, MuSR, and characterized the gaps that remain for techniques like chain-of-thought to perform robust reasoning.

\paragraph{Biases and Errors}
\citet{ando-etal-2023-evaluating,ozeki-etal-2024-exploring,bertolazzi2024systematicanalysislargelanguage,eisape-etal-2024-systematic} found that LLMs exhibit human-like reasoning biases in syllogistic arguments.
\citet{jiang2024peektokenbiaslarge} found that LLMs exibit ``token-biases'' in solving logical reasoning problems.
\citet{aoki2024heuristicrationaldynamicuse} revealed that LMs rely heavily on heuristics, such as lexical overlap, in the earlier stages of reasoning.
\citet{zhao2024exploringcompositionaldeficiencylarge} constructed a \textsc{MathTrap} with carefully designed logical traps into the problem descriptions of MATH and GSM8k and found that while LLMs possess the knowledge required to solve these traps, they do not spontaneously use such knowledge them to handle the problems.
\citet{han2024incontextlearningelicittrustworthy} found that LLMs exhibit A-Not-B errors similar to human infants, failing to suppress the previously established response pattern during ICL.
\citet{liu2024selfcontradictoryreasoningevaluationdetection} found that LLMs often contradict themselves in reasoning tasks involving contextual information understanding or commonsense.
\citet{zhou-etal-2024-paraphrase} found that subtle alterations in the surface form can significantly impact the answer distribution, suggesting that LLMs solve reasoning problems using surface cues.
\citet{pmlr-v235-chen24i} found that the reasoning performance of LLMs is affected by the order of the premises.
\citet{hong-etal-2024-closer,huang2024large} found that LLMs struggle to identify fallacious reasoning steps accurately, suggesting challenges in self-verification methods.

\paragraph{Reasoning in Unknown Situation}
\citet{zhao-etal-2024-uncommonsense} found that LLMs struggle with reasoning in uncommon situations. 
\citet{zhu2024dyval} introduced a framework to dynamically generate reasoning samples, and LLMs perform worse in those samples.
\citet{hu2024largelanguagemodelslimited} found that while LLMs can conduct reasoning when relevant knowledge is given in context, they are not proficient at reasoning with knowledge embedded in the training data.

\subsection{Synthetic Logic Corpus for Training LLMs}

RuleTaker \cite{clark2020transformers} proposed a deduction corpus composed of synthetically generated multistep deductive proofs written in natural languages.
Each deductive proof (dis-)proves a hypothesis by applying deduction rules multiple times to a given set of facts.
They showed that Transformer \cite{vaswani2017attention} LMs can solve these problems in the sense that they can predict the final answer (i.e., ``proved'', ``disproved'', or ``unknown'') of each deductive proof given the fact set.
Later studies \cite{saha-etal-2020-prover,dalvi-etal-2021-explaining,tafjord-etal-2021-proofwriter,sanyal2022fairr} showed that generative LMs can generate even the intermediate proofs as well as the final answer.
Later studies \citep{saha-etal-2020-prover,dalvi-etal-2021-explaining,tafjord-etal-2021-proofwriter,sanyal2022fairr} showed that T5 can generate even the intermediate logical steps as well as the final answer.

PARARULE-Plus \citep{bao2022multi} is the enhanced version of PARARULE \citep{clark2020transformers}, a variation of RuleTaker, that includes more samples and more logical steps.
RoBERTa \citep{liu-et-al-2019-roberta} trained on PARARULE-Plus outperformed the models trained on RuleTaker.

Artificial Argument Corpus \citep{betz-etal-2021-critical} includes single-step deductive reasoning samples constructed from hand-selected deduction rules useful for critical thinking.
They showed that the GPT-2 \citep{radford2019language} trained on this corpus can generalize to solve NLI tasks.
However, at the same time, they found that the LM does not generalize well to solve more challenging reasoning tasks such as ARC \citep{habernal-etal-2018-argument} and LogiQA \citep{ijcai2020p501}.

FLD by \citet{pmlr-v202-morishita23a,morishita-etal-2024-jfld} is the first synthetic logic corpus based on formal logic theory.
It includes multistep deductive reasoning samples constructed from the axioms of first-order predicate logic, which can express any deduction rule due to the completeness theorem.
Due to this nature, T5 trained on FLD generalizes most effectively to other synthetic logic corpora, compared to models trained on other corpora.

\citet{gontier2020measuring} investigated the deductive reasoning capabilities of LMs on a corpus composed of a specific type of multistep inference, kinship relationships on synthetic kinship graphs.
They found that LMs can solve this task when there are relatively few proof steps, but it is difficult for them to generalize to solve proof steps longer than those shown in training data.
\citet{bostrom-etal-2021-flexible} studied how to create realistic natural language expressions that represent deduction rules.
To this end, they scraped sentences from Wikipedia using a template-based method and paraphrased them.
They showed that training on this corpus helps solve real-world deductive reasoning problems such as EntailmentBank \citep{dalvi-etal-2021-explaining}.
\citet{pi-etal-2022-reasoning} used synthetic data from program executors, most notably SQL programs.
They verified that this data can enhance numerical reasoning, logical reasoning, and multi-hop reasoning abilities.
\citet{trinh2024solving} generated 100 million geometry problems and verified that the capability of artificial intelligence can be enhanced to to pass the bronze medal threshold of the International Mathematics Olympiad.
\citet{saeed-etal-2021-rulebert,nafar-etal-2024-teaching} created \textit{soft} reasoning rules involving with probabilistic logic, instead of hard-logic examined by the aformentioned studies.
\citet{sileo2024scalingsyntheticlogicalreasoning} introduced a simpler and more general declarative framework for synthetic generation, and verified its effectiveness.
\citet{zhou2024dont} synthetically generated a large dataset of mathematics, and gained over 12 points on GSM8k.

While these studies partly examined the effect of synthetic logic corpora, whether this approach is promising remains an open question.
It has been unexplored whether the capabilities obtained from synthetic logic corpora generalizes to solve various tasks beyond the original tasks in these corpora.
Additionally, the effect of these corpora has only been examined for small LMs trained on small pre-training corpora such as T5 and RoBERTa; it has been highly questionable whether they can still benefit state-of-the-art LLMs trained on a huge pre-training corpus.
Furthermore, even if their benefits were verified, it remains unclear which design of synthetic logic samples yields the largest benefits due to the lack of systematic discussions on sample designs and empirical verification of these designs.
We aimed to answer these questions in this paper and demonstrate the potential of synthetic logic corpora.

\subsection{Distilling Reasoning Traces from Very Large LLMs}

Recent approaches \citep{ho2023large,magister-etal-2023-teaching,li2022explanations,li-etal-2023-symbolic,shridhar-etal-2023-distilling,wang-etal-2023-scott,mitra2023orca,liu-etal-2023-logicot,benallal2024cosmopedia,lu2024mathgenie} utilize very large LLMs, such as GPT-4, to prepare synthetic reasoning datasets to train smaller LLMs.
A typical procedure is as follows: (i) prepare existing reasoning problems, (ii) prompt large LLMs to generate reasoning traces to solve these problems using techniques such as chain-of-thought prompting \citep{wei2022chain}, and (iii) train smaller LLMs on these reasoning traces.

The distillation approach and the synthetic logic corpora approach examined in this paper have specific advantages and disadvantages, as follows.

The advantage of the distillation approach is its immediate practical effect, as it directly teaches LLMs solutions to various existing problems.
The disadvantages could be that
(i) it is non-trivial for specific solutions to specific problems to generalize to other problems,
(ii) the number of training samples is limited to existing problems in nature,
(iii) the correctness and faithfulness of the reasoning traces are not guaranteed; indeed, some studies \citep{turpin2023language,lanham2023measuring} suggest that large LLMs do not always faithfully follow the ``reasoning traces'' they themselves generate, and
(iv) it cannot enhance the very large LLMs themselves by nature.

The advantages of synthetic logic corpus approaches are that
(i) since they teach the fundamentals of reasoning, such as deductive reasoning, they have the potential to generalize to various problems,
(ii) they can generate an unlimited number of new samples, and
(iii) the correctness of the reasoning traces is guaranteed by nature.
The disadvantage of this approach is that, as it only teaches the basics of reasoning, additional training may be needed to solve more complex real-world problems, as suggested in \Cref{sec:which_task_others}.

We hypothesize that integrating both approaches could be promising. That is, we first train LLMs using \ALT \ to make them understand the fundamentals of reasoning through high-quality samples and then train them using more realistic reasoning traces to solve complex real-world problems.

\section{Limitations}  \label{appendix:sec:limitations}
\begin{itemize}
    \item We only used deductive reasoning samples for \ALT. Future work should examine other reasoning samples, e.g., abductive and inductive reasoning.
    \item We only examined the first-order predicate logic system. Future work should examine other logic systems, such as modal and linear logic.
\end{itemize}

\section{Ethics and Social Impacts}  \label{appendix:sec:social_impacts}
The ultimate goal of the direction of this study is to develop an AI capable of reasoning logically step by step.
If AI can make a decision one logical step at a time, it would be highly explainable and transparent to users.
Furthermore, the user would be able to trace the AI's errors.
We believe that our study is a step towards such AI that will positively impact society.

\section{Details of \PLD}   \label{appendix:sec:PLD}

\Cref{appendix:fig:deduction_example} shows a real sample from \PLDAbbr.
Below, We briefly explain our sample generator.
Please refer to \citet{pmlr-v202-morishita23a} for the details.

\subsection{Answer Labels}   \label{appendix:sec:answer_label}
\begin{figure*}[t!]
    \centering
    \includegraphics[width=1\linewidth]{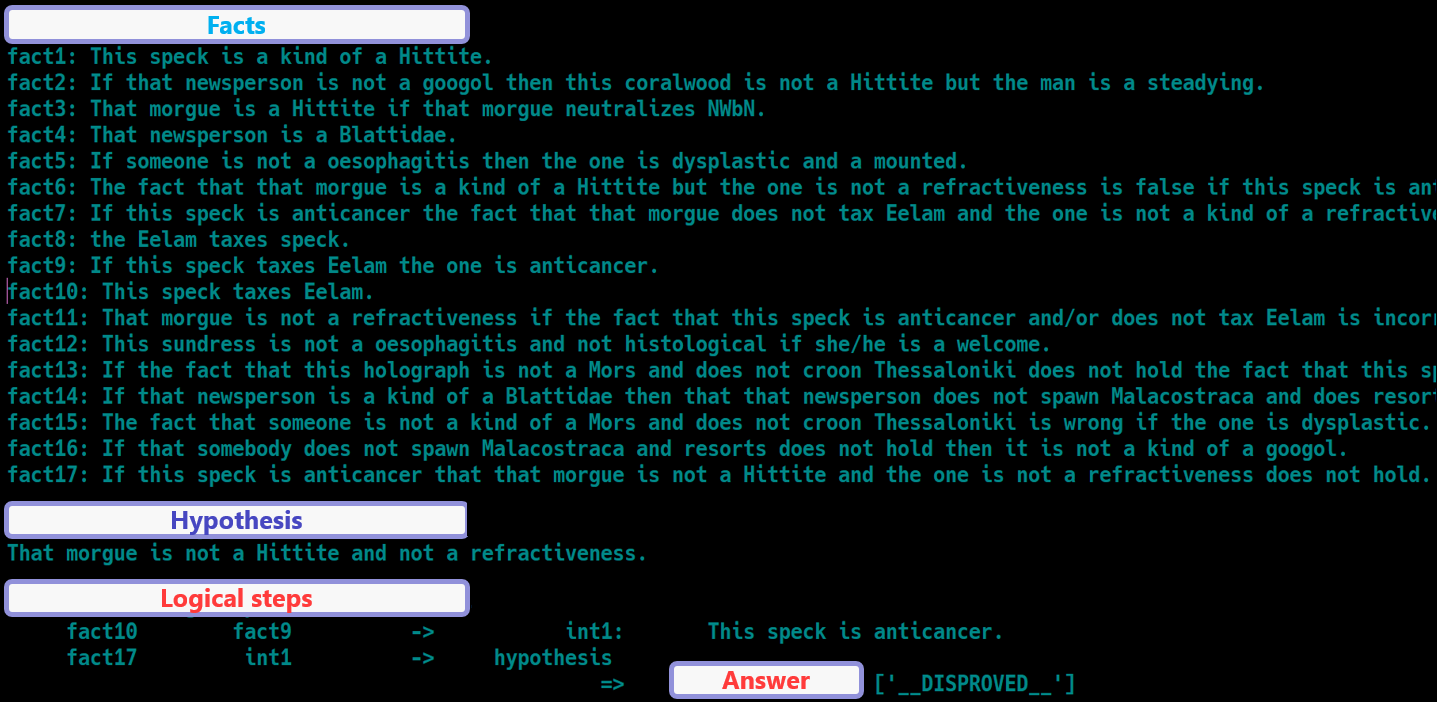}
    \caption{
        A real deduction sample included in \PLD.
        \textbf{\colorBlueFacts{Facts}} and \textbf{\colorVioletHypothesis{hypothesis}} are given to LLMs, then the LLMs are required to generate \textbf{\colorRedLogicalSteps{logical steps}} to (dis-)prove the hypothesis based on the facts, and an \textbf{\colorRedLogicalSteps{answer}} label (see \Cref{appendix:sec:PLD_statistics}).
        \label{appendix:fig:deduction_example}
    }
\end{figure*}

In addition to the logical steps, the samples of \PLDAbbr \ and previous corpora include \textit{answer labels} (\Cref{appendix:fig:deduction_example}): ``proved'' indicating that the hypothesis can be proved by the logical steps, ``disproved'' indicating that the hypothesis can be disproved, and ``unknown'' indicating that the given facts are insufficient for either proving or disproving the hypothesis.
For samples with ``unknown'' labels, the logical steps are ``None.''.
\PLDAbbr \ have a uniform distribution over the labels.

\subsection{Splits}   \label{appendix:sec:PLD_statistics}

\PLDAbbr \ includes 100k/5k/5k samples for train/valid/test splits.

\subsection{Generation of Multistep Deduction} \label{appendix:sec:FLD_proof_tree_generation}

Our sample generator first randomly generates examples of multistep deduction by forward- and backward random deduction, using the deduction rules specified by a user.

The forward random deduction is done as follows.
The generator first chooses a deduction rule randomly and forms the initial tree where the root node is the conclusion of the chosen deduction rules and the child nodes are the premises of the chosen deduction rule.
The generator next randomly chooses another deduction rule that can be ``jointed'' to the root note of the tree.
A deduction rule can be jointed to the root node of a tree if one of the premises of that deduction rule can be identified with the root node.
Then, the generator updates the tree by jointing this chosen deduction rule.
The generator continues this step multiple times until the tree achieves the required depth.

The backward random deduction is done as follows.
For each step, the generator randomly chooses a leaf node of the tree.
Then, the generator randomly chooses a deduction rule that can be jointed to the leaf node.
Here, a deduction rule can be jointed to the leaf node if the deduction rule's conclusion can be identified with the leaf node.
Then, the generator updates the tree by jointing this chosen deduction rule.
The generator continues this step multiple times until the complexity of branches achieves the required level.

\subsection{Linguistic Expressions}   \label{appendix:sec:PLD_linguistic_diversity}

We prepared linguistic templates for each logical formula, exemplified as follows:
\begin{alignSmall}
    \langle(A \land B) \rightarrow C\rangle: & \ \text{If} \ \langle(A \land B)\text{.predicate\_phrase}\rangle, \ \text{then} \ \langle C\text{.predicate\_phrase}\rangle. \nonumber \\
    : & \ \langle (A \land B).\text{noun\_pharse} \rangle \ \langle \text{cause\_synonyms} \rangle \ \langle C\text{.noun\_phrase} \rangle. \nonumber \\
    : & \ (\dots) \nonumber \\
    \langle(A \land B)\text{.predicate\_phrase}\rangle: & \ A \ \langle \text{occur\_synonyms} \rangle \ \text{and also} \ B \ \langle \text{occur\_synonyms} \rangle.  \nonumber \\
    : & \ A \ \text{and also} \ B \ \langle \text{occur\_synonyms} \rangle. \nonumber \\
    : & \ \text{Both} \ A \ \text{and} \ B \ \langle \text{occur\_synonyms} \rangle.  \nonumber \\
    : & \ (\dots)   \nonumber \\ 
    \langle C\text{.predicate\_phrase}\rangle: & \ C \ \langle \text{occur\_synonyms} \rangle.  \nonumber \\
    : & \ (\dots)   \nonumber \\
    \langle \text{occur\_synonyms} \rangle: & \ \text{occur} \nonumber \\
    : & \ \text{happen} \nonumber \\
    : & \ \text{take place} \nonumber \\
    : & \ (\dots) \nonumber \\
    \langle (A \land B).\text{noun\_pharse}\rangle: & \ A \ \text{and} \ B \nonumber \\ 
    : & \ A \ \text{and also} \ B \nonumber \\ 
    : & \ \text{Both} \ A \ \text{and} \ B \nonumber \\ 
    : & \ \text{That} \ A \ \text{and} \ B \ \langle \text{occur\_synonyms} \rangle \nonumber \\ 
    : & \ (\dots) \nonumber \\ 
    \langle \text{cause\_synonyms} \rangle: & \ \text{cause} \nonumber \\
    : & \ \text{result in} \nonumber \\
    : & \ \text{lead to} \nonumber \\
    : & \ \text{bring about} \nonumber \\
    : & \ (\dots) \nonumber \\
    ( & \dots) 
    \label{appendix:eq:linguistic_templates}
\end{alignSmall}
As can be seen, the templates can be nested deeply, yielding combinatorially diverse linguistic expressions.

Expanding these templates beforehand is intractable due to the combinatorial explosion, so we expand these templates on the fly to randomly sample a single expression at a time.
Estimating the exact number of expressions is intractable for the same reason.

We manually crafted several additional English templates per logical formula (i.e., the left-hand sides of \Cref{appendix:eq:linguistic_templates}) compared to those used in FLD, which yield combinatorially more diverse English expressions.
We observed that at least dozens of expressions, including minor variations, are yielded for each formula.

\section{Details of Experimental Setup}   \label{appendix:sec:experiments}

\subsubsection{Prevention of Knowledge Forgetting by Recall Adam Optimizer}   \label{appendix:sec:experiments_recadam}

We employed the Recall Adam (RecAdam) optimizer \citep{recadam}, which regularizes parameter updates to prevent them from being too far from the pre-training parameters.
Recall Adam stands out for LLM training as it does not require access to the pre-training corpus, which is often inaccessible or too huge to handle, nor does it require changes to the model architecture, and it has a proven track record of usage in language models such as BERT.

\subsection{Benchmarks}   \label{appendix:sec:benchmarks}
\Cref{appendix:tb:benchmarks} details the benchmarks used in the experiments.

\subsection{Experimental Runs}   \label{appendix:sec:experiments_experimental_runs}
We show the average and standard deviations over five seeds.

\subsection{Computational Resources}   \label{appendix:sec:experiments_computation}
The entire experiment, including preliminary ones, took about 1 week x 128 NVIDIA H100 GPUs of our own.

\begin{table*}[t]
    \footnotesize
    \tabcolsep 5pt
    \centering
    \caption{
        \numBenchmarks\ benchmarks used in the experiments.
        These benchmarks cover a wide range of tasks and are prominent for LLM evaluation.
        We also show the form of reasoning and the type of knowledge required to solve the problems in each benchmark.
        \label{appendix:tb:benchmarks}    }

\begin{tabular}{@{}clcc@{}}
\toprule
Set                         & \multicolumn{1}{c}{Benchmarks}                                                & \begin{tabular}[c]{@{}c@{}}Reasoning\\ form\end{tabular}                                                     & \begin{tabular}[c]{@{}c@{}}Required\\ knowledge\end{tabular}                \\ \midrule
\multirow{9}{*}{Logic}      & bAbi deduction {\tiny \citep{weston2015towards}},                              & \multirow{7}{*}{deduction}                                                                                   & \multirow{4}{*}{\begin{tabular}[c]{@{}c@{}}-\\ (not required)\end{tabular}} \\
                            & FOLIO {\tiny\citep{han2022folio}}                                              &                                                                                                              &                                                                             \\
                            & LogicNLI {\tiny \citep{tian2021diagnosing}}                                    &                                                                                                              &                                                                             \\
                            & RobustLR  {\tiny \citep{sanyal2022robustlr}}                                   &                                                                                                              &                                                                             \\ \cmidrule(l){4-4} 
                            & AR-LSAT {\tiny \citep{zhong2021ar}}                                            &                                                                                                              & \multirow{3}{*}{commonsense}                                                \\
                            & LogiQA2 {\tiny \citep{logiqa2}}                                                &                                                                                                              &                                                                             \\
                            & ReClor {\tiny \citep{yu2020reclor}}                                            &                                                                                                              &                                                                             \\ \cmidrule(lr){3-3}
                            & AbductionRules {\tiny \citep{young2022abductionrules}}                         & \multirow{2}{*}{abduction}                                                                                   &                                                                             \\ \cmidrule(l){4-4} 
                            & ART {\tiny \citep{bhagavatula2019abductive}}                                   &                                                                                                              & commonsense                                                                 \\ \midrule
\multirow{4}{*}{NLI}        & HELP {\tiny\citep{yanaka2019help}}                                             & \multirow{4}{*}{\begin{tabular}[c]{@{}c@{}}validate\\ a conclusion\\ based on\\ given premises\end{tabular}} & \multirow{4}{*}{commonsense}                                                \\
                            & MultiNLI {\tiny\citep{williams2018broad}}                                      &                                                                                                              &                                                                             \\
                            & RTE {\tiny \citep{dagan2005pascal, giampiccolo2007third, bentivogli2009fifth}} &                                                                                                              &                                                                             \\
                            & SNLI {\tiny \citep{bowmanlarge}}                                               &                                                                                                              &                                                                             \\ \midrule
\multirow{3}{*}{Math}       & GSM8k {\tiny \citep{cobbe2021gsm8k}}                                           & \multirow{3}{*}{Math}                                                                                        & \multirow{3}{*}{Math}                                                       \\
                            & MATH {\tiny \citep{hendrycksmath2021}}                                         &                                                                                                              &                                                                             \\
                            & MathQA {\tiny \citep{amini-etal-2019-mathqa}}                                  &                                                                                                              &                                                                             \\ \midrule
\multirow{4}{*}{Coding}     & HumanEval {\tiny \citep{chen2021codex}}                                        & \multirow{4}{*}{Coding}                                                                                      & \multirow{4}{*}{Coding}                                                     \\
                            & MBPP {\tiny \citep{austin2021program}}                                         &                                                                                                              &                                                                             \\
                            & MBPP+ {\tiny \citep{evalplus}}                                                 &                                                                                                              &                                                                             \\
                            & MultiPL-E (cpp/go) {\tiny \citep{multiple}}                                    &                                                                                                              &                                                                             \\ \midrule
\multirow{8}{*}{Others}     & CommonsenseQA {\tiny \citep{talmor2018commonsenseqa}}                          & \multirow{8}{*}{\begin{tabular}[c]{@{}c@{}}complicated\\ procedures\end{tabular}}                            & \multirow{4}{*}{commonsense}                                                \\
                            & HellaSWAG {\tiny \citep{zellers2019hellaswag}}                                 &                                                                                                              &                                                                             \\
                            & SQuAD2 {\tiny \citep{rajpurkar-etal-2018-know}}                                &                                                                                                              &                                                                             \\
                            & WinoGrande {\tiny \citep{sakaguchi2021winogrande}}                             &                                                                                                              &                                                                             \\ \cmidrule(lr){2-2} \cmidrule(l){4-4} 
                            & ARC (easy/challenge) {\tiny \citep{clark2018think}}                            &                                                                                                              & \multirow{4}{*}{science}                                                    \\
                            & GPQA  {\tiny \citep{rein2023gpqa}}                                             &                                                                                                              &                                                                             \\
                            & OpenBookQA {\tiny \citep{OpenBookQA2018}}                                      &                                                                                                              &                                                                             \\
                            & SciQ {\tiny \citep{welbl-etal-2017-crowdsourcing}}                             &                                                                                                              &                                                                             \\ \midrule
\multirow{3}{*}{aggregated} & MMLU {\tiny \citep{hendryckstest2021}}                                         & \multirow{3}{*}{various}                                                                                     & \multirow{3}{*}{various}                                                    \\
                            & MMLU-Pro {\tiny \citep{wang2024mmluprorobustchallengingmultitask}}             &                                                                                                              &                                                                             \\
                            & BBH {\tiny \citep{suzgun2022challenging}}                                      &                                                                                                              &                                                                             \\ \bottomrule
\end{tabular}

\end{table*}

\section{Results without using Recall Adam}   \label{appendix:sec:results_other_LLMs}

\Cref{appendix:tb:performance_aggregated} shows the results of LLMs trained without using Recall Adam.

\clearpage
\begin{table*}[t!]
    \centering
    \tabcolsep 4pt
    \footnotesize
    \caption{
    5-shot performance of LLMs before and after \ALT.
    $\oplus$\textbf{\ALT}-$x$ denotes the LLM trained with \ALT \ on the synthetic logic corpus $x$ from \Cref{tb:corpora}.
    Color shows the rank in each column (darker is better).
    ``Logic'', ``Math'', ``Code'', and ``Others'' each comprises various benchmarks (see \Cref{appendix:tb:benchmarks}).
    ``Avg.'' represents the micro-average of all the benchmarks.
    ``w/o RecAdam'' denotes that LLM was trained without knowledge forgetting prevention by Recall Adam optimizer.
    \label{appendix:tb:performance_aggregated}
    }
    \vspace{-0.5mm}

    \begin{subfigure}{1.0\linewidth}
        \centering
        \subcaption{\llamaThreeBaseline.    \label{appendix:tb:performance_aggregated_llama_three}}
        \vspace{-1mm}
        \resizebox{\textwidth}{!}{

\begin{tabular}{lcccccccccccc}
\toprule
{} & Avg. & Logic & Math & Code & NLI & Others & \multicolumn{2}{c}{BBH (3-shot)} & \multicolumn{2}{c}{BBH (0-shot)} & \multicolumn{2}{c}{MMLU} \\
\cmidrule(l){2-2} \cmidrule(l){3-3} \cmidrule(l){4-4} \cmidrule(l){5-5} \cmidrule(l){6-6}  \cmidrule(l){7-7} \cmidrule(l){8-9} \cmidrule(l){10-11} \cmidrule(l){12-13}
{} & {} & {} & {} & {} & {} & {} & {} & CoT & {} & CoT & {} & Pro \\
\midrule

\llamaThreeBaseline & \cellcolor{blue!18} 47.9 & \cellcolor{blue!18} 42.8\scalebox{0.8}{$_{\pm\text{0.4}}$} & \cellcolor{blue!26} 39.6\scalebox{0.8}{$_{\pm\text{0.5}}$} & \cellcolor{blue!18} 35.4 & \cellcolor{blue!18} 65.4\scalebox{0.8}{$_{\pm\text{0.3}}$} & \cellcolor{blue!10} 60.7\scalebox{0.8}{$_{\pm\text{0.3}}$} & \cellcolor{blue!26} 44.9\scalebox{0.8}{$_{\pm\text{0.4}}$} & \cellcolor{blue!26} 61.9\scalebox{0.8}{$_{\pm\text{0.4}}$} & \cellcolor{blue!26} 8.2\scalebox{0.8}{$_{\pm\text{0.2}}$} & \cellcolor{blue!26} 36.5\scalebox{0.8}{$_{\pm\text{0.4}}$} & \cellcolor{blue!35} 65.3\scalebox{0.8}{$_{\pm\text{0.4}}$} & \cellcolor{blue!35} 35.8\scalebox{0.8}{$_{\pm\text{0.4}}$} \\
\llamaThreePRSingle & \cellcolor{blue!10} 43.5 & \cellcolor{blue!10} 39.5\scalebox{0.8}{$_{\pm\text{0.2}}$} & \cellcolor{blue!10} 29.1\scalebox{0.8}{$_{\pm\text{0.3}}$} & \cellcolor{blue!10} 35.3 & \cellcolor{blue!10} 57.8\scalebox{0.8}{$_{\pm\text{0.2}}$} & \cellcolor{blue!26} 61.0\scalebox{0.8}{$_{\pm\text{0.2}}$} & \cellcolor{blue!10} 40.5\scalebox{0.8}{$_{\pm\text{0.2}}$} & \cellcolor{blue!10} 47.0\scalebox{0.8}{$_{\pm\text{0.2}}$} & \cellcolor{blue!10} 3.9\scalebox{0.8}{$_{\pm\text{0.1}}$} & \cellcolor{blue!10} 6.3\scalebox{0.8}{$_{\pm\text{0.1}}$} & \cellcolor{blue!10} 64.9\scalebox{0.8}{$_{\pm\text{0.2}}$} & \cellcolor{blue!10} 34.0\scalebox{0.8}{$_{\pm\text{0.2}}$} \\
\llamaThreePRALT & \cellcolor{blue!26} 48.1 & \cellcolor{blue!26} 43.7\scalebox{0.8}{$_{\pm\text{0.2}}$} & \cellcolor{blue!18} 39.2\scalebox{0.8}{$_{\pm\text{0.3}}$} & \cellcolor{blue!26} 35.7 & \cellcolor{blue!26} 65.6\scalebox{0.8}{$_{\pm\text{0.2}}$} & \cellcolor{blue!18} 60.8\scalebox{0.8}{$_{\pm\text{0.2}}$} & \cellcolor{blue!26} 44.9\scalebox{0.8}{$_{\pm\text{0.2}}$} & \cellcolor{blue!18} 61.8\scalebox{0.8}{$_{\pm\text{0.2}}$} & \cellcolor{blue!26} 8.2\scalebox{0.8}{$_{\pm\text{0.1}}$} & \cellcolor{blue!18} 36.4\scalebox{0.8}{$_{\pm\text{0.2}}$} & \cellcolor{blue!35} 65.3\scalebox{0.8}{$_{\pm\text{0.2}}$} & \cellcolor{blue!18} 35.3\scalebox{0.8}{$_{\pm\text{0.2}}$} \\
\llamaThreeRTALT & \cellcolor{blue!35} 50.1 & \cellcolor{blue!35} 46.8\scalebox{0.8}{$_{\pm\text{0.1}}$} & \cellcolor{blue!35} 42.4\scalebox{0.8}{$_{\pm\text{0.2}}$} & \cellcolor{blue!35} 36.5 & \cellcolor{blue!35} 68.6\scalebox{0.8}{$_{\pm\text{0.1}}$} & \cellcolor{blue!35} 61.3\scalebox{0.8}{$_{\pm\text{0.1}}$} & \cellcolor{blue!51} 46.9\scalebox{0.8}{$_{\pm\text{0.2}}$} & \cellcolor{blue!35} 63.5\scalebox{0.8}{$_{\pm\text{0.2}}$} & \cellcolor{blue!51} 13.7\scalebox{0.8}{$_{\pm\text{0.1}}$} & \cellcolor{blue!35} 38.4\scalebox{0.8}{$_{\pm\text{0.2}}$} & \cellcolor{blue!35} 65.3\scalebox{0.8}{$_{\pm\text{0.1}}$} & \cellcolor{blue!26} 35.7\scalebox{0.8}{$_{\pm\text{0.2}}$} \\
\llamaThreeFLDALT & \cellcolor{blue!43} 51.9 & \cellcolor{blue!43} 51.6\scalebox{0.8}{$_{\pm\text{0.1}}$} & \cellcolor{blue!51} 43.4\scalebox{0.8}{$_{\pm\text{0.2}}$} & \cellcolor{blue!51} 38.1 & \cellcolor{blue!43} 70.1\scalebox{0.8}{$_{\pm\text{0.1}}$} & \cellcolor{blue!51} 61.5\scalebox{0.8}{$_{\pm\text{0.1}}$} & \cellcolor{blue!43} 46.7\scalebox{0.8}{$_{\pm\text{0.2}}$} & \cellcolor{blue!43} 64.9\scalebox{0.8}{$_{\pm\text{0.2}}$} & \cellcolor{blue!43} 11.9\scalebox{0.8}{$_{\pm\text{0.1}}$} & \cellcolor{blue!51} 39.6\scalebox{0.8}{$_{\pm\text{0.2}}$} & \cellcolor{blue!43} 65.4\scalebox{0.8}{$_{\pm\text{0.1}}$} & \cellcolor{blue!43} 36.2\scalebox{0.8}{$_{\pm\text{0.2}}$} \\
\llamaThreePLDALTBold & \cellcolor{blue!51} 52.0 & \cellcolor{blue!51} 52.2\scalebox{0.8}{$_{\pm\text{0.1}}$} & \cellcolor{blue!43} 43.2\scalebox{0.8}{$_{\pm\text{0.2}}$} & \cellcolor{blue!43} 38.0 & \cellcolor{blue!51} 70.7\scalebox{0.8}{$_{\pm\text{0.1}}$} & \cellcolor{blue!51} 61.5\scalebox{0.8}{$_{\pm\text{0.1}}$} & \cellcolor{blue!35} 46.5\scalebox{0.8}{$_{\pm\text{0.2}}$} & \cellcolor{blue!51} 65.3\scalebox{0.8}{$_{\pm\text{0.2}}$} & \cellcolor{blue!35} 11.3\scalebox{0.8}{$_{\pm\text{0.1}}$} & \cellcolor{blue!43} 38.7\scalebox{0.8}{$_{\pm\text{0.2}}$} & \cellcolor{blue!51} 65.5\scalebox{0.8}{$_{\pm\text{0.1}}$} & \cellcolor{blue!51} 36.4\scalebox{0.8}{$_{\pm\text{0.2}}$} \\
\bottomrule

\end{tabular}

        }
    \end{subfigure}
    \begin{subfigure}{1.0\linewidth}
        \centering
        \vspace{5mm}
        \subcaption{\llamaThreeLargeBaseline.    \label{appendix:tb:performance_aggregated_llama_three_large}}
        \resizebox{\textwidth}{!}{

\begin{tabular}{lcccccccccccc}
\toprule
{} & Avg. & Logic & Math & Code & NLI & Others & \multicolumn{2}{c}{BBH (3-shot)} & \multicolumn{2}{c}{BBH (0-shot)} & \multicolumn{2}{c}{MMLU} \\
\cmidrule(l){2-2} \cmidrule(l){3-3} \cmidrule(l){4-4} \cmidrule(l){5-5} \cmidrule(l){6-6}  \cmidrule(l){7-7} \cmidrule(l){8-9} \cmidrule(l){10-11} \cmidrule(l){12-13}
{} & {} & {} & {} & {} & {} & {} & {} & CoT & {} & CoT & {} & Pro \\
\midrule

LLaMA-3.1-70B & \cellcolor{blue!18} 60.0 & \cellcolor{blue!18} 57.4\scalebox{0.8}{$_{\pm\text{0.4}}$} & \cellcolor{blue!26} 60.0\scalebox{0.8}{$_{\pm\text{0.5}}$} & \cellcolor{blue!10} 46.2 & \cellcolor{blue!26} 73.7\scalebox{0.8}{$_{\pm\text{0.3}}$} & \cellcolor{blue!26} 67.7\scalebox{0.8}{$_{\pm\text{0.3}}$} & \cellcolor{blue!26} 60.4\scalebox{0.8}{$_{\pm\text{0.3}}$} & \cellcolor{blue!18} 82.1\scalebox{0.8}{$_{\pm\text{0.2}}$} & \cellcolor{blue!26} 6.5\scalebox{0.8}{$_{\pm\text{0.1}}$} & \cellcolor{blue!26} 50.1\scalebox{0.8}{$_{\pm\text{0.3}}$} & \cellcolor{blue!26} 78.7\scalebox{0.8}{$_{\pm\text{0.3}}$} & \cellcolor{blue!18} 50.7\scalebox{0.8}{$_{\pm\text{0.4}}$} \\
\llamaThreeLargePRSingle & \cellcolor{blue!10} 58.8 & \cellcolor{blue!10} 54.3\scalebox{0.8}{$_{\pm\text{0.4}}$} & \cellcolor{blue!10} 59.2\scalebox{0.8}{$_{\pm\text{0.5}}$} & \cellcolor{blue!18} 48.2 & \cellcolor{blue!10} 72.7\scalebox{0.8}{$_{\pm\text{0.3}}$} & \cellcolor{blue!10} 65.9\scalebox{0.8}{$_{\pm\text{0.3}}$} & \cellcolor{blue!26} 60.4\scalebox{0.8}{$_{\pm\text{0.4}}$} & \cellcolor{blue!10} 81.5\scalebox{0.8}{$_{\pm\text{0.3}}$} & \cellcolor{blue!18} 6.1\scalebox{0.8}{$_{\pm\text{0.2}}$} & \cellcolor{blue!10} 48.3\scalebox{0.8}{$_{\pm\text{0.4}}$} & \cellcolor{blue!10} 78.5\scalebox{0.8}{$_{\pm\text{0.3}}$} & \cellcolor{blue!18} 50.7\scalebox{0.8}{$_{\pm\text{0.4}}$} \\
\llamaThreeLargePRALT & \cellcolor{blue!26} 60.4 & \cellcolor{blue!26} 57.7\scalebox{0.8}{$_{\pm\text{0.4}}$} & \cellcolor{blue!18} 59.8\scalebox{0.8}{$_{\pm\text{0.5}}$} & \cellcolor{blue!26} 49.2 & \cellcolor{blue!18} 73.5\scalebox{0.8}{$_{\pm\text{0.3}}$} & \cellcolor{blue!18} 67.6\scalebox{0.8}{$_{\pm\text{0.3}}$} & \cellcolor{blue!26} 60.4\scalebox{0.8}{$_{\pm\text{0.4}}$} & \cellcolor{blue!26} 82.2\scalebox{0.8}{$_{\pm\text{0.3}}$} & \cellcolor{blue!10} 6.0\scalebox{0.8}{$_{\pm\text{0.2}}$} & \cellcolor{blue!26} 50.1\scalebox{0.8}{$_{\pm\text{0.4}}$} & \cellcolor{blue!26} 78.7\scalebox{0.8}{$_{\pm\text{0.3}}$} & \cellcolor{blue!26} 50.9\scalebox{0.8}{$_{\pm\text{0.4}}$} \\
\llamaThreeLargeRTALT & \cellcolor{blue!35} 62.7 & \cellcolor{blue!35} 61.4\scalebox{0.8}{$_{\pm\text{0.2}}$} & \cellcolor{blue!35} 62.1\scalebox{0.8}{$_{\pm\text{0.3}}$} & \cellcolor{blue!35} 50.8 & \cellcolor{blue!43} 75.4\scalebox{0.8}{$_{\pm\text{0.2}}$} & \cellcolor{blue!35} 68.4\scalebox{0.8}{$_{\pm\text{0.2}}$} & \cellcolor{blue!35} 64.1\scalebox{0.8}{$_{\pm\text{0.3}}$} & \cellcolor{blue!35} 82.5\scalebox{0.8}{$_{\pm\text{0.2}}$} & \cellcolor{blue!43} 11.5\scalebox{0.8}{$_{\pm\text{0.2}}$} & \cellcolor{blue!35} 59.2\scalebox{0.8}{$_{\pm\text{0.3}}$} & \cellcolor{blue!35} 79.0\scalebox{0.8}{$_{\pm\text{0.2}}$} & \cellcolor{blue!35} 52.4\scalebox{0.8}{$_{\pm\text{0.3}}$} \\
\llamaThreeLargeFLDALT & \cellcolor{blue!43} 64.2 & \cellcolor{blue!43} 65.7\scalebox{0.8}{$_{\pm\text{0.1}}$} & \cellcolor{blue!51} 63.6\scalebox{0.8}{$_{\pm\text{0.2}}$} & \cellcolor{blue!43} 52.0 & \cellcolor{blue!35} 75.3\scalebox{0.8}{$_{\pm\text{0.1}}$} & \cellcolor{blue!51} 68.5\scalebox{0.8}{$_{\pm\text{0.1}}$} & \cellcolor{blue!43} 65.0\scalebox{0.8}{$_{\pm\text{0.2}}$} & \cellcolor{blue!51} 83.6\scalebox{0.8}{$_{\pm\text{0.1}}$} & \cellcolor{blue!51} 12.1\scalebox{0.8}{$_{\pm\text{0.1}}$} & \cellcolor{blue!43} 59.9\scalebox{0.8}{$_{\pm\text{0.2}}$} & \cellcolor{blue!43} 79.3\scalebox{0.8}{$_{\pm\text{0.1}}$} & \cellcolor{blue!51} 54.4\scalebox{0.8}{$_{\pm\text{0.2}}$} \\
\llamaThreeLargePLDALTBold & \cellcolor{blue!51} 64.4 & \cellcolor{blue!51} 66.1\scalebox{0.8}{$_{\pm\text{0.1}}$} & \cellcolor{blue!43} 63.3\scalebox{0.8}{$_{\pm\text{0.2}}$} & \cellcolor{blue!51} 52.4 & \cellcolor{blue!51} 76.1\scalebox{0.8}{$_{\pm\text{0.1}}$} & \cellcolor{blue!51} 68.5\scalebox{0.8}{$_{\pm\text{0.1}}$} & \cellcolor{blue!51} 65.4\scalebox{0.8}{$_{\pm\text{0.2}}$} & \cellcolor{blue!51} 83.6\scalebox{0.8}{$_{\pm\text{0.2}}$} & \cellcolor{blue!35} 11.4\scalebox{0.8}{$_{\pm\text{0.1}}$} & \cellcolor{blue!51} 60.8\scalebox{0.8}{$_{\pm\text{0.2}}$} & \cellcolor{blue!51} 79.5\scalebox{0.8}{$_{\pm\text{0.1}}$} & \cellcolor{blue!51} 54.4\scalebox{0.8}{$_{\pm\text{0.2}}$} \\
\bottomrule

\end{tabular}

        }
    \end{subfigure}

\vspace{-2mm}
\end{table*}

\begin{table}[t!]
\centering
\footnotesize
\tabcolsep 1.5pt

\caption{
    Benchmark-wise 5-shot performance of \llamaThreeBaseline \ before and \textbf{after} \ALT\ on \PLDAbbr.
    \label{appendix:tb:performance_details}
}

\begin{subfigure}{\linewidth}
    \centering
    \subcaption{Logic.   \label{appendix:tbwhich_task_logical_reasoning}}
    \vspace{-1mm}
    \tabcolsep 2pt
    
\resizebox{\textwidth}{!}{

\begin{tabular}{lccccccccc}
\toprule
{} & bAbiD & FOLIO & LogicNLI & RobustLR & AR-LSAT & LogiQA & ReClor & AbductionR & ART \\
\midrule
\llamaThreeBaseline & 48.7\scalebox{0.8}{$_{\pm\text{1.6}}$} & 50.0\scalebox{0.8}{$_{\pm\text{1.6}}$} & 28.5\scalebox{0.8}{$_{\pm\text{1.0}}$} & 43.2\scalebox{0.8}{$_{\pm\text{0.9}}$} & 20.7\scalebox{0.8}{$_{\pm\text{1.0}}$} & 39.6\scalebox{0.8}{$_{\pm\text{1.2}}$} & 28.7\scalebox{0.8}{$_{\pm\text{0.7}}$} & 52.4\scalebox{0.8}{$_{\pm\text{0.9}}$} & 73.4\scalebox{0.8}{$_{\pm\text{1.1}}$} \\
\llamaThreePLDALTBold & \textbf{55.8}\scalebox{0.8}{$_{\pm\text{0.6}}$} & \textbf{54.5}\scalebox{0.8}{$_{\pm\text{0.6}}$} & \textbf{42.0}\scalebox{0.8}{$_{\pm\text{0.4}}$} & \textbf{62.6}\scalebox{0.8}{$_{\pm\text{0.3}}$} & \textbf{21.1}\scalebox{0.8}{$_{\pm\text{0.4}}$} & \textbf{42.8}\scalebox{0.8}{$_{\pm\text{0.4}}$} & \textbf{29.4}\scalebox{0.8}{$_{\pm\text{0.2}}$} & \textbf{85.5}\scalebox{0.8}{$_{\pm\text{0.2}}$} & \textbf{76.1}\scalebox{0.8}{$_{\pm\text{0.4}}$} \\
\bottomrule
\end{tabular}

}

\end{subfigure}

\begin{subfigure}{\linewidth}
    \centering
    \subcaption{Math.   \label{appendix:tbwhich_task_math}}
    \vspace{-1mm}
    \tabcolsep 3pt

\begin{tabular}{lccccc}
\toprule

{} & \multicolumn{3}{c}{GSM8k} & MATH & MathQA \\
\cmidrule(l){2-4} \cmidrule(l){5-5} \cmidrule(l){6-6}
{} & {} & CoT & CoT (0-shot) & - & - \\
\midrule

\llamaThreeBaseline & 50.2\scalebox{0.8}{$_{\pm\text{1.4}}$} & 51.5\scalebox{0.8}{$_{\pm\text{1.4}}$} & 39.5\scalebox{0.8}{$_{\pm\text{1.3}}$} & 14.1\scalebox{0.8}{$_{\pm\text{0.5}}$} & 42.8\scalebox{0.8}{$_{\pm\text{0.9}}$} \\
\llamaThreePLDALTBold & \textbf{53.6}\scalebox{0.8}{$_{\pm\text{0.5}}$} & \textbf{56.4}\scalebox{0.8}{$_{\pm\text{0.5}}$} & \textbf{48.4}\scalebox{0.8}{$_{\pm\text{0.5}}$} & \textbf{14.3}\scalebox{0.8}{$_{\pm\text{0.2}}$} & \textbf{43.3}\scalebox{0.8}{$_{\pm\text{0.3}}$} \\
\bottomrule
\end{tabular}

\end{subfigure}

\begin{subfigure}{\linewidth}
    \centering
    \subcaption{Coding.   \label{appendix:tbwhich_task_coding}}
    \vspace{-1mm}
    \tabcolsep 4pt

\begin{tabular}{lccccc}

\toprule
{} & HumanEval & MBPP & MBPP+ & MultiPL-E (cpp) & MultiPL-E (go) \\
\midrule
\llamaThreeBaseline & 22.6 & 31.6 & 38.1 & 21.7 & 63.0 \\
\llamaThreePLDALTBold & \textbf{25.9} & \textbf{34.0} & \textbf{39.9} & \textbf{23.0} & \textbf{67.1} \\
\bottomrule

\end{tabular}

\end{subfigure}

\begin{subfigure}{\linewidth}
    \centering
    \subcaption{Natural language inference (NLI).   \label{appendix:tbwhich_task_NLI}}
    \vspace{-1mm}
    \tabcolsep 4pt

\begin{tabular}{lcccc}
\toprule
{} & HELP & MNLI & RTE & SNLI \\
\midrule
\llamaThreeBaseline & 46.4\scalebox{0.8}{$_{\pm\text{0.5}}$} & 68.1\scalebox{0.8}{$_{\pm\text{0.5}}$} & 74.6\scalebox{0.8}{$_{\pm\text{0.9}}$} & 72.6\scalebox{0.8}{$_{\pm\text{0.4}}$} \\
\llamaThreePLDALTBold & \textbf{47.9}\scalebox{0.8}{$_{\pm\text{0.2}}$} & \textbf{75.3}\scalebox{0.8}{$_{\pm\text{0.2}}$} & \textbf{83.1}\scalebox{0.8}{$_{\pm\text{0.3}}$} & \textbf{76.5}\scalebox{0.8}{$_{\pm\text{0.1}}$} \\
\bottomrule
\end{tabular}

\end{subfigure}

\begin{subfigure}{\linewidth}
    \centering
    \subcaption{Others.   \label{appendix:tbwhich_task_others}}
    \vspace{-1mm}
    \tabcolsep 2pt

\resizebox{\textwidth}{!}{

\begin{tabular}{lccccccccc}
\toprule
{} & CommonsenseQA & HellaSwag & SQuAD & WinoGrande & ARCe & ARCc & GPQA & OpenBookQA & SciQ \\
\midrule
\llamaThreeBaseline & 73.9\scalebox{0.8}{$_{\pm\text{1.3}}$} & 61.2\scalebox{0.8}{$_{\pm\text{0.5}}$} & 30.8\scalebox{0.8}{$_{\pm\text{0.0}}$} & 77.4\scalebox{0.8}{$_{\pm\text{1.2}}$} & 84.2\scalebox{0.8}{$_{\pm\text{0.7}}$} & 54.7\scalebox{0.8}{$_{\pm\text{1.5}}$} & \textbf{31.1}\scalebox{0.8}{$_{\pm\text{1.3}}$} & 35.3\scalebox{0.8}{$_{\pm\text{0.7}}$} & \textbf{97.7}\scalebox{0.8}{$_{\pm\text{0.5}}$} \\
\llamaThreePLDALTBold & \textbf{74.8}\scalebox{0.8}{$_{\pm\text{0.4}}$} & \textbf{61.5}\scalebox{0.8}{$_{\pm\text{0.2}}$} & \textbf{33.5}\scalebox{0.8}{$_{\pm\text{0.0}}$} & \textbf{78.1}\scalebox{0.8}{$_{\pm\text{0.5}}$} & \textbf{85.0}\scalebox{0.8}{$_{\pm\text{0.3}}$} & \textbf{55.6}\scalebox{0.8}{$_{\pm\text{0.5}}$} & \textbf{31.1}\scalebox{0.8}{$_{\pm\text{0.5}}$} & \textbf{36.3}\scalebox{0.8}{$_{\pm\text{0.2}}$} & 97.6\scalebox{0.8}{$_{\pm\text{0.2}}$} \\
\bottomrule
\end{tabular}

}

\end{subfigure}

\vspace{-8mm}

\end{table}

\clearpage

\clearpage

\clearpage
\section*{NeurIPS Paper Checklist}

The checklist is designed to encourage best practices for responsible machine learning research, addressing issues of reproducibility, transparency, research ethics, and societal impact. Do not remove the checklist: {\bf The papers not including the checklist will be desk rejected.} The checklist should follow the references and precede the (optional) supplemental material.  The checklist does NOT count towards the page
limit. 

Please read the checklist principles carefully for information on how to answer these questions. For each question in the checklist:
\begin{itemize}
    \item You should answer \answerYes{}, \answerNo{}, or \answerNA{}.
    \item \answerNA{} means either that the question is Not Applicable for that particular paper or the relevant information is Not Available.
    \item Please provide a short (1–2 sentence) justification right after your answer (even for NA). 
\end{itemize}

{\bf The checklist answers are an integral part of your paper submission.} They are visible to the reviewers, area chairs, senior area chairs, and ethics reviewers. You will be asked to also include it (after eventual revisions) with the final version of your paper, and its final version will be published with the paper.

The reviewers of your paper will be asked to use the checklist as one of the factors in their evaluation. While "\answerYes{}" is generally preferable to "\answerNo{}", it is perfectly acceptable to answer "\answerNo{}" provided a proper justification is given (e.g., "error bars are not reported because it would be too computationally expensive" or "we were unable to find the license for the dataset we used"). In general, answering "\answerNo{}" or "\answerNA{}" is not grounds for rejection. While the questions are phrased in a binary way, we acknowledge that the true answer is often more nuanced, so please just use your best judgment and write a justification to elaborate. All supporting evidence can appear either in the main paper or the supplemental material, provided in appendix. If you answer \answerYes{} to a question, in the justification please point to the section(s) where related material for the question can be found.

IMPORTANT, please:
\begin{itemize}
    \item {\bf Delete this instruction block, but keep the section heading ``NeurIPS paper checklist"},
    \item  {\bf Keep the checklist subsection headings, questions/answers and principles below.}
    \item {\bf Do not modify the questions and only use the provided macros for your answers}.
\end{itemize}

\begin{enumerate}

\item {\bf Claims}
    \item[] Question: Do the main claims made in the abstract and introduction accurately reflect the paper's contributions and scope?
    \item[] Answer: \answerYes{} %
    \item[] Justification: Claims stated in \Cref{sec:introduction} is supported by the experimental results in \Cref{sec:results_and_discussions_method,sec:results_and_discussions_tasks}.
    \item[] principles:
    \begin{itemize}
        \item The answer NA means that the abstract and introduction do not include the claims made in the paper.
        \item The abstract and/or introduction should clearly state the claims made, including the contributions made in the paper and important assumptions and limitations. A No or NA answer to this question will not be perceived well by the reviewers. 
        \item The claims made should match theoretical and experimental results, and reflect how much the results can be expected to generalize to other settings. 
        \item It is fine to include aspirational goals as motivation as long as it is clear that these goals are not attained by the paper. 
    \end{itemize}

\item {\bf Limitations}
    \item[] Question: Does the paper discuss the limitations of the work performed by the authors?
    \item[] Answer: \answerYes{} %
    \item[] Justification: \Cref{appendix:sec:limitations}
    \item[] principles:
    \begin{itemize}
        \item The answer NA means that the paper has no limitation while the answer No means that the paper has limitations, but those are not discussed in the paper. 
        \item The authors are encouraged to create a separate "Limitations" section in their paper.
        \item The paper should point out any strong assumptions and how robust the results are to violations of these assumptions (e.g., independence assumptions, noiseless settings, model well-specification, asymptotic approximations only holding locally). The authors should reflect on how these assumptions might be violated in practice and what the implications would be.
        \item The authors should reflect on the scope of the claims made, e.g., if the approach was only tested on a few datasets or with a few runs. In general, empirical results often depend on implicit assumptions, which should be articulated.
        \item The authors should reflect on the factors that influence the performance of the approach. For example, a facial recognition algorithm may perform poorly when image resolution is low or images are taken in low lighting. Or a speech-to-text system might not be used reliably to provide closed captions for online lectures because it fails to handle technical jargon.
        \item The authors should discuss the computational efficiency of the proposed algorithms and how they scale with dataset size.
        \item If applicable, the authors should discuss possible limitations of their approach to address problems of privacy and fairness.
        \item While the authors might fear that complete honesty about limitations might be used by reviewers as grounds for rejection, a worse outcome might be that reviewers discover limitations that aren't acknowledged in the paper. The authors should use their best judgment and recognize that individual actions in favor of transparency play an important role in developing norms that preserve the integrity of the community. Reviewers will be specifically instructed to not penalize honesty concerning limitations.
    \end{itemize}

\item {\bf Theory Assumptions and Proofs}
    \item[] Question: For each theoretical result, does the paper provide the full set of assumptions and a complete (and correct) proof?
    \item[] Answer: \answerNA{} %
    \item[] Justification: Our paper does not include theoretical results.
    \item[] principles:
    \begin{itemize}
        \item The answer NA means that the paper does not include theoretical results. 
        \item All the theorems, formulas, and proofs in the paper should be numbered and cross-referenced.
        \item All assumptions should be clearly stated or referenced in the statement of any theorems.
        \item The proofs can either appear in the main paper or the supplemental material, but if they appear in the supplemental material, the authors are encouraged to provide a short proof sketch to provide intuition. 
        \item Inversely, any informal proof provided in the core of the paper should be complemented by formal proofs provided in appendix or supplemental material.
        \item Theorems and Lemmas that the proof relies upon should be properly referenced. 
    \end{itemize}

    \item {\bf Experimental Result Reproducibility}
    \item[] Question: Does the paper fully disclose all the information needed to reproduce the main experimental results of the paper to the extent that it affects the main claims and/or conclusions of the paper (regardless of whether the code and data are provided or not)?
    \item[] Answer: \answerYes{} %
    \item[] Justification: \Cref{sec:experiments,appendix:sec:experiments}. Further, we release all the resources, including (i) the corpus, (ii) the trained model, and (iii) code for corpus generation, LLM training, and LLM evaluation \footnote{\scriptsize \url{https://anonymous.4open.science/r/ALT/README.md}}.

    \item[] principles:
    \begin{itemize}
        \item The answer NA means that the paper does not include experiments.
        \item If the paper includes experiments, a No answer to this question will not be perceived well by the reviewers: Making the paper reproducible is important, regardless of whether the code and data are provided or not.
        \item If the contribution is a dataset and/or model, the authors should describe the steps taken to make their results reproducible or verifiable. 
        \item Depending on the contribution, reproducibility can be accomplished in various ways. For example, if the contribution is a novel architecture, describing the architecture fully might suffice, or if the contribution is a specific model and empirical evaluation, it may be necessary to either make it possible for others to replicate the model with the same dataset, or provide access to the model. In general. releasing code and data is often one good way to accomplish this, but reproducibility can also be provided via detailed instructions for how to replicate the results, access to a hosted model (e.g., in the case of a large language model), releasing of a model checkpoint, or other means that are appropriate to the research performed.
        \item While NeurIPS does not require releasing code, the conference does require all submissions to provide some reasonable avenue for reproducibility, which may depend on the nature of the contribution. For example
        \begin{enumerate}
            \item If the contribution is primarily a new algorithm, the paper should make it clear how to reproduce that algorithm.
            \item If the contribution is primarily a new model architecture, the paper should describe the architecture clearly and fully.
            \item If the contribution is a new model (e.g., a large language model), then there should either be a way to access this model for reproducing the results or a way to reproduce the model (e.g., with an open-source dataset or instructions for how to construct the dataset).
            \item We recognize that reproducibility may be tricky in some cases, in which case authors are welcome to describe the particular way they provide for reproducibility. In the case of closed-source models, it may be that access to the model is limited in some way (e.g., to registered users), but it should be possible for other researchers to have some path to reproducing or verifying the results.
        \end{enumerate}
    \end{itemize}

\item {\bf Open access to data and code}
    \item[] Question: Does the paper provide open access to the data and code, with sufficient instructions to faithfully reproduce the main experimental results, as described in supplemental material?
    \item[] Answer: \answerYes{} %
    \item[] Justification: we release the code, data, and model.
    \item[] principles:
    \begin{itemize}
        \item The answer NA means that paper does not include experiments requiring code.
        \item Please see the NeurIPS code and data submission principles (\url{https://nips.cc/public/guides/CodeSubmissionPolicy}) for more details.
        \item While we encourage the release of code and data, we understand that this might not be possible, so “No” is an acceptable answer. Papers cannot be rejected simply for not including code, unless this is central to the contribution (e.g., for a new open-source benchmark).
        \item The instructions should contain the exact command and environment needed to run to reproduce the results. See the NeurIPS code and data submission principles (\url{https://nips.cc/public/guides/CodeSubmissionPolicy}) for more details.
        \item The authors should provide instructions on data access and preparation, including how to access the raw data, preprocessed data, intermediate data, and generated data, etc.
        \item The authors should provide scripts to reproduce all experimental results for the new proposed method and baselines. If only a subset of experiments are reproducible, they should state which ones are omitted from the script and why.
        \item At submission time, to preserve anonymity, the authors should release anonymized versions (if applicable).
        \item Providing as much information as possible in supplemental material (appended to the paper) is recommended, but including URLs to data and code is permitted.
    \end{itemize}

\item {\bf Experimental Setting/Details}
    \item[] Question: Does the paper specify all the training and test details (e.g., data splits, hyperparameters, how they were chosen, type of optimizer, etc.) necessary to understand the results?
    \item[] Answer: \answerYes{} %
    \item[] Justification: \Cref{sec:experiments,appendix:sec:experiments}.
    \item[] principles:
    \begin{itemize}
        \item The answer NA means that the paper does not include experiments.
        \item The experimental setting should be presented in the core of the paper to a level of detail that is necessary to appreciate the results and make sense of them.
        \item The full details can be provided either with the code, in appendix, or as supplemental material.
    \end{itemize}

\item {\bf Experiment Statistical Significance}
    \item[] Question: Does the paper report error bars suitably and correctly defined or other appropriate information about the statistical significance of the experiments?
    \item[] Answer: \answerYes{} %
    \item[] Justification: As stated in \Cref{appendix:sec:experiments}.
    \item[] principles:
    \begin{itemize}
        \item The answer NA means that the paper does not include experiments.
        \item The authors should answer "Yes" if the results are accompanied by error bars, confidence intervals, or statistical significance tests, at least for the experiments that support the main claims of the paper.
        \item The factors of variability that the error bars are capturing should be clearly stated (for example, train/test split, initialization, random drawing of some parameter, or overall run with given experimental conditions).
        \item The method for calculating the error bars should be explained (closed form formula, call to a library function, bootstrap, etc.)
        \item The assumptions made should be given (e.g., Normally distributed errors).
        \item It should be clear whether the error bar is the standard deviation or the standard error of the mean.
        \item It is OK to report 1-sigma error bars, but one should state it. The authors should preferably report a 2-sigma error bar than state that they have a 96 \% CI, if the hypothesis of Normality of errors is not verified.
        \item For asymmetric distributions, the authors should be careful not to show in tables or figures symmetric error bars that would yield results that are out of range (e.g. negative error rates).
        \item If error bars are reported in tables or plots, The authors should explain in the text how they were calculated and reference the corresponding figures or tables in the text.
    \end{itemize}

\item {\bf Experiments Compute Resources}
    \item[] Question: For each experiment, does the paper provide sufficient information on the computer resources (type of compute workers, memory, time of execution) needed to reproduce the experiments?
    \item[] Answer: \answerYes{} %
    \item[] Justification: \Cref{appendix:sec:experiments_computation}.
    \item[] principles:
    \begin{itemize}
        \item The answer NA means that the paper does not include experiments.
        \item The paper should indicate the type of compute workers CPU or GPU, internal cluster, or cloud provider, including relevant memory and storage.
        \item The paper should provide the amount of compute required for each of the individual experimental runs as well as estimate the total compute. 
        \item The paper should disclose whether the full research project required more compute than the experiments reported in the paper (e.g., preliminary or failed experiments that didn't make it into the paper). 
    \end{itemize}
    
\item {\bf Code Of Ethics}
    \item[] Question: Does the research conducted in the paper conform, in every respect, with the NeurIPS Code of Ethics \url{https://neurips.cc/public/Ethicsprinciples}?
    \item[] Answer: \answerYes{} %
    \item[] Justification: 
    \item[] principles:
    \begin{itemize}
        \item The answer NA means that the authors have not reviewed the NeurIPS Code of Ethics.
        \item If the authors answer No, they should explain the special circumstances that require a deviation from the Code of Ethics.
        \item The authors should make sure to preserve anonymity (e.g., if there is a special consideration due to laws or regulations in their jurisdiction).
    \end{itemize}

\item {\bf Broader Impacts}
    \item[] Question: Does the paper discuss both potential positive societal impacts and negative societal impacts of the work performed?
    \item[] Answer: \answerYes{} %
    \item[] Justification: \Cref{appendix:sec:social_impacts}
    \item[] principles:
    \begin{itemize}
        \item The answer NA means that there is no societal impact of the work performed.
        \item If the authors answer NA or No, they should explain why their work has no societal impact or why the paper does not address societal impact.
        \item Examples of negative societal impacts include potential malicious or unintended uses (e.g., disinformation, generating fake profiles, surveillance), fairness considerations (e.g., deployment of technologies that could make decisions that unfairly impact specific groups), privacy considerations, and security considerations.
        \item The conference expects that many papers will be foundational research and not tied to particular applications, let alone deployments. However, if there is a direct path to any negative applications, the authors should point it out. For example, it is legitimate to point out that an improvement in the quality of generative models could be used to generate deepfakes for disinformation. On the other hand, it is not needed to point out that a generic algorithm for optimizing neural networks could enable people to train models that generate Deepfakes faster.
        \item The authors should consider possible harms that could arise when the technology is being used as intended and functioning correctly, harms that could arise when the technology is being used as intended but gives incorrect results, and harms following from (intentional or unintentional) misuse of the technology.
        \item If there are negative societal impacts, the authors could also discuss possible mitigation strategies (e.g., gated release of models, providing defenses in addition to attacks, mechanisms for monitoring misuse, mechanisms to monitor how a system learns from feedback over time, improving the efficiency and accessibility of ML).
    \end{itemize}
    
\item {\bf Safeguards}
    \item[] Question: Does the paper describe safeguards that have been put in place for responsible release of data or models that have a high risk for misuse (e.g., pretrained language models, image generators, or scraped datasets)?
    \item[] Answer: \answerNo{} %
    \item[] Justification: 
    \item[] principles:
    \begin{itemize}
        \item The answer NA means that the paper poses no such risks.
        \item Released models that have a high risk for misuse or dual-use should be released with necessary safeguards to allow for controlled use of the model, for example by requiring that users adhere to usage principles or restrictions to access the model or implementing safety filters. 
        \item Datasets that have been scraped from the Internet could pose safety risks. The authors should describe how they avoided releasing unsafe images.
        \item We recognize that providing effective safeguards is challenging, and many papers do not require this, but we encourage authors to take this into account and make a best faith effort.
    \end{itemize}

\item {\bf Licenses for existing assets}
    \item[] Question: Are the creators or original owners of assets (e.g., code, data, models), used in the paper, properly credited and are the license and terms of use explicitly mentioned and properly respected?
    \item[] Answer: \answerYes{} %
    \item[] Justification: All the corpora and benchmarks used in the experiments properly state their licenses.
    \item[] principles:
    \begin{itemize}
        \item The answer NA means that the paper does not use existing assets.
        \item The authors should cite the original paper that produced the code package or dataset.
        \item The authors should state which version of the asset is used and, if possible, include a URL.
        \item The name of the license (e.g., CC-BY 4.0) should be included for each asset.
        \item For scraped data from a particular source (e.g., website), the copyright and terms of service of that source should be provided.
        \item If assets are released, the license, copyright information, and terms of use in the package should be provided. For popular datasets, \url{paperswithcode.com/datasets} has curated licenses for some datasets. Their licensing guide can help determine the license of a dataset.
        \item For existing datasets that are re-packaged, both the original license and the license of the derived asset (if it has changed) should be provided.
        \item If this information is not available online, the authors are encouraged to reach out to the asset's creators.
    \end{itemize}

\item {\bf New Assets}
    \item[] Question: Are new assets introduced in the paper well documented and is the documentation provided alongside the assets?
    \item[] Answer: \answerYes{} %
    \item[] Justification: We will release our corpus.
    \item[] principles:
    \begin{itemize}
        \item The answer NA means that the paper does not release new assets.
        \item Researchers should communicate the details of the dataset/code/model as part of their submissions via structured templates. This includes details about training, license, limitations, etc. 
        \item The paper should discuss whether and how consent was obtained from people whose asset is used.
        \item At submission time, remember to anonymize your assets (if applicable). You can either create an anonymized URL or include an anonymized zip file.
    \end{itemize}

\item {\bf Crowdsourcing and Research with Human Subjects}
    \item[] Question: For crowdsourcing experiments and research with human subjects, does the paper include the full text of instructions given to participants and screenshots, if applicable, as well as details about compensation (if any)? 
    \item[] Answer: \answerNA{} %
    \item[] Justification: This paper does not involve crowdsourcing nor research with human objects.
    \item[] principles:
    \begin{itemize}
        \item The answer NA means that the paper does not involve crowdsourcing nor research with human subjects.
        \item Including this information in the supplemental material is fine, but if the main contribution of the paper involves human subjects, then as much detail as possible should be included in the main paper. 
        \item According to the NeurIPS Code of Ethics, workers involved in data collection, curation, or other labor should be paid at least the minimum wage in the country of the data collector. 
    \end{itemize}

\item {\bf Institutional Review Board (IRB) Approvals or Equivalent for Research with Human Subjects}
    \item[] Question: Does the paper describe potential risks incurred by study participants, whether such risks were disclosed to the subjects, and whether Institutional Review Board (IRB) approvals (or an equivalent approval/review based on the requirements of your country or institution) were obtained?
    \item[] Answer: \answerNA{} %
    \item[] Justification: THis paper does not involve crowdsourcing nor research with human objects.
    \item[] principles:
    \begin{itemize}
        \item The answer NA means that the paper does not involve crowdsourcing nor research with human subjects.
        \item Depending on the country in which research is conducted, IRB approval (or equivalent) may be required for any human subjects research. If you obtained IRB approval, you should clearly state this in the paper. 
        \item We recognize that the procedures for this may vary significantly between institutions and locations, and we expect authors to adhere to the NeurIPS Code of Ethics and the principles for their institution. 
        \item For initial submissions, do not include any information that would break anonymity (if applicable), such as the institution conducting the review.
    \end{itemize}

\end{enumerate}

\end{document}